\documentclass[11pt]{article}

\usepackage[preprint]{acl}

\usepackage{times}
\usepackage{latexsym}

\usepackage[T1]{fontenc}

\usepackage[utf8]{inputenc}

\usepackage{microtype}
\usepackage{tablefootnote} 

\usepackage{inconsolata}

\usepackage{graphicx}
\usepackage{tabularx}
\usepackage{makecell}
\usepackage{subcaption} 
\usepackage{needspace}
\usepackage{times} 
\usepackage{helvet} 
\usepackage{courier} 
\usepackage{float}
\usepackage{placeins}
\usepackage{natbib}  
\usepackage[utf8]{inputenc}
\usepackage{booktabs}
\usepackage{adjustbox}
\usepackage{xcolor}
\usepackage{threeparttable}
\usepackage{amssymb}
\usepackage{caption}
\usepackage{chngcntr}
\usepackage{subcaption}
\usepackage{adjustbox}
\newcolumntype{P}[1]{>{\centering\arraybackslash}p{#1}}
\newcolumntype{M}[1]{>{\centering\arraybackslash}m{#1}}

\usepackage{newfloat}
\usepackage{listings}
\DeclareCaptionStyle{ruled}{labelfont=normalfont,labelsep=colon,strut=off} 
\floatstyle{ruled}
\newfloat{listing}{tb}{lst}{}
\floatname{listing}{Listing}

\usepackage{algorithm}
\usepackage[noend]{algorithmic}
\usepackage{appendix}
\usepackage{tabularx}

\usepackage{multirow}
\usepackage{amsmath}
\usepackage{enumitem}
\usepackage{comment}
\usepackage{physics}
\usepackage{anyfontsize}
\usepackage{lipsum}
\usepackage{listings}
\usepackage{xcolor}
\usepackage[most]{tcolorbox}
\usepackage{lipsum} 
\usepackage{booktabs}
\usepackage{array}
\usepackage{tabularx}
\usepackage{multirow}
\usepackage{graphicx}
\usepackage{subcaption}
\title{DocHop-QA: Towards Multi-Hop Reasoning \\over Multimodal Document Collections}

\newcounter{eqfn}

\newcounter{corrfn}

\author{
    Jiwon Park\textsuperscript{1} \thanks{Equal contribution.}\setcounter{eqfn}{\value{footnote}}, 
    Seohyun Pyeon\textsuperscript{1}\footnotemark[\value{eqfn}], 
    Jinwoo Kim\textsuperscript{1,2}\footnotemark[\value{eqfn}]\thanks{Corresponding author.}\setcounter{corrfn}{\value{footnote}},\\ 
    {\bf Rina Carines Cabral\textsuperscript{2},
    Zhenuyan He\textsuperscript{2},
    Yihao Ding\textsuperscript{3},
    Soyeon Caren Han\textsuperscript{2}\footnotemark[\value{corrfn}]}\\
    \mdseries
    \textsuperscript{1}Pohang University of Science and Technology, Pohang-si, South Korea\\
    \textsuperscript{2}The University of Melbourne, Melbourne, Australia \\
    \textsuperscript{3}The University of Western Australia, Perth, Australia\\
    \texttt{\{jiwon23, seohyun\}@postech.ac.kr}\\
    \texttt{yihao.ding@uwa.edu.au} \\
    \texttt{\{jinwoo.kim.1, rina.cabral, zhenuyan.he, caren.han\}@unimelb.edu.au}\\
}

\begin{document}
\maketitle
\begin{abstract}
Despite rapid progress in large language models (LLMs), current QA benchmarks still overlook the core challenge of real-world scientific information seeking: synthesizing multimodal evidence scattered across multiple documents and structural formats. Existing QAs remain narrow in scope, relying on unimodal text and short-span reasoning that fail to capture the complexity of real information-seeking. We introduce \textbf{DocHop-QA}, a benchmark of 11,379 instances for evaluating multimodal, multi-document, multi-hop scientific QA. Built from publicly available PubMed articles, DocHop-QA incorporates textual passages, tables, and layout cues, enabling cross-document inference without explicit hyperlinks. To scale realistic QA construction, we develop an LLM-driven generation pipeline grounded in 11 scientific reasoning concepts, producing diverse and coherent question--answer pairs. To highlight the utility and versatility of the dataset, we propose a task-driven evaluation framework spanning four settings, including generative answering, multimodal evidence integration and structured index prediction. Experiments show that current models struggle with DocHop-QAs long-context, multi-evidence demands, establishing it as a rigorous testbed for advancing next-generation scientific QA systems.

\end{abstract}

\vspace{-0.3em}
\section{Introduction}
\label{sec:introduction}
Recent advances in Large Language Models (LLMs) have substantially improved Question Answering (QA) performance, particularly in single-document and short-context settings. However, real-world information-seeking rarely conforms to such simplified conditions. In practice, answering complex questions often requires integrating fragmented evidence across multiple documents and heterogeneous formats, including narrative text, tables, and structured layouts \cite{yoon2022sequence}. This setting naturally gives rise to \emph{multi-hop QA}, where models must retrieve and reason over multiple pieces of evidence rather than rely on localized textual cues.
In scientific domains, for example, answering questions such as “What proteins are involved in pathway X, and what experiments validate their roles?” typically demands synthesizing various evidence across several publications.

Despite recent progress, deploying LLMs for multi-hop QA in realistic document-centric environments remains challenging, reflecting fundamental NLP failure modes including sensitivity to retrieval quality, difficulty grounding semi-structured tables, and error accumulation across long reasoning chains. Specifically, 
omitting a single critical evidence node can break the reasoning 
chain, while retrieving loosely related passages introduces 
distraction and confusion ~\cite{zhu2025mitigating, bai2025longbench}. Yet existing benchmarks insufficiently capture these failure modes, remaining largely confined to Wikipedia-sourced, unimodal settings with short extractive answers. 

To fill this gap, we introduce DocHop-QA, a large-scale benchmark for complex, multi-hop QA over multimodal, multi-document corpora, constructed from scientific articles. Its generation methodology is domain-agnostic and scalable, making it applicable to a wide range of scientific domains. Constructed from PubMed articles, it includes unstructured text, structured tables, and layout features, without relying on predefined hyperlinks or annotated reasoning chains. To efficiently generate diverse and challenging QA pairs, we present a scalable LLM-driven pipeline based on 11 real-world scientific reasoning concepts.
To evaluate the applicability and flexibility of DocHop-QA, we benchmark its performance across four representative QA tasks: (1) Generative Reasoning, (2) Structured Generative Answering, (3) BBox Entity Index Extraction, and (4) XML Entity Index Extraction. DocHop-QA intentionally isolates the multi-document evidence integration stage in a controlled setting, rather than modeling the full end-to-end retrieval pipeline. Each instance requires identifying multiple relevant evidence spans across documents and synthesising a coherent natural-language answer, targeting the core capability of cross-document reading and reasoning. This design enables rigorous evaluation of evidence aggregation and cross-document synthesis, a fundamental yet underexplored challenge in scientific document understanding.

Our key contributions are as follows:
1) We introduce \textbf{DocHop-QA}, a large-scale benchmark for multi-hop QA over multimodal, multi-document scientific corpora, incorporating text, tables, and layout features.
2) We develop a \textbf{scalable LLM-driven generation pipeline} guided by 11 reasoning concepts inspired by real-world scientific inquiry, enabling the automated creation of diverse and challenging multi-hop QA instances without gold chains or hyperlinks.
3) We \textbf{evaluate DocHop-QA across four QA tasks}, ranging from structured index prediction to generative answering, demonstrating its broad applicability for assessing both discriminative and generative reasoning over textual, structural, and visual information.

\vspace{-0.3em}
\section{Related Work}
\label{sec:related_works}
\paragraph{Multi-hop QA}  

Multi-hop QA tasks span multiple documents, pages, and modalities, reflecting a realistic information seeking process. Early benchmarks such as QAngaroo \cite{welbl2018constructing} and HotpotQA \cite{yang2018hotpotqa} pioneered multi-hop reasoning over text, but rely on predefined knowledge bases or relatively short contexts.  Later, HybridQA \cite{chen2020hybridqa} and WebQA \cite{chang2022webqa} expanded evidence modalities by incorporating semi-structured data such as tables, yet still lean toward web-derived resources and exhibit evidence that is relatively locally concentrated.
More recent datasets, including FanOutQA \cite{zhu2024fanoutqa}, MEQA \cite{li2024meqa}, LLeQA \cite{louis2024interpretable} and Loong \cite{wang2024leave}  increase difficulty along axes such as hop depth or context length to probe model reasoning under greater complexity. Nonetheless, many remain text-centric and predominantly feature short answers, leaving a gap to the paragraph-level synthesis demanded in scientific information seeking. Moreover, many still depend on explicitly structured representations such as relational DBs \cite{zhu2025mitigating} and knowledge graphs \cite{kim2025biohopr} to support complex reasoning. More recently, NOVELHOP \cite{gupta2025novelhopqa} targets multi-hop reasoning over long narrative contexts but is constrained to unimodal text. MEBench \cite{lin2025mebench} extends this to entity-dense scenarios but remains text-centric, while M3SciQA \cite{li2024m3sciqa} introduces multimodal multi-document QA yet relies on citation-based anchor links. In contrast, DocHop-QA constructs semantically related pairs of real scientific documents, requiring models to aggregate and synthesize paragraph- and table-level evidence across documents without relying on explicit structural anchors.

\paragraph{Visually Rich Document QA}
Visually rich document QA extends question answering beyond flat text to real-world documents with complex layouts, tables, and figures.  Early datasets such as DocVQA \cite{mathew2021docvqa} and InfographicVQA \cite{mathew2022infographicvqa} focus on single page document images, often framed as span extraction or retrieval tasks.  SlideVQA \cite{tanaka2023slidevqa} and BLIVA \cite{hu2024bliva}   introduce multi-hop reasoning, but are primarily studied in single-document settings. 
More recent benchmarks such as ECG-QA \cite{oh2023ecg}, SPIQA \cite{pramanick2024spiqa} , MMVQA \cite{ding2024mvqa}, and VisDoMBench \cite{suri2025visdom} further extend VRD QA toward multimodal and multi-document settings. However, they largely emphasize retrieval-oriented evaluation, offer limited open-ended multi-hop reasoning, and mostly manually constructed for specific document types, limiting scalability. In contrast, DocHop-QA supports scientific document reasoning, incorporating text, tables, and layout features without manual construction constraints. Table~\ref{tab:qa_datasets_comparison} summarizes the comparison of prior datasets with ours across key dimensions.

\begin{table}[t]
    \centering
    \renewcommand{\arraystretch}{1.1}
    \begin{adjustbox}{max width=\linewidth}
    \begin{tabular}{l|c|c|c|c|c|c|c}
        \hline
        \textbf{Dataset} & \textbf{M Hop} & \textbf{VRD} & \textbf{\#QA} & \textbf{Q Gen} & \textbf{Source} & \textbf{M Doc} & \textbf{Modality} \\
        \hline
        HotpotQA & {\color{green!60!black}O} & X & 113K & H & Wiki & X & Tx \\
        MEQA & {\color{green!60!black}O} & X & 2.2K & LLM + H & WikiEvents & X & Tx \\
        NOVELHOPQA & {\color{green!60!black}O} & X & 3.8K & LLM & Gutenberg & X & Tx \\
        HybridQA & {\color{green!60!black}O} & X & 70K  & H & Wiki & {\color{green!60!black}O} & Tx+Tab \\
        Loong & {\color{green!60!black}O} & X & 1.6K & LLM + H & Industry & {\color{green!60!black}O} & Tx \\
        FanOutQA & {\color{green!60!black}O} & X & 1K & H & Wiki & {\color{green!60!black}O} & Tx \\
        M3SciQA & {\color{green!60!black}O} & X & 1.4K & H & EMNLP Papers & {\color{green!60!black}O} & Tx+Tab+Img \\
        \hline
        DocVQA & X & {\color{green!60!black}O} & 50K & H & Industry & X & Tx+Img \\
        SlideVQA & X & {\color{green!60!black}O} & 14K & H & Slideshare & X & Tx+Img \\
        ECG-QA & X & {\color{green!60!black}O} & 414K & H & PTB-XL & X & Tx+Img \\
        SPIQA & X & {\color{green!60!black}O} & 270K & LLM + H & arXiv & X & Tx+Tab+Img \\
        MMVQA & X & {\color{green!60!black}O} & 263K & LLM + H & PubMed & X & Tx+Tab+Img \\
        VisDoMBench & X & {\color{green!60!black}O} & 2.2K & LLM & Wikip.+arXiv &{\color{green!60!black}O} & Tx+Tab+Img\\
        \hline

        \textbf{Ours} & {\color{green!60!black}O} & {\color{green!60!black}O} & 11K & LLM + H & PubMed & {\color{green!60!black}O} & Tx+Tab+Img \\
        \hline
    \end{tabular}
    \end{adjustbox}
    \caption{\textbf{Comparison of QA dataset benchmarks.} M Hop: Multi Hop, VRD: Visually Rich Document, Q Gen.: Question Generation method (H: Human, LLM: Large Language Model), M Doc: Multiple Document, Modality (Tx: Text, Tab: Table, Img: Image). Further details about Multi-hop and reasoning are in Table~\ref{tab:qa_datasets_comparison_recent}}.
    \label{tab:qa_datasets_comparison}
    \vspace{-3em}
\end{table}

\vspace{-0.5em}
\begin{figure*}[t]

    \centering

    \includegraphics[width=\textwidth]{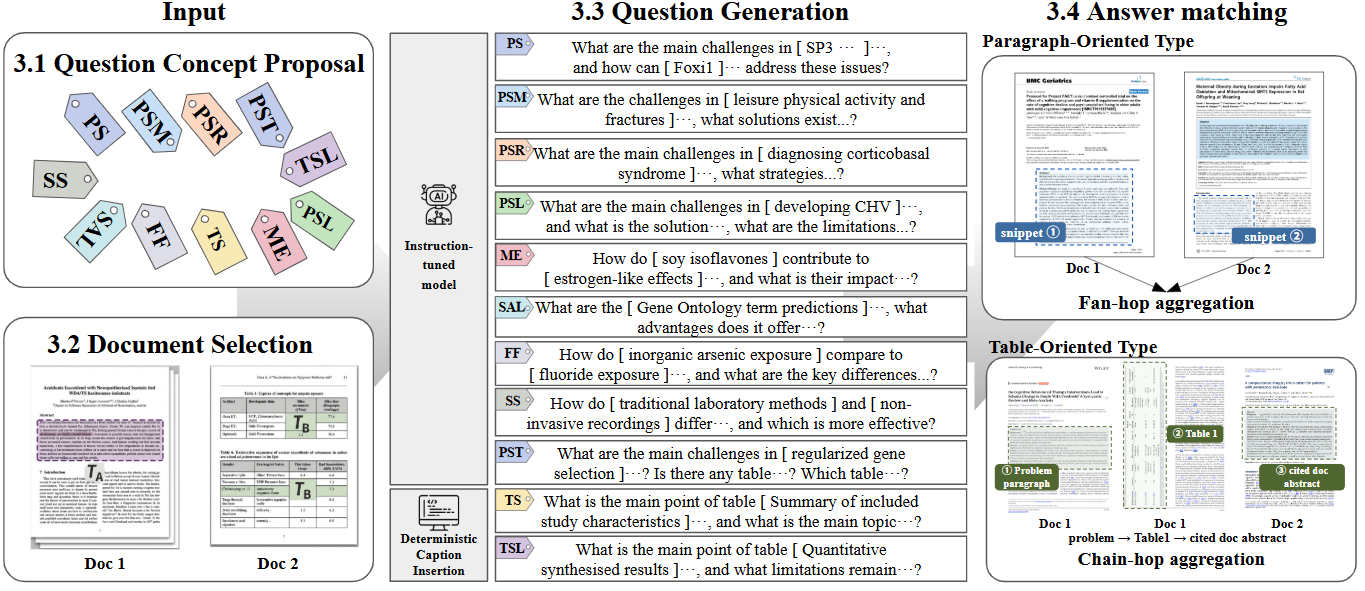}
    \includegraphics[width=0.48\textwidth]{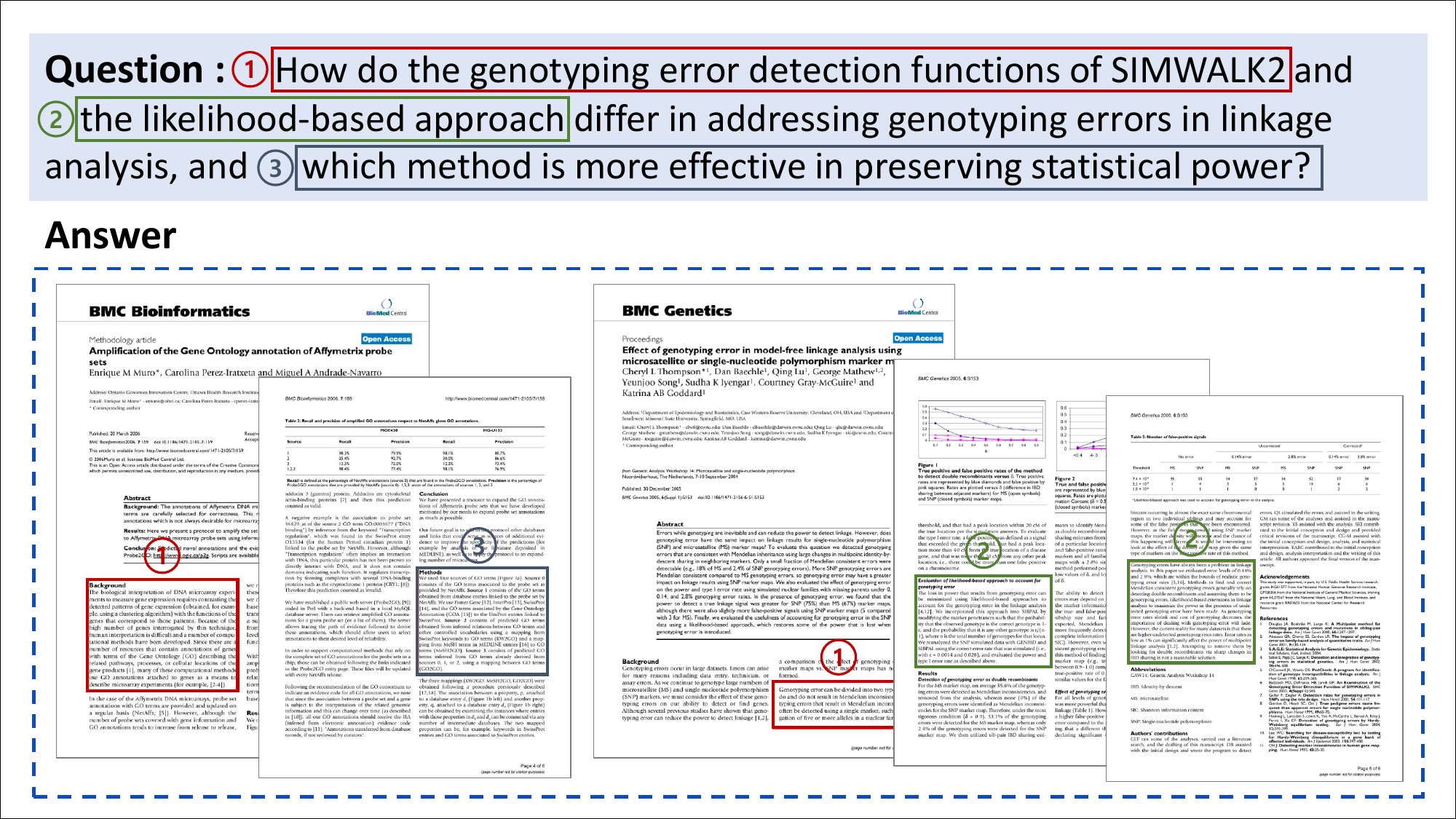} 
    \hfill    \includegraphics[width=0.48\textwidth]{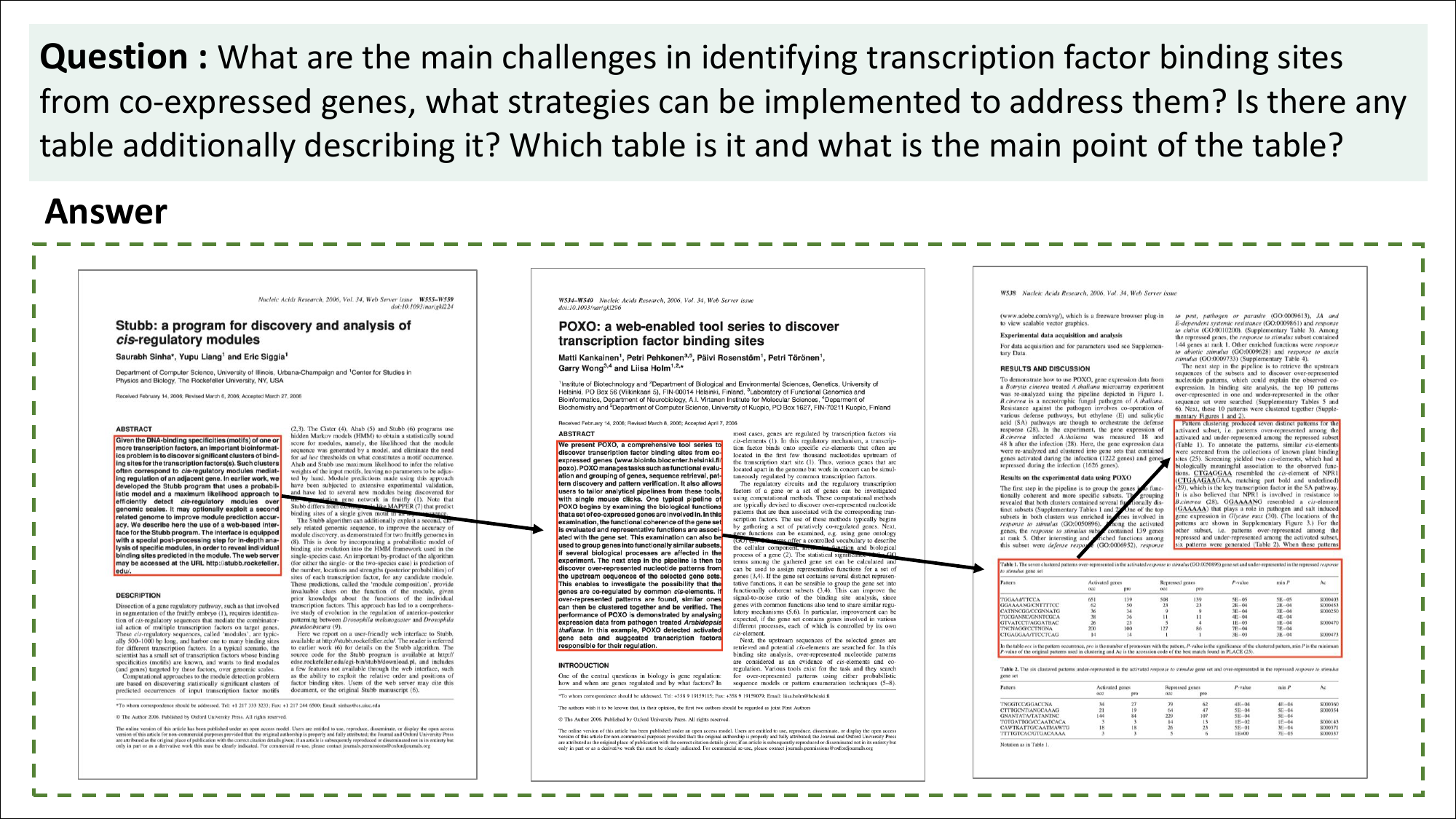} 
    \caption{\textbf{Overview of the DocHop-QA construction pipeline across four stages} (1) Question Concept Proposal, (2) Document Selection, (3) Question Generation, and (4) Answer Matching. Bottom row shows representative Comparison-type (left) and PST-type (right) QA instances.}

    \label{fig:pipeline}
\vspace{-0.8em}
\end{figure*}

\begin{table*}[t]
\centering
\begin{adjustbox}{max width =\linewidth}
{\fontsize{11pt}{12pt}\selectfont
\renewcommand{\arraystretch}{1.2}
\centering
\begin{tabular}{M{1.8cm}|M{2.2cm}|M{6cm}|M{1.4cm}|M{7cm}|M{11cm}}
\hline
\textbf{Type} &  \textbf{Subtype} & \textbf{Keyword} & \textbf{Q Type}& \textbf{Question Concept} & \textbf{Question Example} \\
\hline
    & & \textless \textbf{P}roblem\textgreater\textless sep\textgreater\textless \textbf{S}olution\textgreater & PS(1215) & What are the main challenges in [ ], and how can [ ] address these issues? 
    & What are the main challenges in understanding the dual role of the transcription factor Sp3 in tumor progression, and how can the forkhead transcription factor Foxi1 address these issues by regulating the expression of key subunits of the vacuolar H+ ATPase?\\
\cline{3-6}
    & & \textless \textbf{P}roblem\textgreater\textless sep\textgreater\textless \textbf{S}olution\textgreater\newline\textless sep\textgreater\textless \textbf{M}echanism\textgreater & PSM (720) & What are the challenges in [ ], what solutions exist to address these challenges, and how do these solutions help resolve the issues? 
    & What are the challenges in the relationship between leisure physical activity and the risk of osteoporotic fractures in men, what solutions exist to address these challenges, and how do these solutions help resolve the issues?\\
    \cline{3-6}
    &Non-Comparison & \textless \textbf{P}roblem\textgreater\textless sep\textgreater\textless \textbf{S}olution\textgreater\newline\textless sep\textgreater\textless \textbf{R}esult/effect\textgreater & PSR (977) & What are the main challenges in [ ], what strategies can be implemented to address them, and what are the observed effects of these strategies? 
    & What are the main challenges in accurately diagnosing corticobasal syndrome, what strategies can be implemented to address them, and what are the observed effects of these strategies?\\
    \cline{3-6}
    \shortstack{Paragraph-\\Oriented\\(75.3\%)}& & \textless \textbf{P}roblem\textgreater\textless sep\textgreater\textless \textbf{S}olution\textgreater\newline\textless sep\textgreater\textless \textbf{L}imitation\textgreater & PSL (1,269) & What are the main challenges in [ ], and how can [ ] address these issues? Additionally, what are the limitations of this approach? 
    & What are the main challenges in developing consumer health vocabulary (CHV) due to the heterogeneity and ambiguity of consumer expressions, and what is the solution to this challenge? Additionally, what are the limitations of this approach? \\
    \cline{3-6}
    & & \textless \textbf{M}echanism\textgreater\textless sep\textgreater\textless \textbf{E}ffect\textgreater & ME (1,155) & How does [ ] contribute to [ ], and what is its impact on [ ]? 
    & How do soy isoflavones contribute to estrogen-like effects, and what is their impact on male breast cancer incidence? \\
    \cline{3-6}
    & & \textless \textbf{S}tudytopic\textgreater\textless sep\textgreater\textless \textbf{A}dvantage\textgreater\newline\textless sep\textgreater\textless \textbf{L}imitation\textgreater & SAL (850) & What are the [ ], what advantages does it offer, and what limitations exist? 
    & What are the Gene Ontology term predictions, what advantages does it offer, and what limitations exist?\\
\cline{2-6}
    & \multirow{2}{*}{\vspace{-1.1cm} Comparison}
    & \textless \textbf{F}eature\textgreater\textless sep\textgreater\textless \textbf{F}eature\textgreater & FF (1,229) & How does [ ] compare to [ ], and what are the key similarities and differences? 
    & How do the effects of inorganic arsenic exposure on children's intellectual function compare to those of fluoride exposure, and what are the key similarities and differences?\\
    \cline{3-6}
    & & \textless \textbf{S}olution\textgreater\textless sep\textgreater\textless \textbf{S}olution\textgreater & SS (1,152) & How do [ ] and [ ] differ in addressing [ ], and which approach is more effective? 
    & How do traditional laboratory and clinical research methods and non-invasive abdominal recordings differ in addressing the prediction of preterm labor, and which approach is more effective? \\
\hline
  &Non-Refer Table & \textless \textbf{P}roblem\textgreater\textless sep\textgreater\textless \textbf{S}olution\textgreater\newline\textless sep\textgreater\textless \textbf{T}able\textgreater & PST (874) & What are the main challenges in [ ], what strategies can be implemented to address them? Is there any table additionally describing it? Which table is it and what is the main point of the table? 
    & What are the main challenges in regularized gene selection in cancer microarray meta-analysis, what strategies can be implemented to address them? Is there any table additionally describing it? Which table is it and what is the main point of the table?\\
    \cline{2-6}
  \shortstack{Table-\\Oriented\\(24.7\%)}& \multirow{2}{*}{\vspace{-0.7cm} Refer Table}& \textless \textbf{T}able\textgreater\textless sep\textgreater\textless \textbf{S}tudytopic\textgreater\newline\textless sep\textgreater\textless \textbf{L}imitation\textgreater& TSL (573) & What is the main point of Table [ ] in document [ ], and what is the main topic of the referenced paper? Also, what limitations remain in the referenced paper? 
    & What is the main point of table Quantitative synthesised results in given document, and what is the main topic of the referenced paper? Also, what limitations remain in the referenced paper?\\
    \cline{3-6}
    & & \textless \textbf{T}able\textgreater\textless sep\textgreater\textless \textbf{S}tudytopic\textgreater& TS (1,365) &What is the main point of Table [ ] in document [ ], and what is the main topic of the referenced paper? 
    & What is the main point of table Summary of included study characteristics in given document, and what is the main topic of the referenced paper?\\
\hline
\end{tabular}
}
\end{adjustbox}
\caption[\textbf{Illustrative examples of the proposed question concepts}, categorized by (1) \textbf{Type}, denoting the primary answer modality; (2) \textbf{Subtype}, indicating the reasoning structure; (3) \textbf{Keyword}, representing the core reasoning elements and semantic placeholders used in question generation; (4) \textbf{Q Type}, a shorthand for the placeholder composition; (5) \textbf{Question Concept}, the abstract question structure using placeholders; and (6) \textbf{Question Example}, a concrete instantiation of each question concept. This design enables controlled, diverse, and semantically grounded multi-hop QA generation across heterogeneous document modalities.]{\textbf{Illustrative examples of the proposed question concepts}, categorized by (1) \textbf{Type}, denoting the primary answer modality; (2) \textbf{Subtype}, indicating the reasoning structure; (3) \textbf{Keyword}, representing the core reasoning elements and semantic placeholders used in question generation; (4) \textbf{Q Type}, a shorthand for the placeholder composition; (5) \textbf{Question Concept}, the abstract question structure using placeholders; and (6) \textbf{Question Example}, a concrete instantiation of each question concept. This design enables controlled, diverse, and semantically grounded multi-hop QA generation across heterogeneous document modalities.\footnotemark}
\label{tab:concept_type_examples}
\vspace{-1em}
\end{table*}

\section{Dataset Construction}
\label{sec:dataset construction}
DocHop-QA is constructed through a four-stage pipeline illustrated in
Figure~\ref{fig:pipeline}: (1) proposing diverse question concepts,
(2) selecting document pairs suitable for multi-hop reasoning,
(3) generating questions and (4) matching answers.
We build on the PubMed Central (PMC) open access repository which
provides structured XML and PDF versions of biomedical and life science
literature.
From roughly 2M full text scientific articles we obtained approximately
20K document pairs covering 9,250 unique documents forming a controlled
testbed for cross-document evidence integration over heterogeneous
content.

\vspace{-0.5em}
\subsection{Question Concept Proposal}
\label{sec:question_concept_proposal}
To enable systematic and scalable multi-hop question generation we
define 11 \emph{question concepts} grounded in reasoning patterns
frequently observed in PubMed articles.
The concepts were derived through a human-in-the-loop process.
We first used NotebookLM to propose cross-document question candidates
grounded in uploaded source documents which authors then reviewed and
refined into 90 seed QA pairs.
Qualitative analysis identified core semantic keywords such as
\texttt{Problem}, \texttt{Solution} and \texttt{Effect} capturing
fundamental components of scientific reasoning.
Using these as scaffolds we generated an additional 1,000 multi-hop
QA pairs with ChatGPT through prompts tailored to elicit multi-step
reasoning.
We then consolidated structurally overlapping or infrequent scaffolds
into 11 representative question concepts summarised in Table~\ref{tab:concept_type_examples}. Paragraph-Oriented concepts follow a fan-hop aggregation pattern while
Table-Oriented concepts follow a chain-hop sequential inference pattern
as illustrated in Figure~\ref{fig:pipeline}.

\vspace{-0.5em}

\footnotetext{See Fig.~\ref{fig:distance_by_qtypes} for concept-wise distributions of section and paragraph level answer snippet positions.}

\subsection{Document Selection}
\label{sec:document_selection}
\vspace{-0.5em}
We selected document pairs by
ensuring topical alignment without predefined explicit links between documents.
The selection strategy differs by question type to reflect the distinct reasoning structures each type demands.

\vspace{-0.5em}
\paragraph{Paragraph-Oriented Type}

Document pairs are selected by keyword cooccurrence and lightweight
semantic similarity over titles and abstracts without predefined
citation links between documents.
This encourages open ended evidence aggregation across semantically
related but structurally unlinked documents.
The detailed process is provided in Appendix~\ref{appendix:document_selection}.

\vspace{-0.5em}
\paragraph{Table-Oriented Type}
Unlike paragraph-oriented settings, Table-Oriented types intentionally use citation-linked tables to model structured chain-hop reasoning. For the PST type we retained only pairs in which at least one table appears in a results section as these tables typically summarise intervention outcomes.
For TS and TSL types which model reasoning over tables citing external studies we targeted systematic review articles containing
citation linked tables. These citation links serve as intra document structural anchors for
sequential evidence transitions rather than as inter-document pairing criteria.

\vspace{-0.6em}
\subsection{Question Generation}
\label{sec:qa_generation}
\paragraph{Paragraph-Oriented Type}
We tuned an instruction following model on instances pairing question concept templates with document abstracts and gold question examples 
(Appendix~\ref{appendix:traindatasetexample}).
We instruction-tuned the Qwen2-Instruct model using Unsloth’s 4-bit LoRA framework\cite{hu2021loralowrankadaptationlarge}\footnote{\url{https://huggingface.co/unsloth/Qwen2-7B-Instruct-bnb-4bit}}, chosen for its open-source availability, enabling reproducible and cost-efficient large-scale question generation. To preserve controllability and semantic consistency, we trained separate models for each question category (comparison, non-comparison, and PST). Joint training proved suboptimal, frequently omitting contrastive cues in comparison questions and producing fragmented reasoning in non-comparison and PST cases. During inference, we supplied a few-shot prompt that includes the concept, 2 abstracts, and 2 example questions using a step by step reasoning skeleton.

\vspace{-0.3em}
\paragraph{Table-Oriented Type}
For the PST type, questions were generated using the same model based approach as the Paragraph-Oriented type, but differed in the composition of the input to better reflect the table content. For TS and TSL types we designed a deterministic procedure in which
table captions are inserted directly into predefined question
placeholders.
This design is appropriate because the reasoning transitions for these
types are explicitly defined by table references making LLM-based generation unnecessary while preserving full structural
controllability. Post processing filters applied to all types are described in Appendix~\ref{appendix:postprocessing}.
\vspace{-0.5em}


\vspace{-1em}
\subsection{Answer Matching}
\label{sec:answer_matching}
In DocHop-QA, answer retrieval is guided by two distinct reasoning patterns\footnote{Further details are provided in Appendix~\ref{appendix:detailed_answer_matching}}: \textit{fan-hop} for Paragraph-Oriented types, where semantically related content is aggregated from documents \cite{zhu2024fanoutqa}, and \textit{chain-hop} for Table-Oriented types, which require explicit links between structured and unstructured content \cite{xu2021exploiting, chen2019multi}. Following the reasoning patterns illustrated in Figure~\ref{fig:pipeline} our matching pipeline adopts unified retrieval for Paragraph-Oriented types and tailored strategies for Table-Oriented ones.

\vspace{-0.2em}
\paragraph{Paragraph-Oriented Type} For Paragraph-Oriented types, evidence is dispersed across documents. 
Semantically complementary five snippets are retrieved independently from each source document and aggregated without explicit cross-referencing between sources.
Candidate segments are ranked by hybrid lexical and semantic similarity to the question and only instances with sufficiently
coherent evidence are retained.

\vspace{-0.2em}
\paragraph{Table-Oriented Type}
For Table-Oriented types, evidence is sequential. We constructed up to four snippets per instance, connecting problem and solution paragraphs, tables, and referenced content across documents. Retrieval combines semantic similarity with keyword filtering and
applies logical chaining when anchor references are present.
Instances with low similarity were filtered to preserve dataset
integrity.
Detailed retrieval logic is provided in Appendix~\ref{appendix:detailed_answer_matching}.

\section{Dataset Quality Assurance}
\label{sec:dataset_evaluation}
\vspace{-0.3em}
To evaluate the quality of the DocHop-QA dataset, we define three criteria assessed via a 4-point rubric: \textbf{\textit{Reality/Fluency}} ensures that questions are natural, plausible, and meaningful within the context. \textbf{\textit{Accuracy}} assesses whether each snippet meaningfully addresses the question and contributes to forming a complete answer. \textbf{\textit{Completeness}} evaluates whether the retrieved snippets collectively form a sufficient answer. 
\vspace{-0.8em}
\paragraph{Human-LLM Quality assurance}
To validate dataset quality, we conduct a human evaluation on 50 carefully selected QA instances, covering all 11 question concepts, both paragraph- and table-oriented types, and diverse hop patterns(annotator demographics, IRB statement, and evaluation interface details are provided in Appendix~\ref{appendix:human_quality_assurance}). We report Gwet's AC2 as the primary agreement metric, as it is robust to marginal distribution skewness commonly observed in quality annotation tasks. Human annotations are compared against three instruction-tuned LLM judges: GPT-4o\footnote{\url{https://openai.com/index/hello-gpt-4o/}}, Gemini2.5~Flash\footnote{\url{https://cloud.google.com/vertex-ai/generative-ai/docs/models/gemini/2-5-flash}}, and Qwen3.5-9B, using the same rubric (prompt details in Appendix~\ref{appendix:llm-quality-assurance}), all of which show substantial agreement with human judgments across all criteria (Table~\ref{tab:gwet_ac2_summary}). Additional metrics including Spearman's $\rho$ and Krippendorff's $\alpha$ are provided in Appendix~\ref{appendix:quality-assurance}.
The evaluation is then extended to the full dataset using Qwen3.5-9B, an open-source model that achieves competitive agreement while remaining locally deployable. Across all 11,379 instances, approximately \textbf{92\%}, \textbf{87\%}, and \textbf{76\%} of QA pairs receive a score of 3 or above on Reality/Fluency, Accuracy, and Completeness, respectively, suggesting that the quality trends confirmed by human evaluation generalize to the entire DocHop-QA corpus. Detailed results are provided in Appendix~\ref{appendix:quality-assurance}.

\begin{table}[h]
\centering
\small
\setlength{\tabcolsep}{3pt}
\resizebox{0.85\columnwidth}{!}{
\begin{tabular}{l l c}
\toprule
\textbf{Criterion} & \textbf{Model} & \textbf{AC2 (ordinal)} \\
\midrule
\multirow{3}{*}{\textbf{Completeness}} 
    & GPT-4o & 0.555 \\
    & Gemini2.5    & 0.331 \\
    & Qwen3.5-9B    & 0.342 \\
\midrule
\multirow{3}{*}{\textbf{Accuracy}} 
    & GPT-4o & 0.632 \\
    & Gemini2.5    & 0.424 \\
    & Qwen3.5-9B    & 0.741 \\
\midrule
\multirow{3}{*}{\textbf{Reality}} 
    & GPT-4o & 0.874 \\
    & Gemini2.5    & 0.911 \\
    & Qwen3.5-9B    & 0.935 \\
\bottomrule
\end{tabular}
}
\caption{Human–LLM agreement measured by Gwet's AC2 agreement coefficients on 50 selected QA instances}
\label{tab:gwet_ac2_summary}
\end{table}

\vspace{-1.2em}
\section{Dataset Analysis}
\label{sec:Dataset_Analysis}

\begin{figure}[!t]
    \centering
\begin{subfigure}[t]{0.48\textwidth}
    \includegraphics[width=\linewidth]{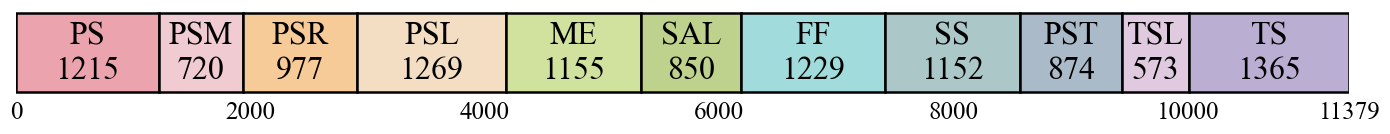}
    \caption{\textbf{Q Types distribution}}
\end{subfigure}
\begin{subfigure}[b]{0.22\textwidth}
\includegraphics[width=\linewidth,height=3cm]{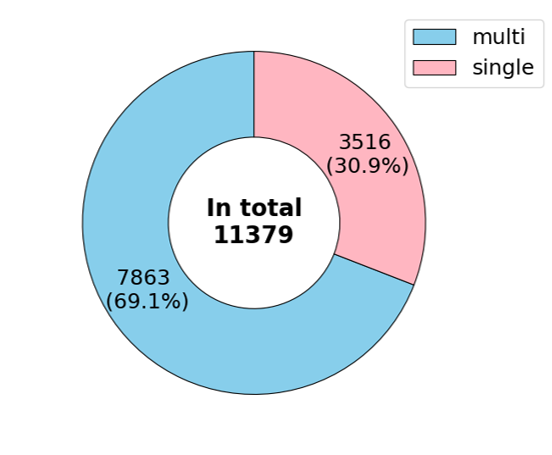}
    \vspace{-0.6cm}
    \caption{\textbf{Used Document Type}}
    \label{fig:Used Document Type}
    \end{subfigure}
\begin{subfigure}[b]{0.23\textwidth}
    \includegraphics[width=\linewidth,height=2.8cm]{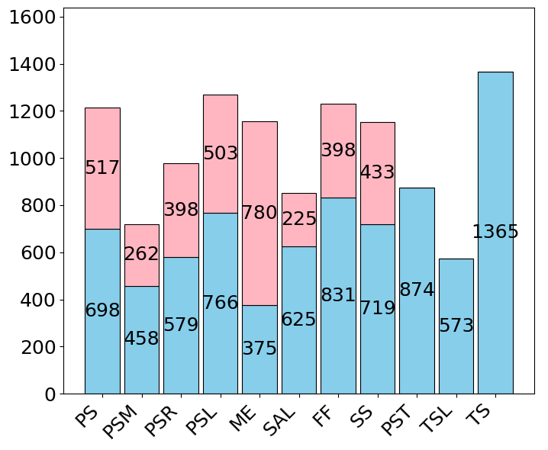}
    \vspace{-0.6cm}
    \caption{\textbf{QA Distribution}}
    \label{fig:QA Distribution}
\end{subfigure}

\vspace{-0.5em}    
\caption{\textbf{Overview of document usage across question types.} Multi-document reasoning dominates.}
\label{fig:overview_of_q_types_and_document_usage}
\vspace{-0.8em}
\end{figure}

DocHop-QA comprised 11,379 instances spanning 11 evenly distributed question concepts, each representing approximately 10\% of dataset, with both single- and multi-document instances across Question concept types shown in Figure~\ref{fig:overview_of_q_types_and_document_usage}. The following analyses characterize DocHop-QA from three perspectives: (1) Document-level complexity, (2) Answer Dispersion, and (3) Reasoning complexity.

\vspace{-0.5em}
\subsection{Dataset Coverage}
\label{sec:document_complexity}
\begin{figure}[!t]
    \centering
\includegraphics[width=0.48\textwidth]{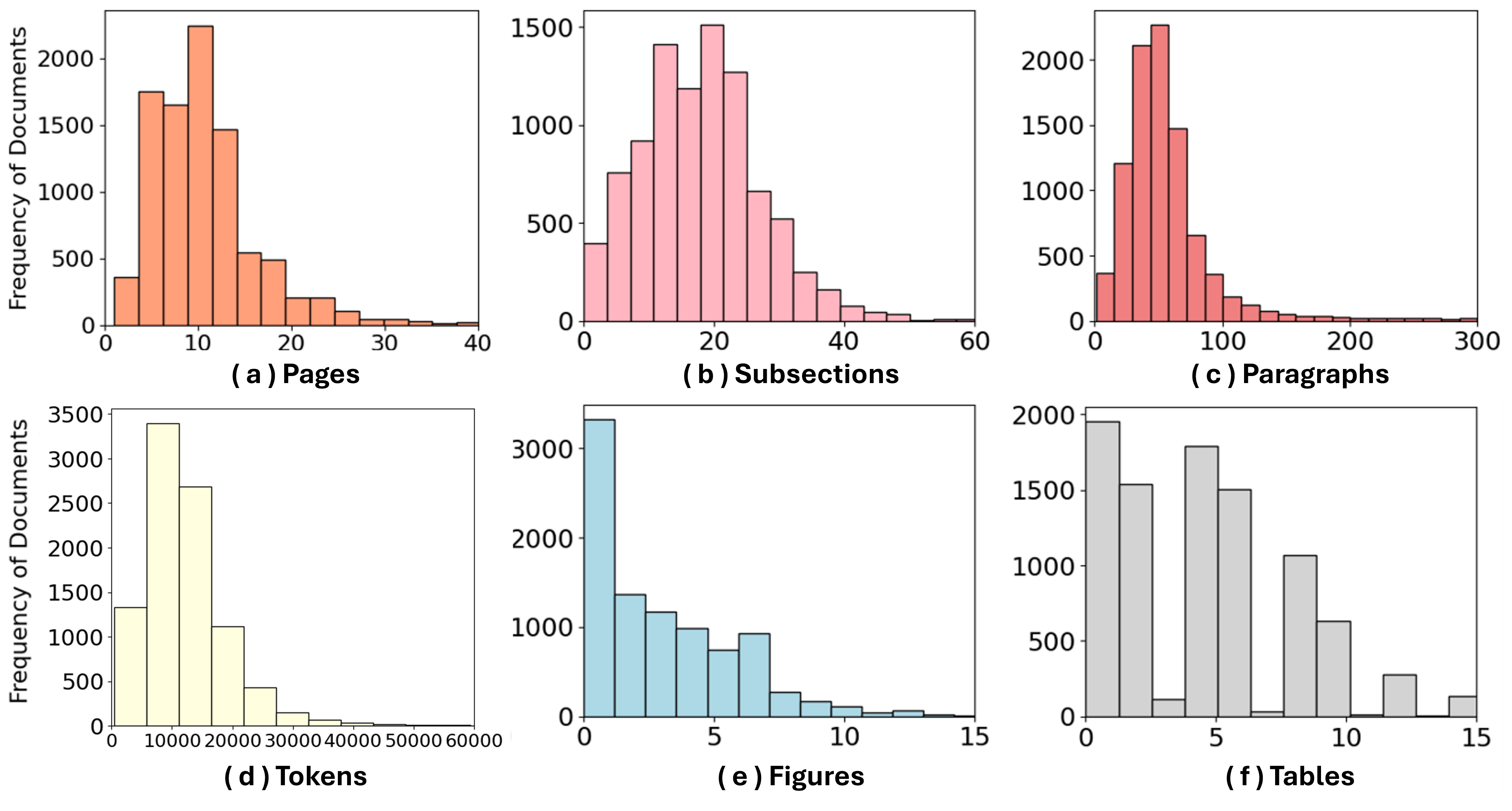}
    \vspace{-0.6cm}
    \caption{\textbf{Distribution of document component types.} DocHop-QA includes multi-page, multi-section VRDs, supporting complex multi-hop and multimodal QA.}
    \label{fig:Document analysis}
    \vspace{-1.5em}
\end{figure}

As shown in Figure~\ref{fig:Document analysis}, most documents span \textasciitilde10 pages, contain \textasciitilde20 subsections and \textasciitilde60 paragraphs totaling 10,000 to 15,000 tokens, closely reflecting the scale and complexity of real-world scientific literature. The questions constructed from these documents inherit their semantic richness, reflecting the depth and complexity of biomedical literature. Furthermore, keyword analysis across 11 question concepts (see Appendix~\ref{appendix:wordclouds}) reveals both topical diversity and domain specificity, with biomedical terms dominating, as expected from PubMed-sourced data.

\subsection{Answer Dispersion}
\label{sec:answer_dispersion}
\begin{figure}[h]
    \centering
\begin{subfigure}[hb]{0.23\textwidth}
\includegraphics[width=\linewidth,height=2cm]{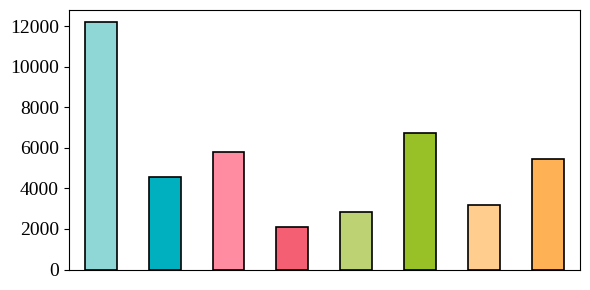}
\caption{Paragraph-Oriented}
\label{fig:sec_dist_para}
\end{subfigure}
\begin{subfigure}[hb]{0.23\textwidth}
\includegraphics[width=\linewidth,height=2cm]{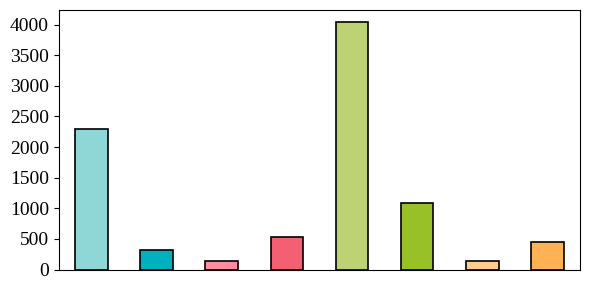}
\caption{Table-Oriented}
\label{fig:sec_dist_tabular}
\end{subfigure}
\begin{subfigure}[hb]{0.90\columnwidth}
\includegraphics[width=\linewidth,height=1cm]{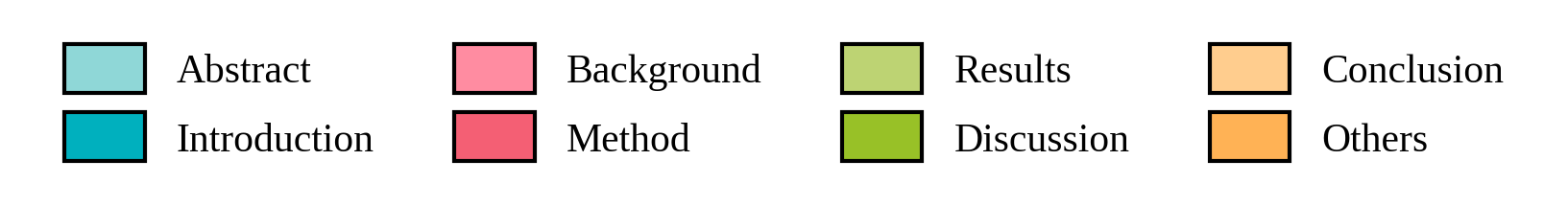}
\end{subfigure}
\caption{\textbf{Answer snippet distribution for each question types}; \textbf{Abstract} is the most frequent for Paragraph-Oriented whereas \textbf{Result} dominates Table-Oriented.}
\label{fig:Super-section-dist-grouped}
\end{figure}

DocHop-QA encodes structurally diverse and non-position-biased reasoning behaviors across its 11 question concepts. To support section level analysis, diverse section titles are mapped into 8 unified Super-Section tags described in Appendix~\ref{appendix:answer_dispersion}. As shown in Figure~\ref{fig:Super-section-dist-grouped}, Paragraph-Oriented types predominantly retrieve from Abstract while Table-Oriented types draw from Results and Discussion, consistent with their respective fan-hop and chain-hop strategies (see Figure~\ref{fig:super-section-dist-each-qtype} and Table~\ref{tab:table_section_count}). This section-level divergence is further supported by structural distance analysis in Figure~\ref{fig:Structural_distance_Analysis} and further explained in Appendix~\ref{appendix:answer_dispersion}. Fan-hop dispersion was measured by absolute positional distance and normalized distance ratio within documents, and chain-hop using both element-wise and their average using inter-snippet distances between consecutive steps. Paragraph-Oriented instances using fan-hop strategy  exhibit broadly distributed snippet distances without strong positional bias, while Table-Oriented instances of chain-hop strategy show a heavy-tailed distribution where most transitions are local but a non-negligible fraction spans long ranges, reflecting a mix of short and long-range dependencies. Further analysis of chain-hop transition in Figure~\ref{fig:chainhop_sankey} reveals that tables frequently serve not only as terminal evidence but also as intermediate reasoning bridges. These findings confirm that DocHop-QA systematically encodes diverse retrieval behaviors across document structure, posing genuine challenges for both evidence localization and multi-hop aggregation. 

\begin{figure}[!h]
    \centering
\begin{subfigure}[hb]{0.23\textwidth}
\includegraphics[width=\linewidth,height=2cm]{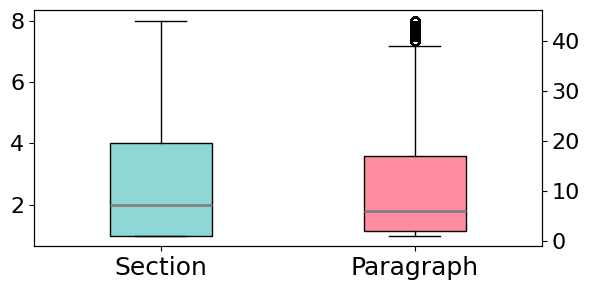}
\caption{Fan-hop (Absolute)}
\label{fig:fan-absolute-distance}
\end{subfigure}
\begin{subfigure}[hb]{0.23\textwidth}
\includegraphics[width=\linewidth,height=2cm]{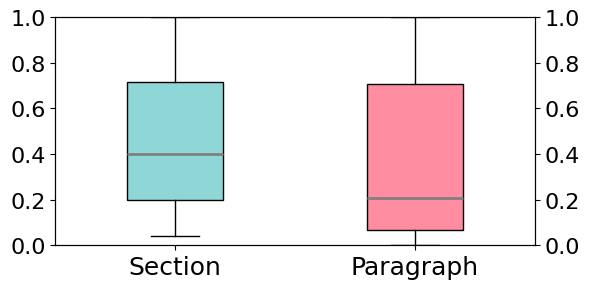}
\caption{Fan-hop (Ratio)}
\label{fig:fan-normalized-distance}
\end{subfigure}
\hfill
\begin{subfigure}[hb]{0.23\textwidth}
\includegraphics[width=\linewidth,height=2cm]{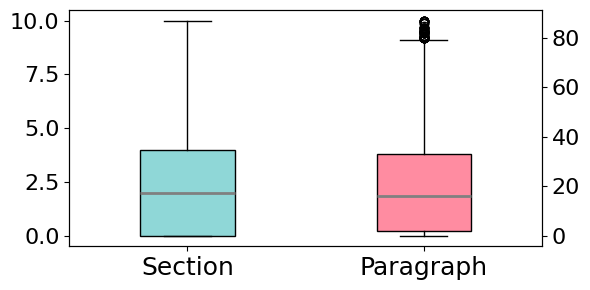}
\caption{Chain-hop (Element)}
\label{fig:chain-element_distance}
\end{subfigure}
\begin{subfigure}[hb]{0.23\textwidth}
\includegraphics[width=\linewidth,height=2cm]{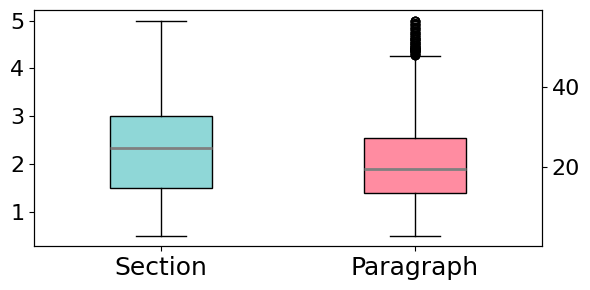}
\caption{Chain-hop (Average)}
\label{fig:chain-average_distance}
\end{subfigure}
\caption{\textbf{Section and paragraph level structural distance of fan-hop and chain-hop questions.}}
\label{fig:Structural_distance_Analysis}
\end{figure}

\vspace{-0.3em}    
\subsection{Reasoning Complexity} 
\label{sec:reasoning_complexity}
Table~\ref{tab:qa_datasets_comparison_recent} reports context source, answer form, hop count and reasoning depth across existing benchmarks, where reasoning depth is defined as the average number of decomposable sub-questions per instance. While DocHop-QA shows comparable hop counts to prior datasets (avg. 4.63 hops), the nature of evidence integration is fundamentally different. Although Loong~\cite{wang2024leave} exhibits higher hop counts and reasoning depth, its reasoning traversal relies on explicit citation links and predefined structural connections between documents, reducing the open-ended discovery burden on the model. Similarly, most existing benchmarks are text-centric, explicitly linked, and rely on short-span answers. In contrast, DocHop-QA draws from real scientific literature and requires synthesizing information across multiple paragraphs and tables, with cross-document reasoning constructed through semantic similarity rather than predefined structural connections. These properties establish DocHop-QA as a benchmark that demands richer semantic integration and higher reasoning complexity than hop statistics alone would suggest.

\renewcommand{\arraystretch}{1.3}
\begin{table}[!t]
    \centering
    \begin{adjustbox}{max width=\columnwidth}
    \renewcommand{\arraystretch}{1.2} 
    \setlength{\tabcolsep}{2pt}
    \begin{tabular}{l|c|c|c|c}
        \hline
        \textbf{Dataset} & \textbf{Context} & \textbf{Answer} & \textbf{\shortstack{Num \\ Hops}} &\textbf{\shortstack{Reasoning \\ Depth}} \\
        \hline
        HotpotQA & Wiki & short span and supporting facts & 2.0 & 1.42 \\
        HybridQA & Wiki & short span & 3.03 & 2.05 \\
        Loong & Industry & short spans or NL sentence & 10.18 & 3.25 \\
        FanOutQA & Wiki & collections of short span & 6.40 & 2.96 \\
        MEQA & Wiki & short span & 2.53 & 2.53 \\
        NovelHopQA & Gutenberg & 1 -2 sentences & 2.50 & 2.50 \\
        \hline
        \textbf{Ours} & \textbf{Pubmed} & \textbf{collection of paragraphs} & \textbf{4.63} & \textbf{2.46} \\
        \hline
    \end{tabular}
    \end{adjustbox}

    \caption{\textbf{Comparison of QA benchmark dataset} described in Table~\ref{tab:qa_datasets_comparison} in terms of context granularity, answer form, and reasoning complexity}
    \label{tab:qa_datasets_comparison_recent}
    \vspace{-1em}
\end{table}

\vspace{-0.3em}

\vspace{-0.1em}

\section{Experiments and Results}
\label{sec:experiments_and_results}
To evaluate the utility of DocHop-QA, we conduct experiments across four QA tasks covering different reasoning paradigms and input modalities, including generative and discriminative settings. These tasks assess models’ ability to reason over complex, document-rich inputs involving structured index prediction, free-form generation, and multimodal integration. Discriminative tasks involving entity index prediction, namely Task 3 (BBox Entity Index Extraction) and Task 4 (XML Entity Index Extraction) are discussed in detail in Appendix~\ref{appendix:detailed_experiments_and_results}, along with detailed experiment setups, metrics, and ablations. Section~\ref{sec:ablationstudies} and Appendix~\ref{appendix:ablationstudies} further discusses different ablation studies.

\begin{table}[!t]
\scriptsize
\centering
\setlength{\tabcolsep}{3pt}
\renewcommand{\arraystretch}{1.1}
\begin{tabularx}{\columnwidth}{l
    >{\centering\arraybackslash}X|
    *{5}{>{\centering\arraybackslash}X}
}
\hline
\textbf{Model} & \textbf{Input} & \textbf{BLEU} & \textbf{R-1} & \textbf{R-2} & \textbf{R-L} & \textbf{R-LS} \\
\hline
InternVL2   & T    & 14.53 & 32.11 & 11.99 & 19.56 & 20.91 \\
InternVL2   & I    & 6.39 & 28.74 & 6.99 & 16.12 & 18.50 \\
InternVL2  & T+I     &  7.48 & 31.23 &  8.55 & 16.46 & 20.26 \\
Qwen2.5-VL   & T    &  7.32 & 27.66 &  8.30 & 15.45 & 19.44 \\
Qwen2.5-VL   & I    &  12.40 & 34.48 &  11.49 & 19.50 & 22.53 \\
Qwen2.5-VL   & T+I     &  9.45 & 31.35 &  9.96 & 17.94 & 21.40 \\
Qwen3-VL 8B  & T & 26.00  &  --   &  --   & 28.80  &  --   \\
Qwen3-VL 8B  & I & 18.20  &  --   &  --   & 21.00  &  --   \\
Qwen3-VL 8B  & T+I & 29.70  &  --   &  --   & 32.10  &  --   \\
GPT-4o    & I     & 28.92 & \textbf{45.79} & \textbf{23.81} & 31.26 & \textbf{34.12} \\
Gemini 2.5  & T     & 11.94 & 24.55 & 12.85 & 16.81 & 19.41 \\
Gemini 2.5  & I     & 11.96 & 25.47 & 10.96 & 15.33 & 18.68 \\
Gemini 2.5   & T+I & 11.71 & 24.57 & 12.61 & 16.30 & 18.95 \\
Gemini 3 Pro  & T+I & \textbf{31.8}  &  --   &  --   & \textbf{34.5}  &  --   \\
\hline
\end{tabularx}
\caption{\textbf{Generative reasoning results under one-shot prompting across models and input configurations.} BLEU and ROUGE-1/2/L/LSum are reported. T:~Text, I: Image.}
\label{tbl:generative_oneshot}
\vspace{-2em}
\end{table}

\vspace{-0.7em}
\paragraph{Task 1: Generative Reasoning.}
This task tests models' ability to generate natural-language answers given different combinations of text and image inputs. We assessed three prompting styles (question-only, zero-shot, and one-shot) and three input modalities (text-only, image-only, text+image). We further compare different input modes with InternVL2-4B, Qwen2.5-VL-7B, and Qwen3-VL-8B for preprocessed inputs and GPT-4o, Gemini2.5 Flash, and Gemini3 Pro\footnote{\url{https://docs.cloud.google.com/gemini-enterprise-agent-platform/models/gemini/3-pro?hl=ko}} for full-document ingestion (XML, PDF). Evaluation is based on BLEU \cite{10.3115/1073083.1073135} and ROUGE \cite{lin-2004-rouge} scores against gold answers\footnote{As gold answers are synthesized rather than exhaustively enumerated, these lexical-overlap metrics do not assume a single canonical phrasing}. 
The generative reasoning task assesses the ability of large vision-language models to produce free-form answers from document-level context, without relying on predefined answer indices. The results of one-shot prompting setting is described in Table~\ref{tbl:generative_oneshot}. Gemini3 Pro demonstrates superior performance (BLEU: 31.8) with full-document ingestion, as opposed to input-processed architectures, where texts and images are incorporated into prompts, consuming valuable input context window. Detailed results are provided in Appendix~\ref{appendix:detailed_experiments_and_results}.

\begin{table}[t]
\scriptsize
\centering
\setlength{\tabcolsep}{3pt}
\renewcommand{\arraystretch}{1.1}
\begin{tabularx}{\columnwidth}{
    l
    c
    c|                     
    *{3}{>{\centering\arraybackslash}X}|  
    *{3}{>{\centering\arraybackslash}X}
}
\hline
\multicolumn{3}{c|}{} 
& \multicolumn{3}{c|}{\textbf{Sample}}   
& \multicolumn{3}{c}{\textbf{Accuracy}} \\
\textbf{Model} & \textbf{S} & \textbf{Img}
& \textbf{F1} & \textbf{Re} & \textbf{Pr}
& \textbf{Be} & \textbf{Co} & \textbf{Ov} \\
\hline
InternVL2 & Z & -- & 7.03 & 9.77 & 7.39 & 0.88 & 0.88 & 28.30 \\
InternVL2 & Z & \checkmark & 4.24 & 6.72 & 4.03 & 0.35 & 0.75 & 17.44 \\
InternVL2 & O & \checkmark & 4.34 & 5.77 & 4.68 & 0.44 & 0.62 & 17.31 \\
\hline
Qwen2.5-VL & Z & -- & 14.06 & 51.62 & 13.69 & 4.04 & 34.05 & 75.48 \\
Qwen2.5-VL & Z & \checkmark & 7.92 & 58.53 & 7.06 & 1.54 & 50.04 & 69.73 \\
Qwen2.5-VL & O & \checkmark & 8.12 & 18.55 & 7.41 & 0.04 & 9.89 & 41.08 \\
\hline
GPT-4o & Z & - & \textbf{16.21} & 54.12 & 15.85 & 5.18 & 36.86 & \textbf{79.38} \\
GPT-4o & Z & \checkmark & 9.37 & 60.27 & 8.91 & 2.12 & 52.46 & 73.15 \\
GPT-4o & O & \checkmark & 9.15 & 21.43 & 8.75 & 0.97 & 12.21 & 45.62 \\

\hline
\end{tabularx}
\caption{
\textbf{Structured Generative Answering results.}
S:~Setup (Z: Zero-shot, O: One-shot),
Img: Image input. 
}
\vspace{-1.2em}
\label{tbl:structuredgenerativeresults}
\end{table}

\begin{figure}[t]
\centering
\includegraphics[width=\columnwidth]{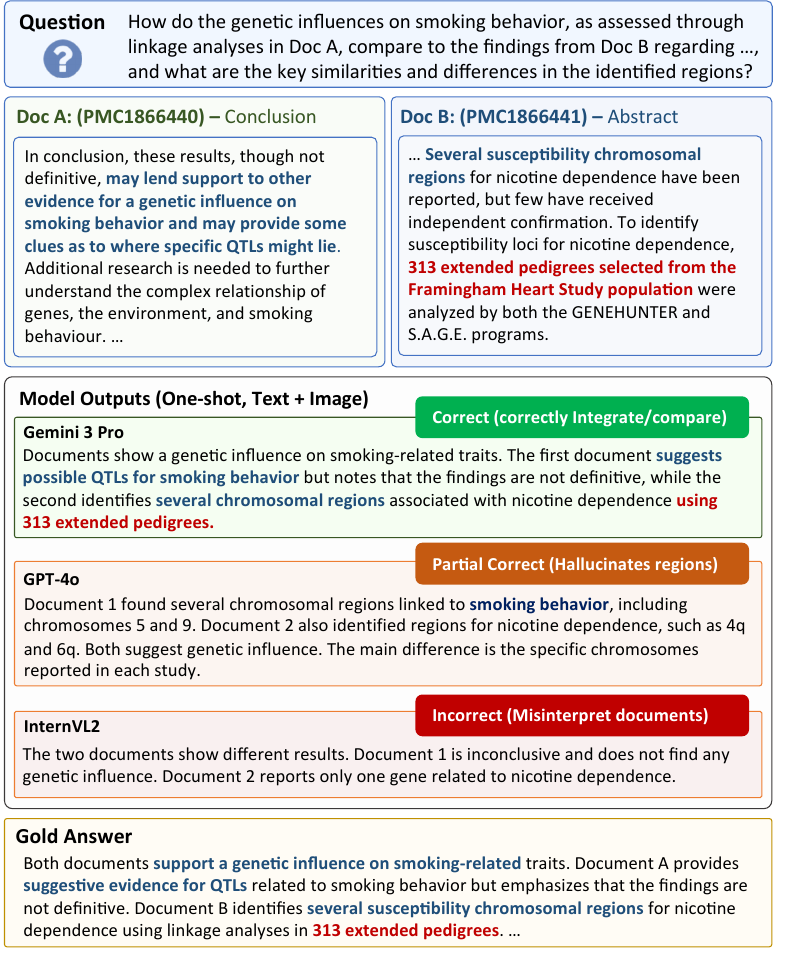}
\caption{Representative qualitative sample for cross-document evidence integration.}
\vspace{-0.3em}
\label{fig:qualitative_example}
\vspace{-1em}
\end{figure}

\paragraph{Task 2: Structured Generative Answering.}
For this task, models generate lists of answer entity indices in response to questions, using labeled XML content and, optionally, associated page images. We compared InternVL2-4B \cite{chen2024internvl}, Qwen2.5-VL-7B \cite{Qwen2.5-VL}, and GPT-4o in zero-shot and one-shot settings. Evaluation spans multi-label spatial accuracy (belong/overlap/contain)\footnote{\url{https://meta-pytorch.org/torcheval/stable/generated/torcheval.metrics.MultilabelAccuracy.html}} and sample-level precision, recall, and F1.
This task assesses whether generative models can produce structured answer indices from labeled contextual inputs. GPT-4o achieves the highest F1 score (16.21) followed by Qwen2.5-VL (14.06) in a zero-shot, text-only setting (Table \ref{tbl:structuredgenerativeresults}), showing strong performance without fine-tuning.
InternVL2 performs poorly across sets and struggles with consistent output formatting, underscoring the need for task-specific pretraining. Adding document images generally hurts performance due to alignment challenges between text and visual layouts.
The structured generative task highlights both the promise and limitations of using instruction-following LLMs for entity-index generation. DocHop-QA provides a challenging testbed for structured reasoning in generative settings, enabling deeper studies of prompt design, modality alignment, and decoding strategies.
\vspace{-0.3em}
\paragraph{Qualitative Analysis: Cross-Document Evidence Integration.}
We present a representative qualitative example to illustrate the reasoning challenges in DocHop-QA.
Figure~\ref{fig:qualitative_example} shows an example from Task~1 (Generative Reasoning), where answering the question requires integrating complementary findings across two independent documents.
Document~A reports tentative linkage-analysis evidence suggesting possible QTLs associated with smoking behavior, while Document~B identifies susceptibility chromosomal regions for nicotine dependence using extended pedigree analysis.
Correct answering requires models to synthesize the shared conclusion that smoking-related traits have genetic influences while distinguishing differences in confidence, phenotype focus, and specificity of the reported loci. Stronger models such as Gemini~3 Pro successfully aggregate evidence across documents and produce coherent comparative summaries.
However, even strong models such as GPT-4o may hallucinate unsupported chromosomal regions not explicitly stated in the evidence.
Weaker models frequently fail to integrate the two studies correctly and instead misinterpret the relationship between smoking behavior and nicotine dependence.
This example demonstrates that DocHop-QA requires more than local evidence retrieval: models must perform cross-document reasoning, comparative interpretation, and coherent answer synthesis from distributed scientific evidence.

\section{Ablation Studies}
\label{sec:ablationstudies}

\paragraph{Modality Contribution Analysis}
To examine the necessity of multimodal reasoning, we conduct a controlled ablation using Qwen3-VL-8B under four input configurations.
As in Table~\ref{tab:ablation_modality}, removing visual input leads to a measurable performance drop relative to the full multimodal setting, while image-only input is substantially weaker, confirming that textual content remains the dominant modality.
Joint multimodal input consistently outperforms text-only, indicating that visual cues contribute non-trivially to answer synthesis.
The oracle-text upper bound further reveals that evidence integration itself remains a key bottleneck, even when relevant snippets are directly provided.

\begin{table}[h]
\centering

\resizebox{\columnwidth}{!}{%
\begin{tabular}{llcc}
\toprule
\textbf{Setting} & \textbf{Input} & \textbf{BLEU} & \textbf{ROUGE-L} \\
\midrule
Full (Text+Image) & OCR text + page images & 29.7 & 32.1 \\
Text-only         & OCR text only          & 26.0 & 28.8 \\
Image-only        & Page images only       & 18.2 & 21.0 \\
Oracle-text       & Gold snippets only     & 35.4 & 38.2 \\
\bottomrule
\end{tabular}
}
\caption{Modality ablation on DocHop-QA using Qwen3-VL-8B.}
\label{tab:ablation_modality}
\vspace{-0.8em}
\end{table}

\paragraph{Shortcut Learning Analysis}
To investigate whether models can exploit parametric knowledge or generic scientific phrasing rather than performing genuine document-grounded reasoning, we evaluate Qwen3-VL-8B in two settings: with full document access and with question input alone.
As reported in Table~\ref{tab:ablation_shortcut}, question-only performance is extremely low (BLEU 6.1, ROUGE-L 9.6), confirming that answers cannot be derived from prior knowledge alone.
Combined with the document-grounded, multi-hop template constraints used during question generation, these results indicate that DocHop-QA requires genuine evidence aggregation across multiple documents rather than superficial semantic matching.

\begin{table}[h]
\centering
\small

\begin{tabular}{lcc}
\toprule
\textbf{Setting} & \textbf{BLEU} & \textbf{ROUGE-L} \\
\midrule
With Documents & 29.7 & 32.1 \\
Question Only  &  6.1 &  9.6 \\
\bottomrule
\end{tabular}
\caption{Shortcut learning analysis on DocHop-QA}
\label{tab:ablation_shortcut}
\vspace{-1em}
\end{table}

\begin{table}
\centering
\footnotesize
\begin{tabular}{lc}
\toprule
\textbf{Setting} & \textbf{ROUGE-L} \\
\midrule
    Question-only & 10.2 \\
    Retrieved (GTR top-5) & 28.2 \\
    Oracle (gold snippets only) & 48.5 \\
    Human (full document) & 58.6 \\
\bottomrule
\end{tabular}
\caption{Controlled evidence-access performance analysis comparing input setups question-only, GTR retrieved context access, and oracle access to human performance with full document access.}
\label{tab:controlledevidence}
\end{table}

\paragraph{Controlled Evidence-Access Analysis} We evaluate Qwen3-VL-8B (one-shot) under different evidence-access settings on the same 50-instance subset used for human evaluation to further illustrate the task difficulty (Table~\ref{tab:controlledevidence}). We find that question-only performance is very low (10.2), indicating that answers cannot be generated from parametric knowledge or general scientific phrasing alone. Oracle, having access to the gold snippets, improves over Retrieved, suggesting that retrieval contributes meaningfully to overall task difficulty. However, even with gold evidence provided, model performance (48.5) remains below human performance (58.6). This suggests that integrating evidence across documents and synthesizing answers remains challenging, even when correct snippets are available. The consistent performance gaps across conditions indicate that DocHop-QA requires non-trivial evidence aggregation rather than superficial summarization or localization. These results provide empirical evidence that DocHop-QA questions are evidence-dependent and reflects reasoning complexity.





\paragraph{Retrieval Quality and Multi-Hop Recall}
Table~\ref{tab:ablation_retrieval} reports snippet-level and complete multi-hop recall for three retrieval systems.
While Top-5 snippet recall is relatively high across all retrievers, complete multi-hop recall, which requires all supporting evidence pieces to be jointly retrieved, remains substantially lower (e.g., 0.39 at Top-5 for GTR-large).
This recall gap directly reflects the core fan-hop aggregation challenge in DocHop-QA: retrieving any single relevant snippet is comparatively tractable, whereas locating the full set of evidence required for a multi-hop answer remains an open challenge.

\begin{table}[h]
\centering
\resizebox{\columnwidth}{!}{%
\begin{tabular}{lcccc}
\toprule
\textbf{Retriever} & \textbf{Top-5 R} & \textbf{Top-10 R} & \textbf{Comp. R@5} & \textbf{Comp. R@10} \\
\midrule
TF-IDF + BERT & 0.68 & 0.77 & 0.32 & 0.41 \\
BM25          & 0.71 & 0.79 & 0.35 & 0.42 \\
GTR-large     & 0.76 & 0.84 & 0.39 & 0.48 \\
\bottomrule
\end{tabular}%
}
\caption{Retrieval quality analysis on DocHop-QA, evaluated with Recall (Top-K and \textbf{Comp}lete Recall)}
\label{tab:ablation_retrieval}
\end{table}

\vspace{-1.0em}

\section{Conclusion}
\vspace{-0.5em}
We presented DocHop-QA, a new benchmark designed to evaluate multi-hop reasoning over multimodal, multi-document scientific corpora. Unlike prior datasets, DocHop-QA captures realistic information seeking scenarios by integrating unstructured text, structured tables, and visual layout features, without relying on explicit links.
Using 11 concepts, we generated diverse QAs and evaluated performance on 4 representative tasks. Results highlight critical insights into model behavior, modality integration challenges, and prompt effects, underscoring the need for further research in multi-hop, multimodal reasoning. We hope DocHop-QA facilitates future research toward more robust QA systems for real world documents.

\section*{Limitations}
DocHop-QA has a few limitations worth noting. First, parts of the dataset are generated using LLM-assisted pipelines, which may introduce regularities in question formulation despite concept-guided prompting and human validation. Second, we evaluate end-to-end reasoning over long, multi-document contexts without requiring access to explicit intermediate-reasoning annotations. Third, while table-oriented questions reflect common scientific reading behavior, they primarily emphasize high-level semantic interpretation rather than fine-grained numerical or cell-level operations; consequently, complex table structures, such as merged cells and hierarchical headers, and scientific diagram understanding are not explicitly targeted. Finally, our evaluation targets general-purpose models in zero-shot or light prompting settings, leaving task-specific adaptation for future work.

\section*{Ethics Statement}
DocHop-QA is constructed from publicly available scientific articles in the PubMed Central Open Access corpus. The dataset does not contain personal, sensitive, or private information, and all source documents are used in accordance with their original licenses.
The benchmark is designed for research purposes to study multimodal, multi-document reasoning and is not intended for clinical decision-making or medical advice. While the dataset is automatically generated with assistance from large language models, we apply concept-guided generation, filtering, and human validation to mitigate hallucinations and low-quality instances.

\section*{Acknowledgments} 
We thank Prof. Minseok Song for his support of this research. This research was supported by the Korea Planning \& Evaluation Institute of Industrial Technology (KEIT) funded by the Ministry of Trade, Industry and Energy (No.RS-2025-25458052, Development of Core Technologies for Manufacturing Foundation Models) and the Institute of Information \& communications Technology Planning \& Evaluation (IITP) grant funded by the Korea government(MSIT) (No.RS-2025-02217259, Development of self-evolving AI bias detection-correctionexplain platform based on international multidisciplinary governance), (RS-2024-00395401, Development of VFX creation and combination using generative AI), ((IITP-2026-RS-2024-00441244) and OTT (RS-2024-00353131), Global Data-X Leader HRD program).


\bibliography{custom}
\clearpage

\appendix
\onecolumn

\section{Table of Contents}
\begin{table}[!h]
\centering
\small
\setlength{\tabcolsep}{5pt}
\renewcommand{\arraystretch}{1.1}
\begin{tabularx}{\textwidth}{
  >{\raggedright\arraybackslash}p{2.5cm}
  >{\raggedright\arraybackslash}p{3cm}
  >{\centering\arraybackslash}p{2.5cm}
  >{\raggedright\arraybackslash}X
}
\toprule
\textbf{Category} & \textbf{Subcategory} & \textbf{Location} & \textbf{Key Items (Purpose)} \\
\midrule

& \multicolumn{3}{c}{\normalsize{\textit{\textbf{How to prove the dataset construction pipeline?}}}} \\[-2pt]
\cmidrule(l){2-4}\cmidrule(l){2-4}\\[-12pt]
\multirow{4}{2.3cm}{\normalsize{\textbf{Pipeline \& Construction}}}
& Question Concept Proposal
& Sec.~\ref{sec:question_concept_proposal}
& Table~\ref{tab:concept_type_examples} (Question Concept Type Examples) \\
\cmidrule(l){2-4}
& Document Pairing Strategy
& Sec.~\ref{sec:document_selection}, App.~\ref{appendix:document_selection}
& Alg.~\ref{algorithm:document_selection_pseudocode} (Document Selection Algorithm) \\
\cmidrule(l){2-4}
& Question Generation
& Sec.~\ref{sec:qa_generation}, App.~\ref{appendix:traindatasetexample}
& Fig.~\ref{fig:pipeline} (Pipeline Overview) \newline
  Tables~\ref{tab:example-comparison}--\ref{tab:example-pst} (Question Generation Prompt) \\
\cmidrule(l){2-4}
& Answer Matching
& Sec.~\ref{sec:answer_matching}, App.~\ref{appendix:detailed_answer_matching}
& Table~\ref{tab:keyword_lists} (Filtering Keyword List) \\
\midrule

& \multicolumn{3}{c}{\normalsize{\textit{\textbf{Is this dataset valid?}}}} \\[-2pt]
\cmidrule(l){2-4}\cmidrule(l){2-4}\\[-12pt]
\multirow{1}{2.3cm}{\normalsize{\textbf{Quality \& Validation}}}
& Quality Assurance
& Sec.~\ref{sec:dataset_evaluation}, App.~\ref{appendix:quality-assurance},~\ref{appendix:supplementary}
& Tables~\ref{tab:gwet_ac2_summary},~\ref{tab:human_llm_agreement} (Inter-annotator Agreement Result) \newline
  Figs.~\ref{fig:llm _result}--\ref{fig:llm_based_quality_assurance_result} (LLM-as-judge Result)\newline
  Tables ~\ref{tab:eval-prompt},~\ref{tab:eval-fewshot} (LLM Evaluation Prompt)\newline
  Fig. ~\ref{fig:human_evaluation_form_all} (Human Evaluation Interface)\\
\midrule

& \multicolumn{3}{c}{\normalsize{\textit{\textbf{What does the dataset cover?}}}} \\[-2pt]
\cmidrule(l){2-4}\cmidrule(l){2-4}\\[-12pt]
\multirow{2}{2.3cm}{\normalsize{\textbf{Diversity \& Coverage}}}
& Dataset Coverage
& Sec.~\ref{sec:document_complexity}, App.~\ref{appendix:document_complexity}
&  Fig.~\ref{fig:overview_of_q_types_and_document_usage} (Dataset Statistics) \newline Fig.~\ref{fig:Document analysis} (Document Analysis) \newline
  Figs.~\ref{fig:wordcloud-ps}--\ref{fig:wordcloud-ts} (Question Keyword) \\
\cmidrule(l){2-4}
& Answer Dispersion \& Reasoning Scope
& Sec.~\ref{sec:answer_dispersion}, App.~\ref{appendix:answer_dispersion}
& Table~\ref{tab:super_section_details}, Fig.~\ref{fig:section_mapping_chart} (Super Section Mapping Result) \newline
  Figs.~\ref{fig:Super-section-dist-grouped},~\ref{fig:super-section-dist-each-qtype}, Table~\ref{tab:table_section_count} (Section-level Distribution) \newline
  Figs.~\ref{fig:Structural_distance_Analysis},~\ref{fig:distance_by_qtypes} (Answer Snippet Distance) \newline
  Fig.~\ref{fig:chainhop_sankey} (Chain-hop Transition) \\
\midrule

& \multicolumn{3}{c}{\normalsize{\textit{\textbf{Why is this dataset hard?}}}} \\[-2pt]
\cmidrule(l){2-4}\cmidrule(l){2-4}\\[-12pt]
\normalsize{\textbf{Reasoning Analysis}}
& Reasoning Complexity
& Sec.~\ref{sec:reasoning_complexity}
& Table~\ref{tab:qa_datasets_comparison_recent} (Reasoning Depth Comparison) \\
\midrule

& \multicolumn{3}{c}{\normalsize{\textit{\textbf{What actually makes the task difficult?}}}} \\[-2pt]
\cmidrule(l){2-4}\cmidrule(l){2-4}\\[-12pt]
\multirow{4}{2.3cm}{\normalsize{\textbf{Controlled Experiment}}}
& Task 1: Generative Reasoning
& Sec.~\ref{sec:experiments_and_results}, App.~\ref{appendix:detailed_experiments_and_results}
& Table~\ref{tbl:generative_oneshot} (Main Result) \newline
  Table~\ref{tbl:generativeresults} (Full Result) \newline
  Fig.~\ref{fig:qwen_tsne} (Question Type Embedding Analysis) \\
\cmidrule(l){2-4}
& Task2: Structured Generative Answering
& Sec.~\ref{sec:experiments_and_results}, App.~\ref{appendix:detailed_experiments_and_results}
& Table~\ref{tbl:structuredgenerativeresults} (Main Result) \\
\cmidrule(l){2-4}
& Task3, 4: BBox \& XML Entity Index Extraction
& App.~\ref{appendix:detailed_experiments_and_results}
& Fig. ~\ref{fig:task3_entitycount_f1} (Result Comparisons) \newline
Tables~\ref{tbl:bboxresults},~\ref{tbl:xmlresults} (Main Result) \\
\cmidrule(l){2-4}
& Ablation Analysis
& Sec.~\ref{sec:ablationstudies}, App.~\ref{appendix:ablationstudies}
& Table ~\ref{tab:ablation_modality} (Modality Ablation)\newline 
Table ~\ref{tab:ablation_shortcut} (Shortcut Ablation)\newline
Table ~\ref{tab:controlledevidence} (Controlled Evidence Ablation)\newline 
Table~\ref{tab:human_answerability} (Human Answerability and Realism Analysis)\newline
Table ~\ref{tab:ablation_retrieval} (Retrieval Ablation) 
\\
\midrule

& \multicolumn{3}{c}{\normalsize{\textit{\textbf{Can this work be reliably reproduced and extended?}}}} \\[-2pt]
\cmidrule(l){2-4}\cmidrule(l){2-4}\\[-12pt]
\multirow{4}{2.5cm}{\normalsize{\textbf{Reproducibility \& Implementation}}}
& Dataset and Code
& App.~\ref{appendix:data_and_code_release}
& \url{https://shorturl.at/rqfcM} \\
\cmidrule(l){2-4}
& License of Artifacts & App.~\ref{appendix:data_and_code_release} & -- \\
\cmidrule(l){2-4}
& Reproducibility Cost
& App.~\ref{computation_efficiency}
& Table ~\ref{tab:qa_generation_cost} (Computational Cost) \\
\cmidrule(l){2-4}
& Implementation Details
& App.~\ref{appendix:training_setup}, ~\ref{appendix:hyperparams}
& Tables~\ref{tbl:hyperparametersgenerative}--~\ref{tbl:hyperparametersxml} (Hyperparameters) \newline  \\

\bottomrule
\end{tabularx}
\caption{
  Structural overview of DocHop-QA.
}
\label{tab:toc}
\end{table}

\clearpage

\twocolumn
\section{Data and Code Release}
\label{appendix:data_and_code_release}

\paragraph{Link to Code and Dataset}
Full implementation and reproduction scripts for Dataset Construction (Section~\ref{sec:dataset construction}), Dataset Quality Assurance (Section~\ref{sec:dataset_evaluation}), and Experiments and Results (Section~\ref{sec:experiments_and_results}) are at the following link. Access it here:
\url{https://shorturl.at/rqfcM}
\paragraph{License for Artifacts}
All experiments are conducted using publicly available datasets and pretrained multimodal foundation models subject to their respective licenses and terms of use.
We release the DocHop-QA dataset and the code necessary to reproduce the overall experimental procedures.
Users are responsible for ensuring compliance with the licenses and usage restrictions associated with the underlying datasets and pretrained backbones.

\section{Related Work}
\label{appendix:related_works}
\paragraph{Table-Oriented QA}
Several prior works explore table-centric QA, but they differ fundamentally from our setting. \citet{pal-etal-2024-table} generate table QA for low-resource languages, yet their method relies on structured web pages and does not evaluate multi-hop or path-level reasoning. Likewise, FORTAP\cite{cheng2022fortapusingformulasnumericalreasoningaware} focuses on numerical reasoning over tabular data, but the framework lacks support for textual inference or multimodal evidence integration. More recently, \citet{wu2025mmqa} propose MMQA, which evaluates multi-hop reasoning at the schema level over relational tables, rather than document-centric evidence integration. Our DocHop-QA targets document-centric multi-hop reasoning, where evidence spans heterogeneous elements such as paragraphs, tables, and visual layouts across multiple documents.

\section{Dataset Construction Details}
\label{appendix:detailed_dataset_construction}
\subsection{Document Selection Details}
\label{appendix:document_selection}
As mentioned in Section 3.2, we formalized the semantic similarity-based document pairing strategy into a concrete algorithmic procedure, as illustrated in Algorithm~\ref{algorithm:document_selection_pseudocode}. Candidate pairs are first identified using the following two filters: (1) at least one overlapping keyword, and (2) a TF-IDF cosine similarity score of at least 0.3\footnote{After testing multiple thresholds (0.1--0.9) and manually inspecting retrieved documents, 0.3 was selected for consistently yielding the most relevant results.}. After this first-stage filtering, we compute a total matching score for each remaining pair as a weighted combination of normalized keyword overlap and TF-IDF similarity. For each source document, we then kept only the top three scoring partners for downstream question generation, limiting repeated reuse of the same documents and promoting document diversity.

\begin{algorithm}[!h]
\small
\caption{\textbf{Document Selection}}
\label{algorithm:document_selection_pseudocode}
\textbf{Input}: Source corpus $\mathcal{C}=\{d_1,\dots,d_N\}$ (PubMed XMLs)\\
\textbf{Parameter}: Keyword extractor (YAKE) $\Phi$ ; TF--IDF vectorizer $\psi$ ; 
similarity threshold $\tau$ ; weight $\alpha$ ; top-$L$ \\
\textbf{Output}: Paired set $\mathcal{P} \subseteq \{(i,j)\mid 1\le i<j\le N\}$\\

\begin{algorithmic}[1] 
\STATE $\mathcal{P} \gets \varnothing$; initialize $S_{ij}\gets -\infty$ for all $i<j$
\FOR{$i \gets 1$ \textbf{to} $N$}
  \STATE extract $\textit{pmc id}_i$, $\textit{title}_i$, $\textit{abstract}_i$ from $d_i$
  \STATE $\textit{fulltext}_i \gets \textit{title}_i \parallel \textit{abstract}_i$
  \STATE $\mathcal{K}_i \gets \Phi(\textit{fulltext}_i)$; remove stopwords
\ENDFOR $\mathbf{X} \gets \psi (\textit{ fulltext}_1,\dots,\textit{fulltext}_N )$ 
\STATE $\mathbf{M} \gets \mathrm{cosine}(\mathbf{X}\mathbf{X}^\top)$

\FOR{$i \gets 1$ \textbf{to} $N$}
  \FOR{$j \gets i+1$ \textbf{to} $N$}
    \STATE $C_{ij} \gets \mathcal{K}_i \cap \mathcal{K}_j$ 
    \IF{$|C_{ij}|>0$ \textbf{and} $\mathbf{M}_{ij}\ge \tau$}
      \STATE $s_{\mathrm{kw}} \gets \dfrac{\log(1+|C_{ij}|)}{\log(1+\max_k|\mathcal{K}_k|)}$
      \STATE $s_{\mathrm{sim}} \gets \mathbf{M}_{ij}$
      \STATE $S_{ij} \gets \alpha\, s_{\mathrm{kw}} + (1-\alpha)\, s_{\mathrm{sim}}$ 
    \ENDIF 
\ENDFOR $\mathcal{N}_i \gets \text{Top-}L \text{ indices } j \text{ by } S_{ij}$
  \FORALL{$j \in \mathcal{N}_i$}
    \STATE $\mathcal{P} \gets \mathcal{P} \cup \{(i,j)\}$
  \ENDFOR
\ENDFOR
\textbf{return} $\mathcal{P}$
\end{algorithmic}
\end{algorithm}

\subsection{Question Generation Details}
\label{appendix:traindatasetexample}
\paragraph{Instruction Tuning}
We presented few shot instances for three types of question generation: Non-Comparison, Comparison, and Non-Refer-Table (PST). Each instance follows a unified schema comprising a prompt, an instruction, two abstracts, and the resulting question. The prompt encapsulates the underlying question concept, the instruction clarifies how the concept should be instantiated, and Contexts 1 and 2 provide the input abstracts. Although the schema is identical across types, we train a distinct model for each category because our preliminary experiments revealed that type-specific models generate a higher proportion of valid questions than a single unified model. Representative examples of each question type are shown in Tables~\ref{tab:example-comparison}, \ref{tab:example-noncomparison}, and~\ref{tab:example-pst}.

\begin{table}[!h]
\small
\centering
\begin{tabular}{p{0.95\linewidth}}
\toprule

\textbf{\textit{Prompt}} \\
Instruction: How does [ ] compare to [ ], and what are the key similarities and differences?\\
Chain-of-Thought (internal use only): Generate a question by replacing only the placeholder [ ] in the question concept below using document-specific information derived from the provided contexts. You must use two documents.\\

\vspace{0.2em}

\textbf{\textit{Instruction}} \\
How does [ ] compare to [ ], and what are the key similarities and differences? \\

\vspace{0.2em}

\textbf{\textit{Input}} : \{\\ 
\hspace{0.5em}\textbf{\textit{Context 1}} \\
\hspace{0.5em}\texttt{Regionalization of pediatric emergency care in the US. The 2006 IOM Report highlighted uneven care. Pediatric emergency services are concentrated in a limited number of children’s hospitals and trauma centers.} \\

\vspace{0.2em}

\hspace{0.5em}\textbf{\textit{Context 2}} \\
\hspace{0.5em}\texttt{Current state of pediatric emergency care in Korea. A 2010 survey showed widespread unpreparedness: many EDs lacked essential pediatric equipment, and consultations were often handled by inexperienced residents.} \} \\

\vspace{0.2em}

\textbf{\textit{Output}} \\
How do the characteristics of the healthcare systems in the United States compare to Korea, and what are the key similarities and differences? \\

\bottomrule
\caption{\textbf{An example of our dataset instance based on comparison-type multi-hop question generation.} Fields such as prompt, instruction, and context are shown in full.}
\label{tab:example-comparison}\end{tabular}
\end{table}

\begin{table}[!h]
\small
\centering
\begin{tabular}{p{0.95\linewidth}}
\toprule

\textbf{\textit{Prompt}} \\
Instruction: What are the main challenges in [ ], and how can [ ] address these issues?\\
Generate a question by replacing only the placeholder [ ] in the question concept below using document-specific information derived from the provided contexts. \\Chain-of-Thought (internal use only): Generate a question by replacing only the placeholder [ ] in the question concept below using document-specific information derived from the provided contexts. You must use two documents.\\

\vspace{0.2em}

\textbf{\textit{Instruction}} \\
What are the main challenges in [ ], and how can [ ] address these issues? \\

\vspace{0.2em}

\textbf{\textit{Input}} : \{\\ 

\hspace{0.5em}\textbf{\textit{Context 1}} \\
\hspace{0.5em}\texttt{Anaemia is a common haematological disorder in chronic disease, often complicating diagnosis and worsening quality of life and survival. This paper reviews its pathogenesis and potential therapeutic targets.} \\

\vspace{0.2em}

\hspace{0.5em}\textbf{\textit{Context 2}} \\
\hspace{0.5em}\texttt{Cloning the human EPO gene (1984) enabled ESA therapies for cancer-related anaemia. Trials show ESAs reduce transfusion needs, provide sustained correction, lower transfusion-related risks, and improve blood flow and quality of life.} 
\} \\

\vspace{0.2em}

\textbf{\textit{Output}} \\
What are the main challenges in iron availability for red cell precursors in cancer-related anaemia, and how can erythropoiesis-stimulating agents address this issue? \\

\bottomrule
\caption{\textbf{An example of our dataset instance based on Non-comparison type multi-hop question generation.} Fields such as prompt, instruction, and context are shown in full.}
\label{tab:example-noncomparison}
\end{tabular}
\end{table}

\begin{table}[!h]
\small
\centering
\begin{tabular}{p{0.95\linewidth}}
\toprule

\textbf{\textit{Prompt}} \\
Instruction: What are the main challenges in [ ], what strategies can be implemented to address them? Is there any table additionally describe it? Which table is it and what is the main point of the table?\\
Chain-of-Thought (internal use only): Generate a question by replacing only the placeholder [ ] in the question concept below using document-specific information derived from the provided contexts. You must use two documents.\\

\vspace{0.2em}

\textbf{\textit{Instruction}} \\
What are the main challenges in [ ], what strategies can be implemented to address them? Is there any table additionally describe it? Which table is it and what is the main point of the table? \\

\vspace{0.2em}

\textbf{\textit{Input}} : \{\\

\hspace{0.5em}\textbf{\textit{Context 1}} \\
\hspace{0.5em}\texttt{Inflammatory diseases, including rheumatic disorders, are a major global health burden. While multiple treatments exist, long-term steroidal or non-steroidal therapy can cause serious adverse effects.} \\

\vspace{0.2em}

\hspace{0.5em}\textbf{\textit{Context 2}} \\
\hspace{0.5em}\texttt{A randomized, double-blind, placebo-controlled trial enrolled 80 UC patients in remission (Sep 2009--Oct 2010). Participants received SM (140 mg/day) or placebo alongside standard therapy for six months.} \}  \\

\vspace{0.2em}

\textbf{\textit{Output}} \\
What are the main challenges in traditional anti-inflammatory in treating inflammatory diseases, what strategies can be implemented to address them? Is there any table additionally describe it? Which table is it and what is the main point of the table?  \\

\bottomrule
\caption{\textbf{An example of our dataset instance based on PST type multi-hop question generation.} Fields such as prompt, instruction, and context are shown in full.}
\label{tab:example-pst}\end{tabular}
\end{table}

\paragraph{Question Post-processing}
\label{appendix:postprocessing}
To ensure that generated questions strictly follow the intended question concepts and remain usable for downstream QA, we apply a lightweight post-processing step to filter invalid generations.
We use the following criteria: (1) Generation failures:
We remove instances where the model produces empty outputs or outputs that do not conform to the expected format (e.g., malformed JSON fields or missing required components). (2) Question--concept mismatch: We filter out questions that fail to instantiate the target concept, operationalized by missing the required concept keywords/placeholders specified in the corresponding prompt. This step prevents concept drift and enforces consistency between the intended reasoning structure and the generated question.
After applying these filters, we retain only question--context pairs that pass all checks, forming the final dataset used in our experiments.

\subsection{Answer Matching Details}
\label{appendix:detailed_answer_matching}
In DocHop-QA, multi-hop reasoning follows two paradigms. Fan-hop aggregates semantically complementary information from multiple, independently retrieved documents. Each snippet contributes to the final answer without explicit cross-referencing. On the other hand, Chain-hop performs step-by-step inference where information in one document directly supports content in another, often forming a logical link-such as when a table summarizes results from an external study. 

\paragraph{Paragraph-Oriented Type}
To ground each QA instance with relevant context, we parsed the full-text files of both documents. The subsection-level content was extracted, and segments with fewer than two sentences were discarded to avoid sparse inputs.
For each segment, we computed semantic similarity to the question using a hybrid scoring method: a weighted combination of TF-IDF (weight 0.2) and BERT-based\cite{DBLP:journals/corr/abs-1904-09675} sentence embeddings (weight 0.8). This configuration prioritized deeper semantic understanding while preserving term-level sensitivity. 
Similarity scores were aggregated across both documents, and the top 5 segments were retrieved. To ensure quality, only answer instances with all retrieved snippets above a similarity score of 0.4 were retained. This threshold was empirically selected based on preliminary validation, which showed that higher similarity was strongly associated with answer plausibility. 
All filtered QA-context pairs were saved in structured JSON.

\paragraph{Table-Oriented Type}
\label{appendix:keywords}
We employed two strategies depending on the snippet’s role: some snippets were retrieved via semantic-similarity scoring augmented with keyword-based filtering, whereas others were extracted using predefined logic. To support the semantic-similarity route, we curated three keyword lists targeting problem-, solution-, and limitation-related paragraphs (Table~\ref{tab:keyword_lists}); these lists served as an initial paragraph filter in each context and were then combined with semantic-similarity scoring to select answer snippets. If no paragraphs are filtered we computed similarity scores between questions and all paragraphs.
We designed logical answer chains tailored to each QA type (PST Type and TS/TSL Type), which are explained as follows. Green-colored snippets used the semantic similarity strategy, while others were retrieved via predefined logic.

\begin{table}[!h]
\centering
\scriptsize
\setlength{\tabcolsep}{4pt}

\begin{tabularx}{\columnwidth}{p{2.3cm} X}
\toprule
\textbf{Category} & \textbf{Keywords} \\
\midrule

Problem-related &
burden, challenge, gap, unmet need, lack of, insufficient,
controversy, uncertainty, poor outcomes, high prevalence,
incidence, rising rates, increasing trend, public health concern,
problem, barrier, limitation, disparity, inequity,
inaccessibility, misdiagnosis, delay, complication,
risk factor, vulnerability, deficiency, inconsistency,
overuse, underuse, shortage, low adherence,
noncompliance, failure, inadequacy, ineffectiveness,
complexity, fragmentation, obstacle, hazard,
disease burden \\
\midrule

Solution-related &
approach, strategy, intervention, treatment, therapy,
model, framework, tool, instrument, protocol,
program, solution, innovation, novel, effective,
efficacy, improvement, enhancement, implementation,
guideline, practice, plan, policy, recommendation,
management, design, structure, evaluation,
algorithm, process, assessment, mechanism,
procedure, initiative, rollout, deployment, trial,
interdisciplinary, collaborative, multimodal,
integrated, systematic, scalable, feasible,
evidence-based, optimization \\
\midrule

Limitation-related &
limitation, limit, lack, weakness, shortcoming, bias \\
\bottomrule
\end{tabularx}

\caption{Keyword categories used for paragraph filtering during semantic retrieval.}
\label{tab:keyword_lists}

\end{table}

\subparagraph{(1) PST Type}

\begin{itemize}
    \item {\color{green!60!black}\textbf{Snippets 1 \& 2}}: Problem and Solution paragraphs were keyword filtered. The table is assumed to be in the solution document; the problem document provides motivation. We then selected the most relevant paragraph per document using 0.2 * TF-IDF + 0.8 * BERTScore.
    \item {\color{green!60!black}\textbf{Snippet 3}}: The most relevant table content, selected based on similarity to Snippet 2.
    \item \textbf{Snippet 4}: Paragraph in the document referring the table.
\end{itemize}
\subparagraph{(2) TS \& TSL Type}
\begin{itemize}
    \item \textbf{Snippet 1}: Table metadata from the source document.
    \item \textbf{Snippet 2}: Paragraph referring to the table.
    \item \textbf{Snippet 3}: Abstract of the referenced document, linked via citation markers in the table.
    \item {\color{green!60!black}\textbf{Snippet 4}}: Paragraph describing the cited document's limitations. Snippet 4 was retrieved via a query-driven similarity search using the prompt “What is the limitation of this paper?” and subsequently keyword-filtered. If the similarity score for snippet 4 fell below 0.4, the snippet was discarded, and the QA instance was reclassified as TS type, containing only the first three snippets.
\end{itemize}


\vspace{-0.3cm}
\FloatBarrier
\subsection{Computational Cost Details}
\label{computation_efficiency}

We measured generation and matching costs to assess the computational scalability of our pipeline. As summarized in Table~\ref{tab:qa_generation_cost}, the measured question/answer generation time and GPU usage indicate that once prompts are prepared, scaling up the QA set can be done efficiently with moderate resource requirements.This demonstrates that the proposed LLM-driven pipeline is computationally scalable for large-scale QA dataset construction.

\begin{table}[t]
\centering
\small
\renewcommand{\arraystretch}{1.15}
\setlength{\tabcolsep}{3pt}
\begin{tabularx}{\columnwidth}{
    l
    >{\centering\arraybackslash}X
    >{\centering\arraybackslash}X
    >{\centering\arraybackslash}X
    >{\centering\arraybackslash}X
}
\toprule
\textbf{Question Type} &
\textbf{Time per Q gen} &
\textbf{GPU peak usage (Q gen)} &
\textbf{Time per A matching} &
\textbf{GPU peak usage (A matching)} \\
\midrule
Comparison     & 6.84s  & 11.82GB & 5.34s  & 0.99GB \\
Non-comparison & 2.22s  & 12.10GB & ---    & --- \\
PST            & 12.56s & 14.89GB & 0.05s  & 15.38GB \\
TSL            & 24.03s & ---     & 4.45s  & --- \\
\bottomrule
\end{tabularx}
\caption{\textbf{Computation efficiency of QA generation and matching.} 
Time and GPU usage measured per question (Q gen) and answer matching (A matching) step 
for each question type. The results show that once prompts are prepared, 
QA scaling can be efficiently achieved with moderate GPU memory usage.}
\label{tab:qa_generation_cost}
\end{table}

\section{Dataset Quality Assurance}
\label{appendix:quality-assurance}
\subsection{LLM-based Dataset Quality Assurance}
\label{appendix:llm-quality-assurance}
\begin{figure}[h]
    \centering
    \includegraphics[width=0.99\linewidth,height=1.9cm]{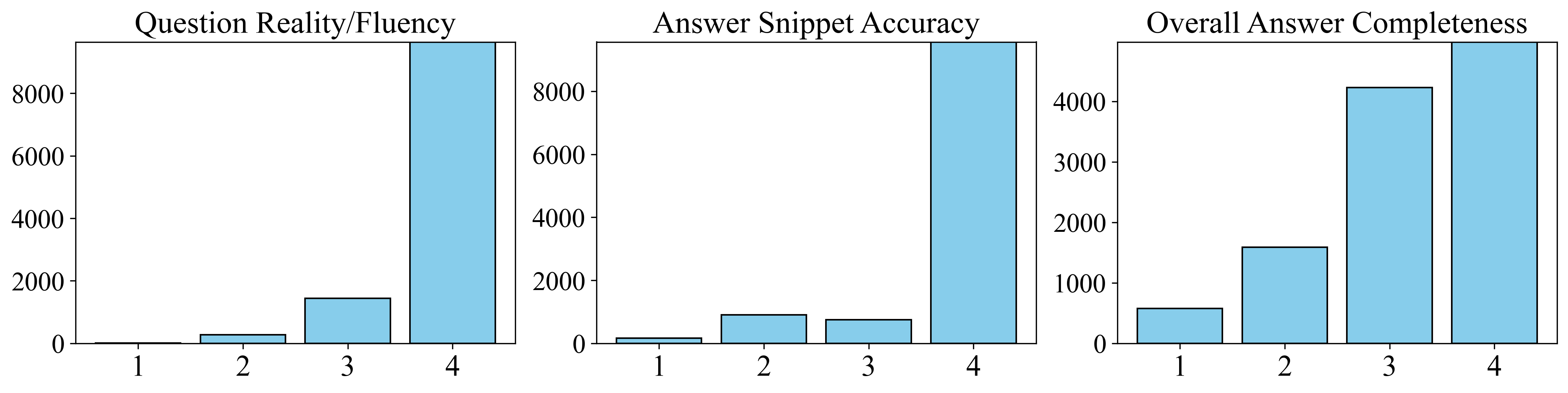}
    \vspace{-0.7cm}
\caption{\textbf{Qwen evaluation across the full dataset}}
\vspace{-1em}
\label{fig:llm _result}
\end{figure}

\begin{figure}[h]
    \centering
\begin{subfigure}[t]{0.45\textwidth}
    \includegraphics[width=\linewidth,height=2.1cm]{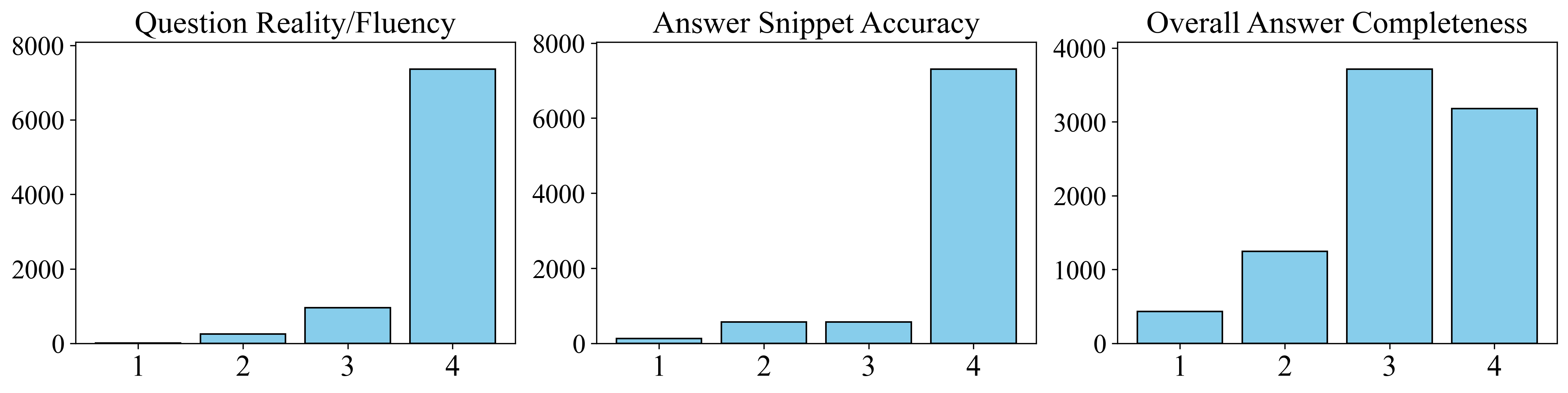}
    \vspace{-0.3cm}
    \caption{Paragraph-Oriented}
    \label{fig:llm_based_paragraph-oriented}
\end{subfigure}
\begin{subfigure}[h]{0.45\textwidth}
    \includegraphics[width=\linewidth,height=2.1cm]{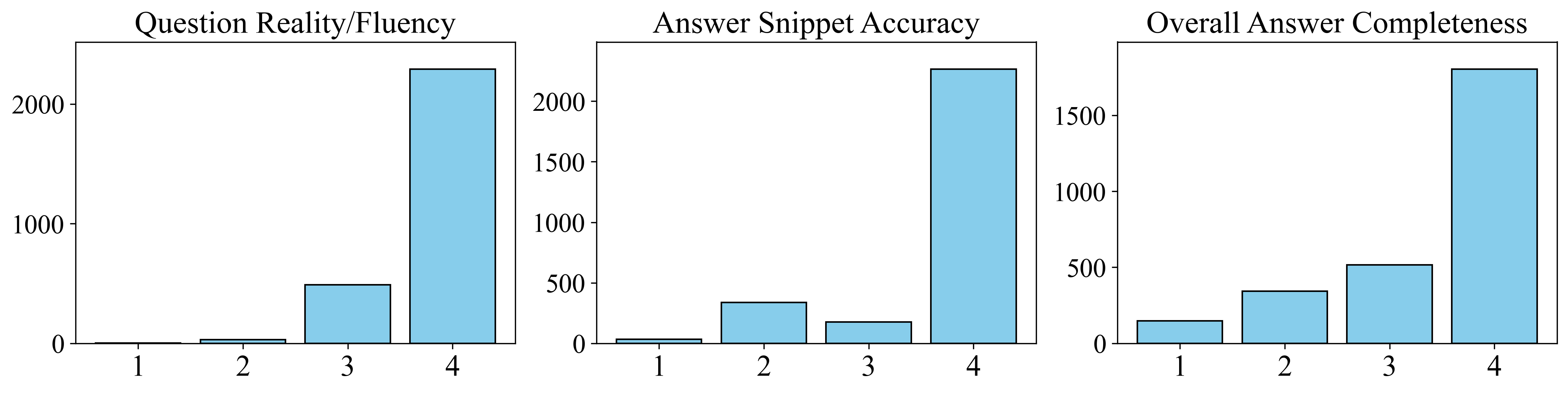}
    \vspace{-0.3cm}
    \caption{Table-Oriented}
    \end{subfigure}
    \label{fig:llm_based_table-oriented}
\caption{\textbf{LLM based Quality Assurance result}, (1) Question Reality/Fluency (2) Answer Snippet Accuracy (3) Overall Answer Completeness. Scores are based on a 4-point scale
(1:poor, 4: excellent).}
\label{fig:llm_based_quality_assurance_result}
\end{figure}

As shown in Figure~\ref{fig:llm _result}, the full dataset achieves consistently high scores across all criteria. Among question types, Paragraph-Oriented QA performs particularly well, with most scores rated 3 or 4 (Figure~\ref{fig:llm_based_quality_assurance_result}). Some instances had lower accuracy, likely due to snippets covering only parts of the question, reducing alignment with its overall intent. Table-Oriented QA showed slightly lower Reality/Fluency due to its format, but still maintained high overall quality. These results confirm that DocHop-QA exhibits robust quality and is suitable for complex QA tasks. All three LLM judges (Qwen3.5-9B, GPT-4o, and Gemini2.5~Flash) were prompted with the same evaluation criteria, output format, and few-shot examples (Table~\ref{tab:eval-prompt}, Table~\ref{tab:eval-fewshot}).

\vspace{-0.5em}
\subsection{Human-based Dataset Quality Assurance}
\label{appendix:human_quality_assurance}
\paragraph{Annotator Demographics}
Fifteen annotators participated in the human evaluation.The sample was gender-balanced (7F/8M) and included 6 undergraduates, 6 master’s, and 3 doctoral students. The participants had academic backgrounds in business (n=3), industrial and management engineering (n=4), computer engineering unrelated to AI (n=2), and AI-related computer science (n=6). The age distribution was as follows: 20 years (n = 8), 30 years (n = 5) and older than 50 years (n = 2).

\paragraph{Institutional Review Board Statement}
The study was approved by the Institutional Review Board (Exemption Number: PIRB-2025-E020, Date: June 9, 2025).

\begin{figure*}[ht]
    \centering

    \begin{minipage}{0.95\textwidth}
        \centering
        \includegraphics[width=\textwidth]{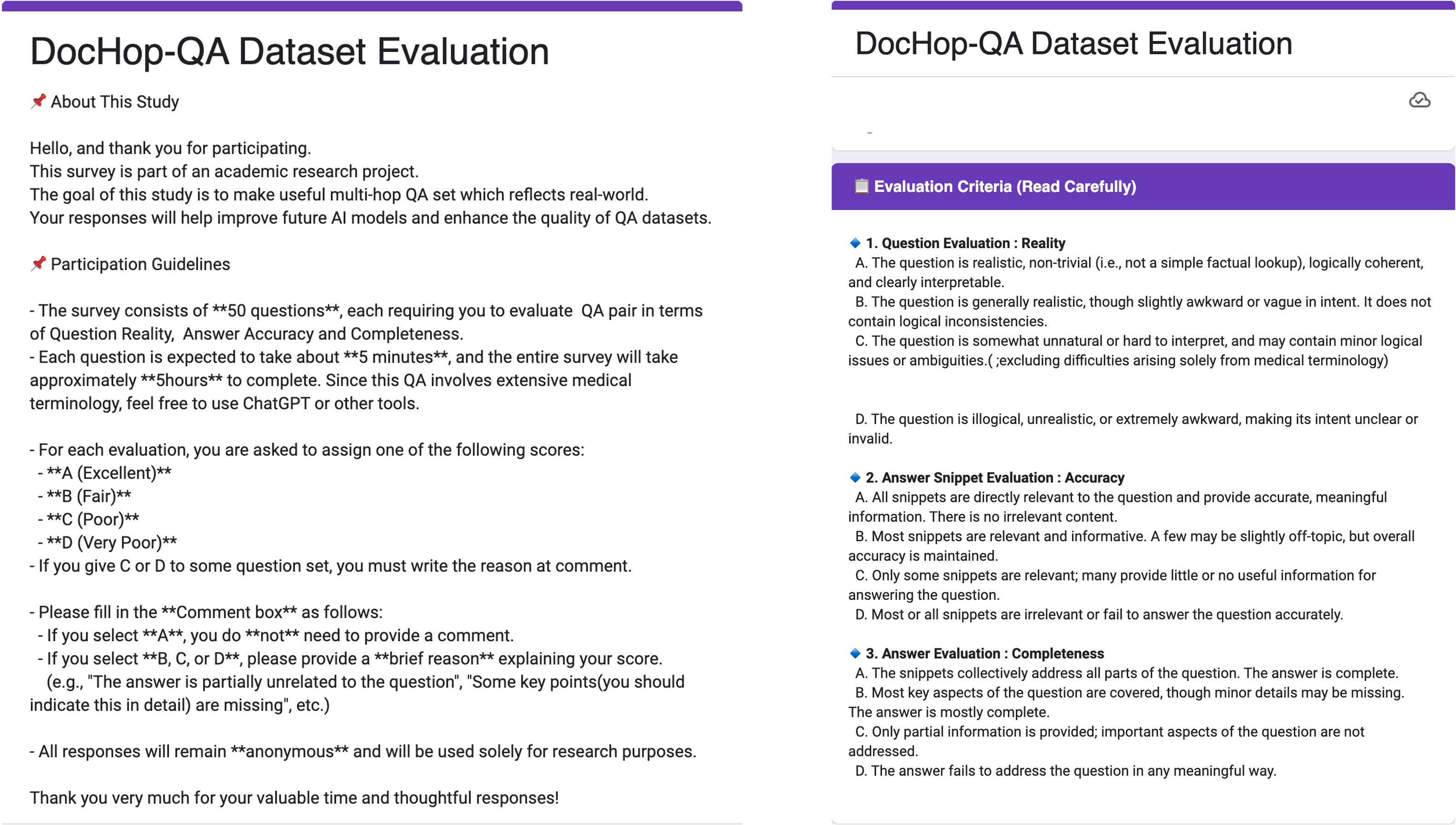}
        \vspace{1em}
        \small (a) Overview page providing the study purpose and outlining the evaluation criteria.
    \end{minipage}

    \vspace{2em}

    \begin{minipage}{0.95\textwidth}
        \centering
        \includegraphics[width=\textwidth]{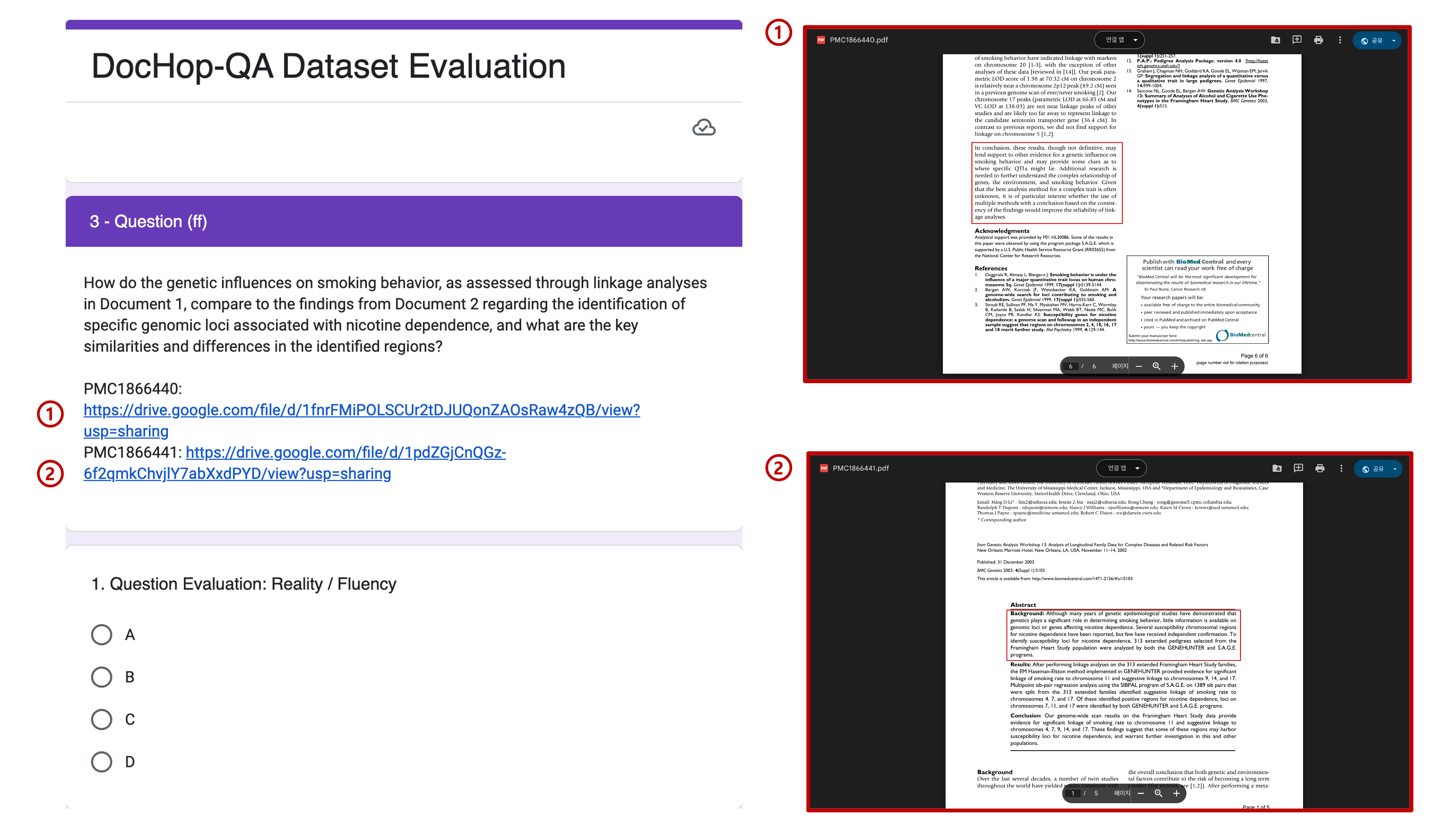}
        \vspace{1em}
        \small (b) Example QA instance with hyperlinks to PDF documents displaying the answer snippet with a bounding box.
    \end{minipage}

    \caption{\textbf{Screenshots of the structured Google Form interface used for human evaluation.}}
    \label{fig:human_evaluation_form_all}
\end{figure*}

\paragraph{Evaluation Interface}
Annotators rated each QA pair via a Google Form interface (Figure~\ref{fig:human_evaluation_form_all}). The form included Likert-scale items for three evaluation criteria and mandatory comment fields for low ratings, ensuring consistent evaluation and facilitating both quantitative and qualitative feedback.

\begin{table}[h]
\centering
\small
\setlength{\tabcolsep}{4pt}
\resizebox{\columnwidth}{!}{
\begin{tabular}{l l c c}
\toprule
\textbf{Evaluation} & \textbf{Model} & \textbf{Spearman’s $\rho$} & \textbf{Krippendorff’s $\alpha$} \\
\midrule
\multirow{3}{*}{\textbf{Completeness}} 
    & GPT-4o & 0.354 & 0.199 \\
    & Gemini-2.5    & 0.606 & 0.382 \\
    & Qwen3.5-9B    & 0.037 & 0.015 \\
\midrule
\multirow{3}{*}{\textbf{Accuracy}} 
    & GPT-4o & 0.240 & 0.144 \\
    & Gemini-2.5    & 0.395 & 0.285 \\
    & Qwen3.5-9B    & 0.027 & -0.441 \\
\midrule
\multirow{3}{*}{\textbf{Reality}} 
    & GPT-4o & -0.163 & -0.160 \\
    & Gemini-2.5    & 0.445 & 0.000 \\
    & Qwen3.5-9B    & 0.099 & 0.113 \\
\bottomrule
\end{tabular}
}
\caption{Human--LLM agreement measured by Spearman’s $\rho$ and Krippendorff’s $\alpha$ on a 50-instance subset.}
\label{tab:human_llm_agreement}
\end{table}

\subsection{Human-LLM Agreement Analysis}
\label{appendix:human_llm_agreement_analysis}
To further assess the alignment between human judgments and automated evaluators in multiple ways, we report Spearman's $\rho$ and Krippendorff's $\alpha$ on the 50-instance subset. Table~\ref{tab:human_llm_agreement} summarizes the agreement across three evaluation dimensions. We observe moderate agreement between human annotators and LLM evaluators, with variation across models and criteria. For the \textit{Reality} dimension, agreement scores are low due to a ceiling effect, where most instances receive high scores, resulting in low variance and unstable statistics. This reflects a known trait of variance-based coefficients such as $\rho$ and $\alpha$. In our setting, AC2, a metric that remains interpretable under such skew, was the most valid choice.
Beyond evaluation scores, we manually verified whether automated snippet matching correctly identifies human-identified supporting evidence. In most cases, retrieved snippets align well with human annotations, indicating limited noise from automated matching. These results suggest that automated evaluation can approximate human judgment when appropriately calibrated.

\vspace{-0.5em}
\section{Dataset Analysis Details}
\label{appendix:dataset_analysis}
\subsection{Dataset Coverage}
\label{appendix:document_complexity}
\begin{figure}[h]
  \centering
  \begin{subfigure}[t]{0.23\textwidth}
    \includegraphics[width=\textwidth]{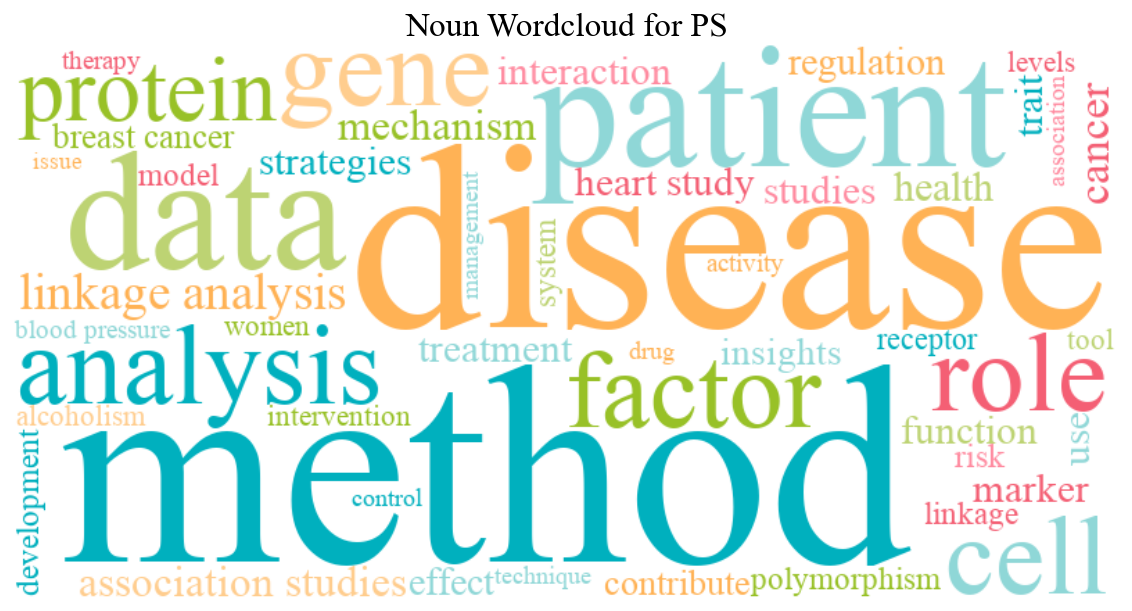}
    \caption{Noun}
  \end{subfigure}
  \hfill
  \begin{subfigure}[t]{0.23\textwidth}
    \includegraphics[width=\textwidth]{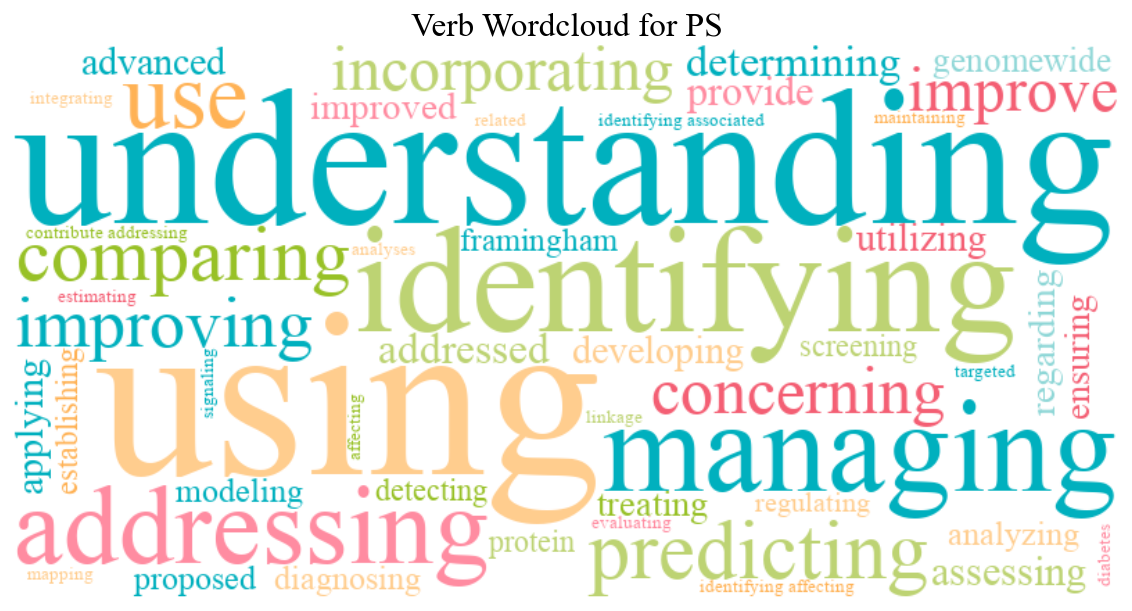}
    \caption{Verb}
  \end{subfigure}
  \hfill
  \begin{subfigure}[t]{0.23\textwidth}
    \includegraphics[width=\textwidth]{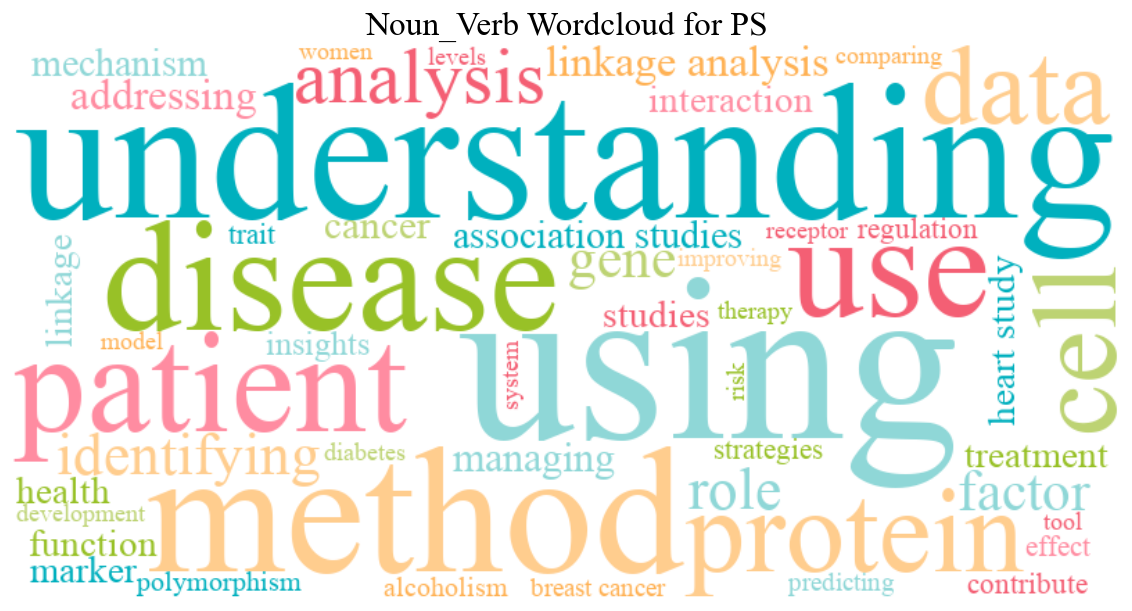}
    \caption{Noun \& Verb}
  \end{subfigure}
  \hfill
  \begin{subfigure}[t]{0.23\textwidth}
    \includegraphics[width=\textwidth]{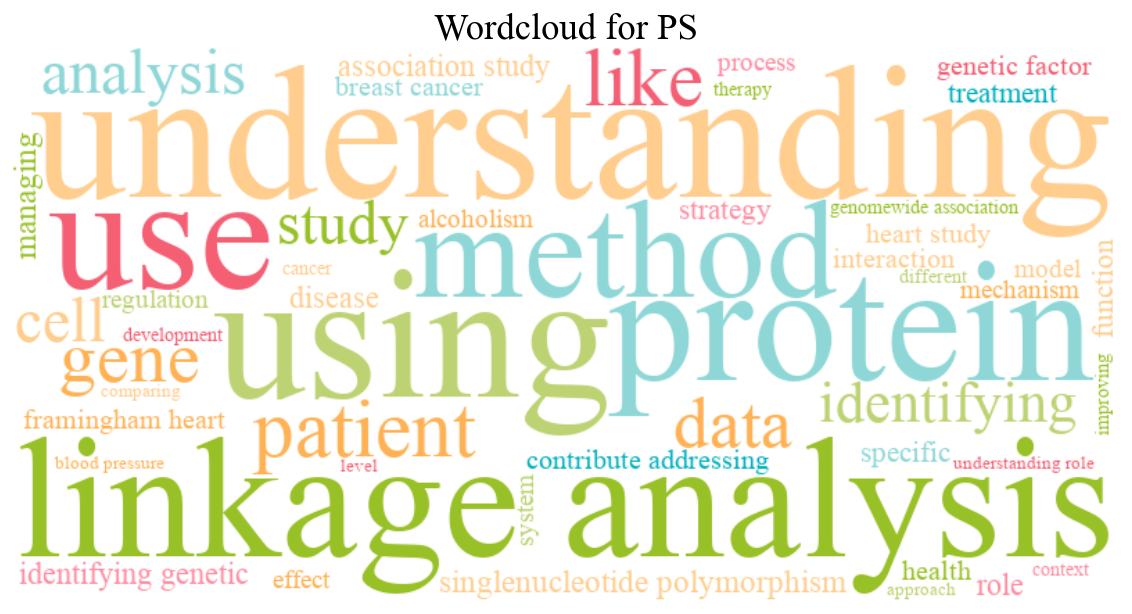}
    \caption{All Words}
  \end{subfigure}
  \caption{\textbf{Frequent words in \textbf{PS-type} questions:} Showing terms like \textit{“method”} and \textit{“disease,”} indicating conceptual understanding in biomedical topics.}
  \label{fig:wordcloud-ps}
\end{figure}

\begin{figure}[H]
  \centering
  \begin{subfigure}[t]{0.23\textwidth}
    \includegraphics[width=\textwidth]{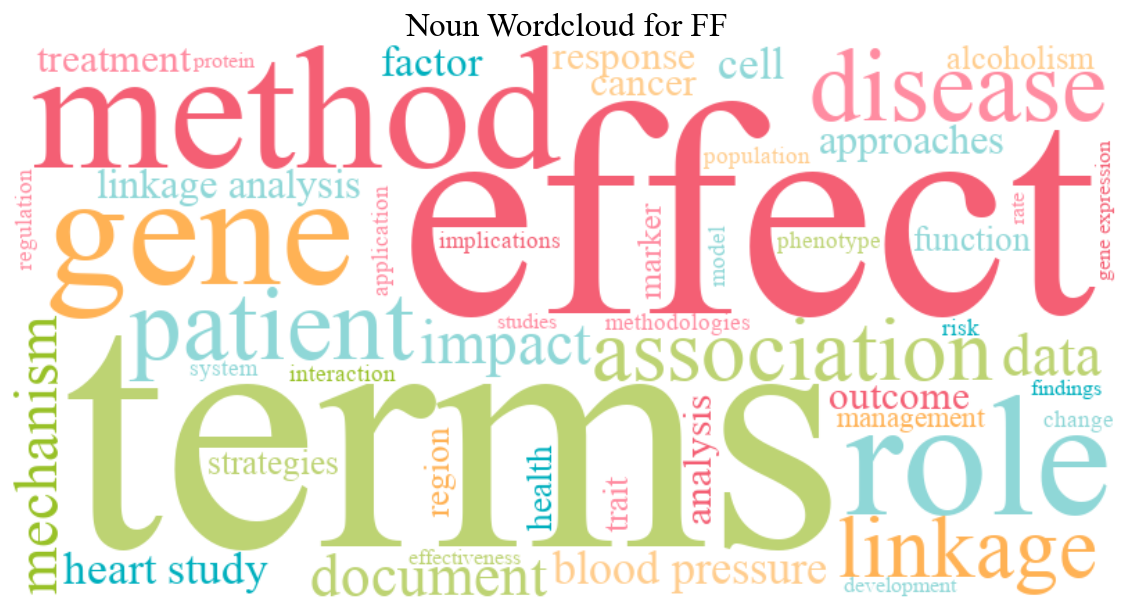}
    \caption{Noun}
  \end{subfigure}
  \hfill
  \begin{subfigure}[t]{0.23\textwidth}
    \includegraphics[width=\textwidth]{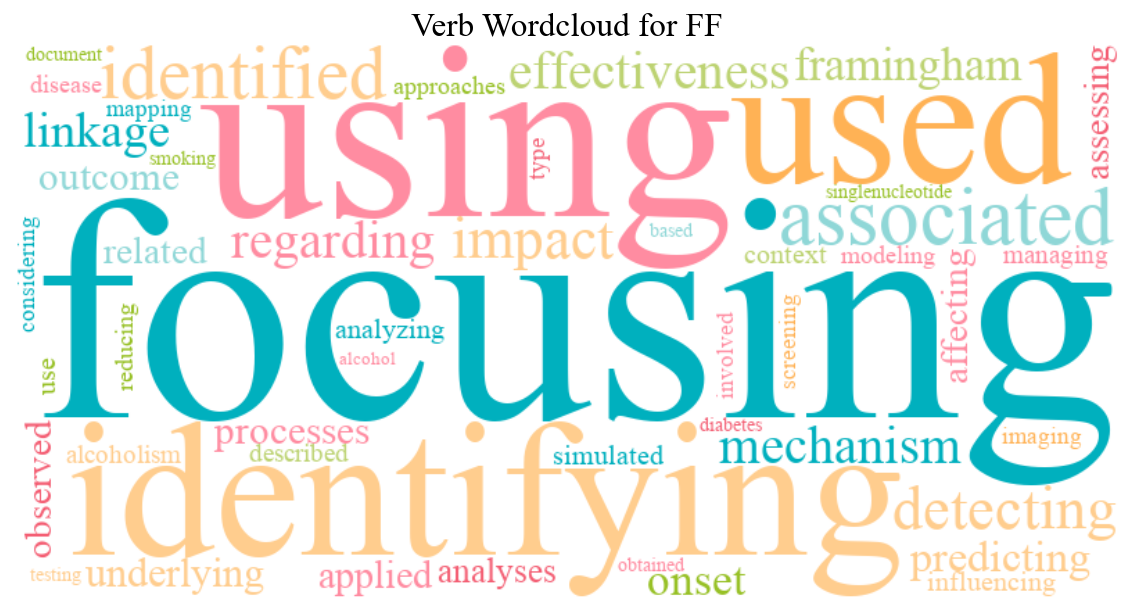}
    \caption{Verb}
  \end{subfigure}
  \hfill
  \begin{subfigure}[t]{0.23\textwidth}
    \includegraphics[width=\textwidth]{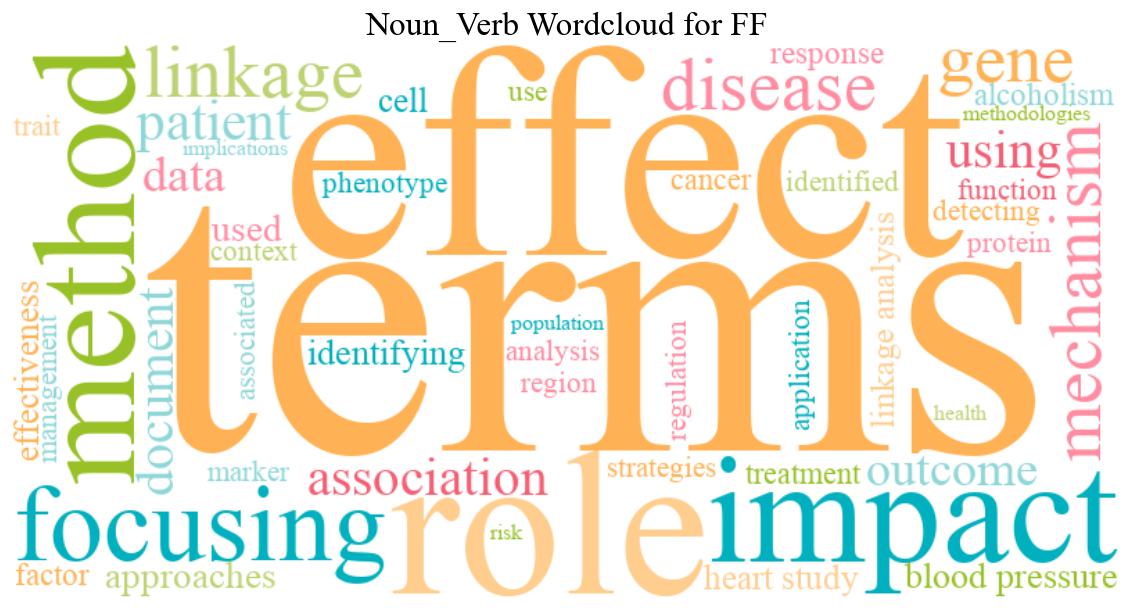}
    \caption{Noun \& Verb}
  \end{subfigure}
  \hfill
  \begin{subfigure}[t]{0.23\textwidth}
    \includegraphics[width=\textwidth]{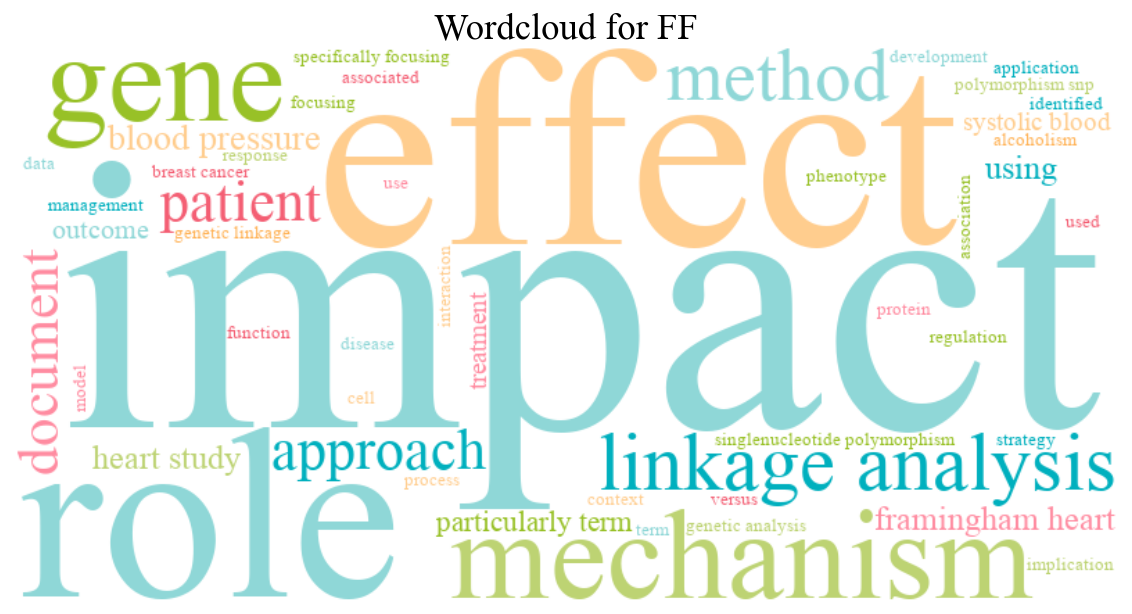}
    \caption{All Words}
  \end{subfigure}
  \caption{\textbf{Frequent words in \textbf{FF-type} questions:} Showing terms like \textit{associated}, \textit{effect}, and \textit{mechanism}, indicating functional comparisons between two entities.}
  \label{fig:wordcloud-ff}
\end{figure}

\begin{figure}[H]
  \centering
  \begin{subfigure}[t]{0.23\textwidth}
    \includegraphics[width=\textwidth]{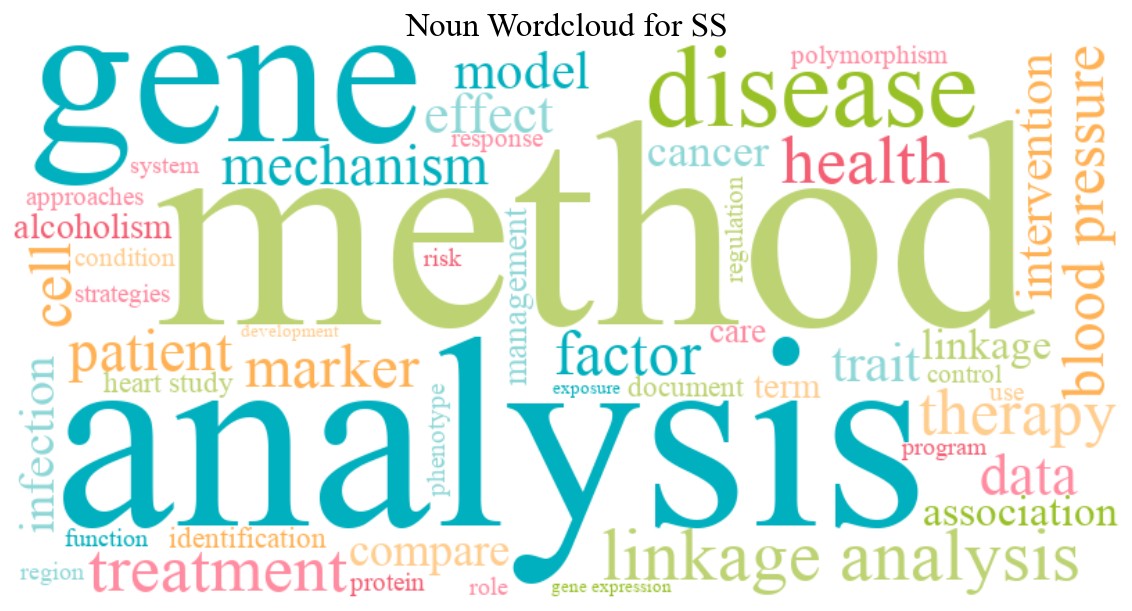}
    \caption{Noun}
  \end{subfigure}
  \hfill
  \begin{subfigure}[t]{0.23\textwidth}
    \includegraphics[width=\textwidth]{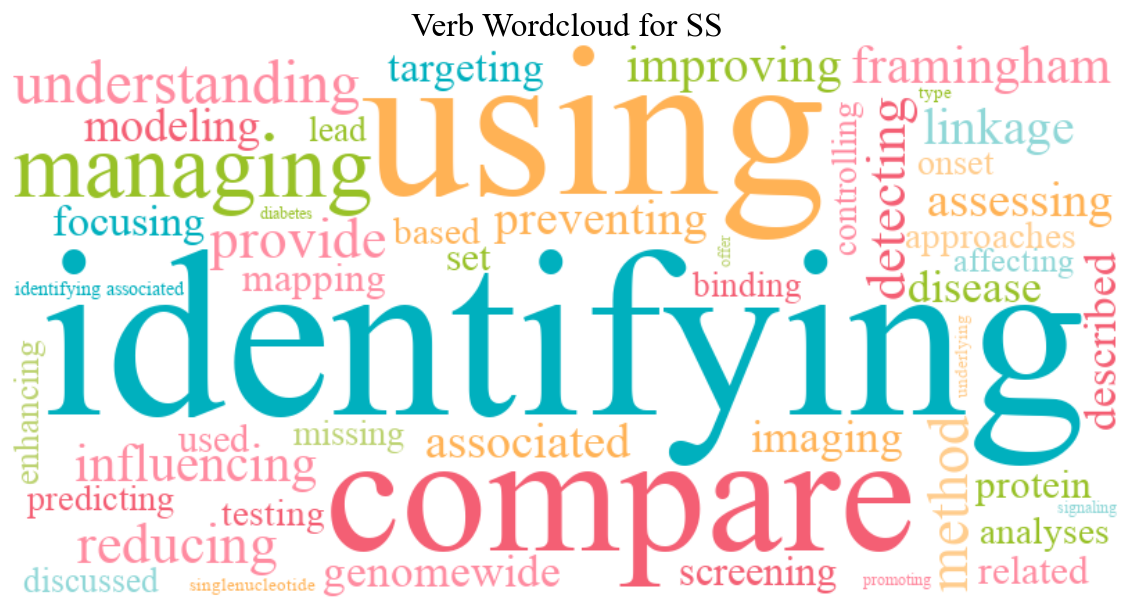}
    \caption{Verb}
  \end{subfigure}
  \hfill
  \begin{subfigure}[t]{0.23\textwidth}
    \includegraphics[width=\textwidth]{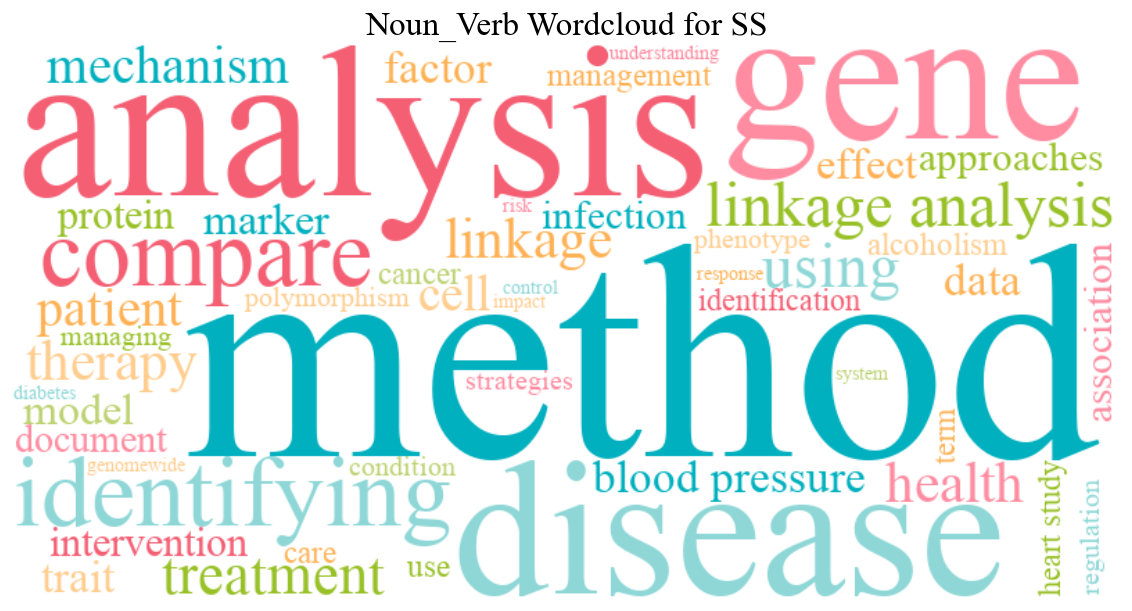}
    \caption{Noun \& Verb}
  \end{subfigure}
  \hfill
  \begin{subfigure}[t]{0.23\textwidth}
    \includegraphics[width=\textwidth]{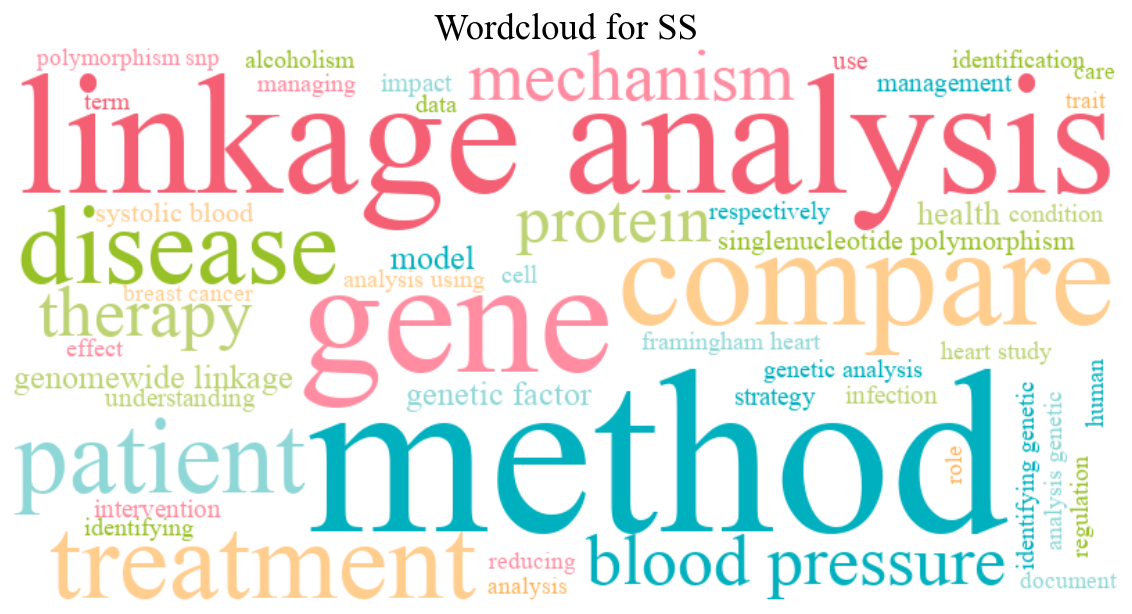}
    \caption{All Words}
  \end{subfigure}
  \caption{\textbf{Frequent words in \textbf{SS-type} questions:} Showing terms like \textit{compare}, \textit{method}, and \textit{analysis}, reflecting contrast of scientific approaches to biomedical problems.}
  \label{fig:wordcloud-ss}
\end{figure}

\begin{figure}[H]
  \centering
  \begin{subfigure}[t]{0.23\textwidth}
    \includegraphics[width=\textwidth]{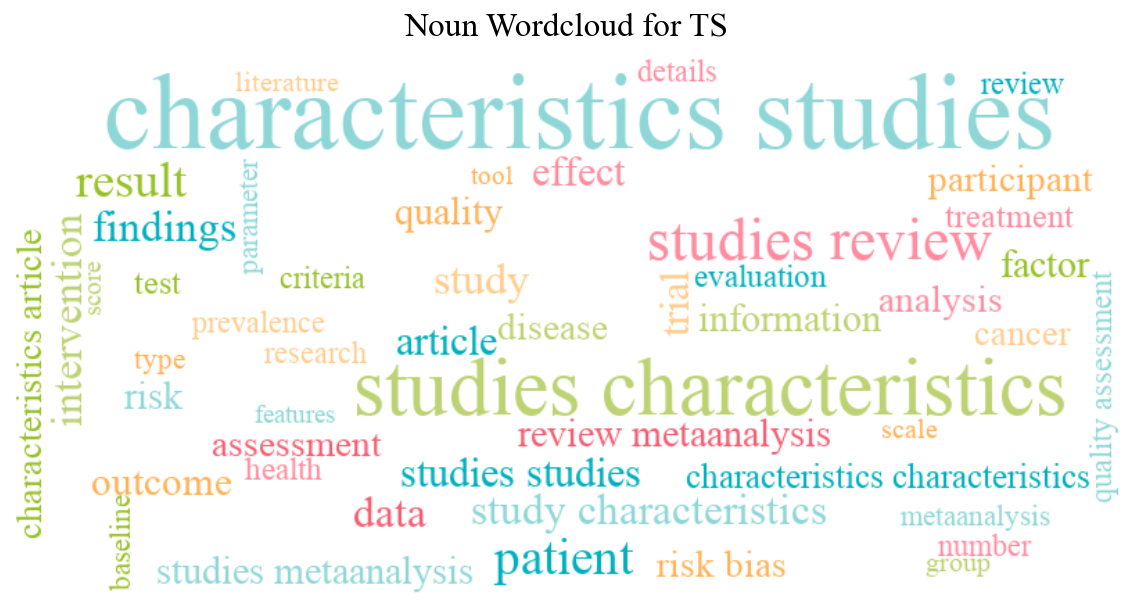}
    \caption{Noun}
  \end{subfigure}
  \hfill
  \begin{subfigure}[t]{0.23\textwidth}
    \includegraphics[width=\textwidth]{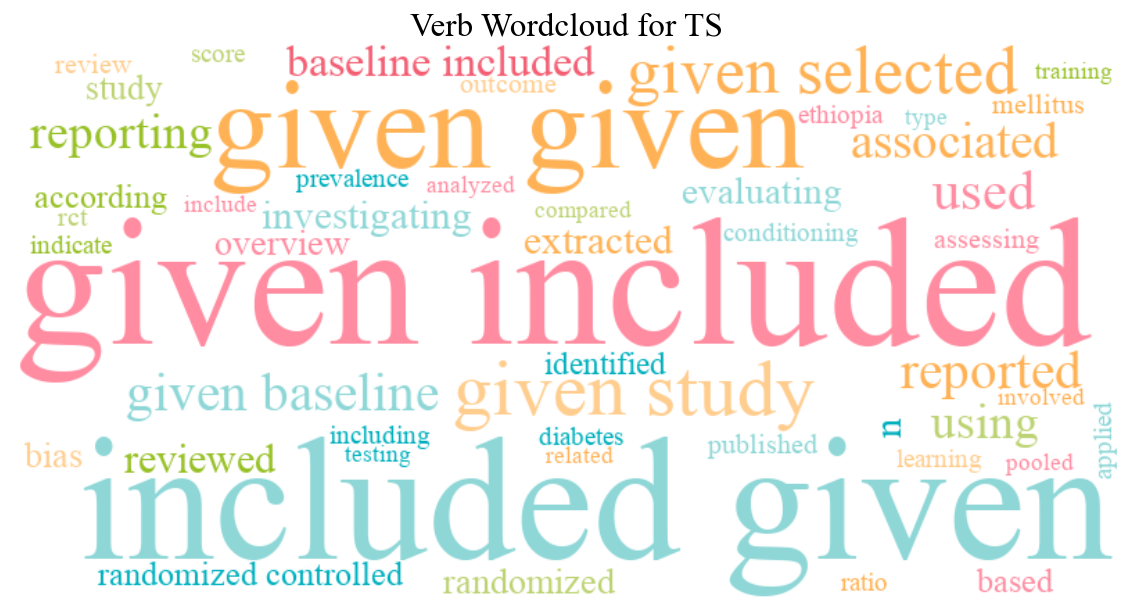}
    \caption{Verb}
  \end{subfigure}
  \hfill
  \begin{subfigure}[t]{0.23\textwidth}
    \includegraphics[width=\textwidth]{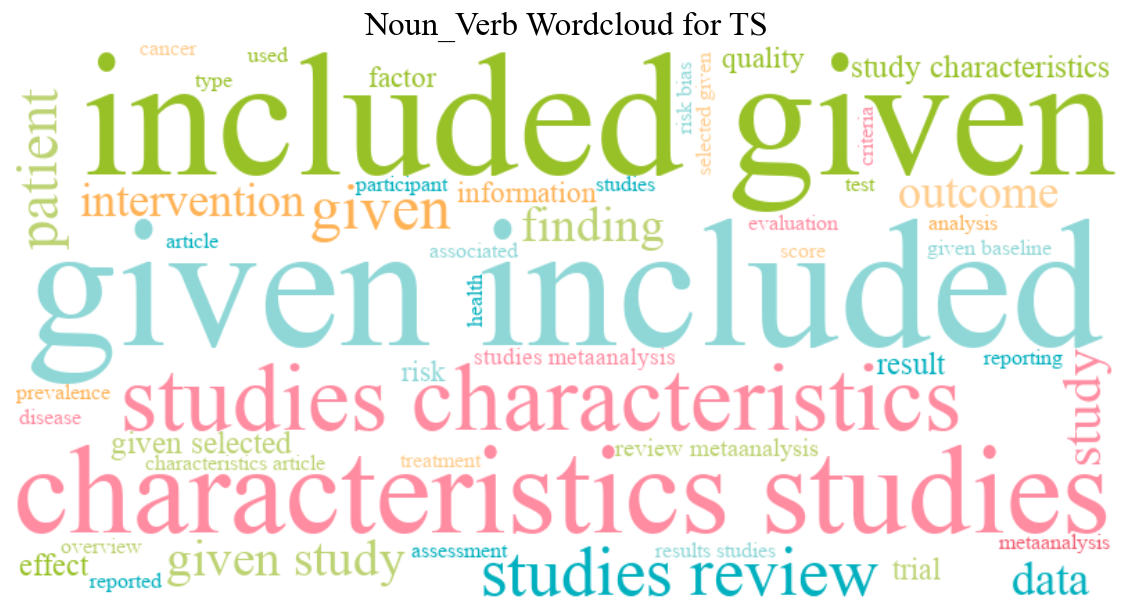}
    \caption{Noun \& Verb}
  \end{subfigure}
  \hfill
  \begin{subfigure}[t]{0.23\textwidth}
    \includegraphics[width=\textwidth]{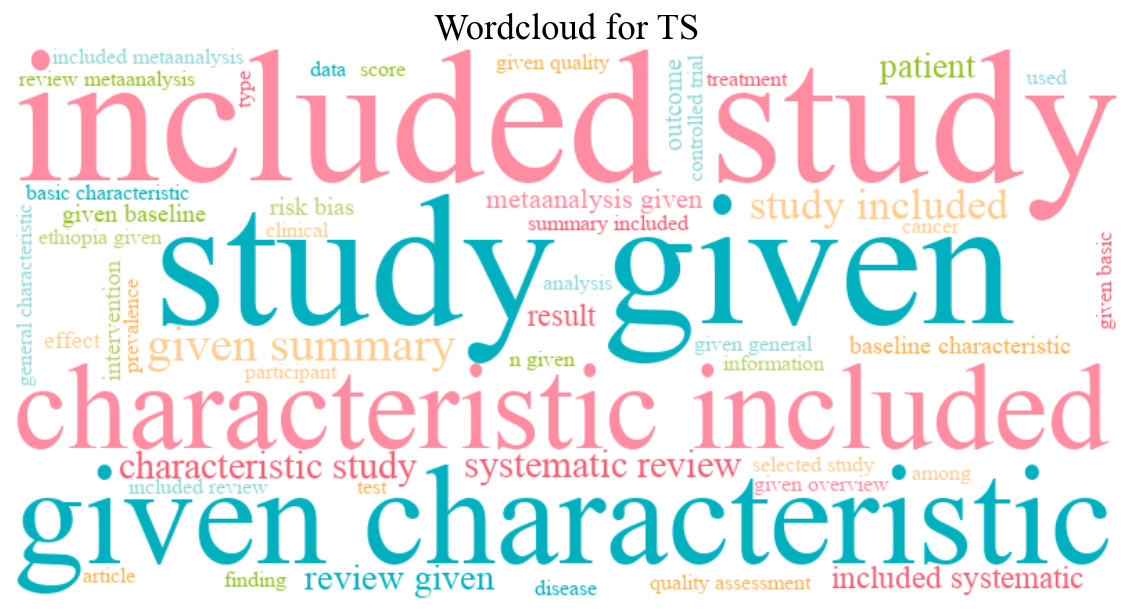}
    \caption{All Words}
  \end{subfigure}
  \caption{\textbf{Frequent words in \textbf{TS-type} questions:} Showing terms like \textit{given}, \textit{included}, and \textit{study}, reflecting their deterministic, table-driven design with limited lexical diversity.}
  \label{fig:wordcloud-ts}
\end{figure}

\paragraph{Keyword Analysis}
\label{appendix:wordclouds}
To validate the topical diversity and domain coverage of DocHop-QA, we conducted word cloud analysis on question texts across all 11 question concepts. After removing stopwords and terms specific to each question concept, we extracted nouns, verbs, and combinations of nouns and verbs from each question. Word clouds were generated per concept (Figures~\ref{fig:wordcloud-ps} to \ref{fig:wordcloud-ts}). As expected from data sourced from PubMed, biomedical terms such as \textit{patient}, \textit{gene}, and \textit{cell} dominate across concepts, confirming domain specificity. Importantly, each concept exhibits distinct keyword distributions. Concepts without comparison focus on mechanistic terms, while comparison concepts surface contrastive terms such as \textit{compare} and \textit{associated}. Together, these patterns confirm that DocHop-QA achieves both broad topical coverage across biomedical subfields and structural diversity across reasoning types.


\begin{figure}[h]
  \centering
  \begin{subfigure}[t]{0.23\textwidth}
    \includegraphics[width=\textwidth]{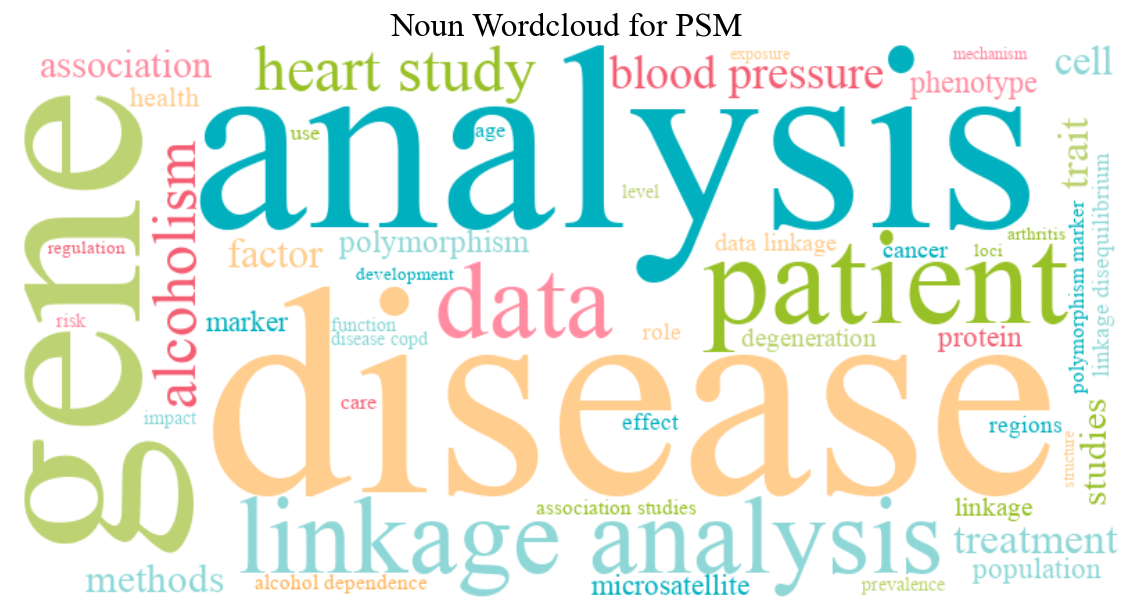}
    \caption{Noun}
  \end{subfigure}
  \hfill
  \begin{subfigure}[t]{0.23\textwidth}
    \includegraphics[width=\textwidth]{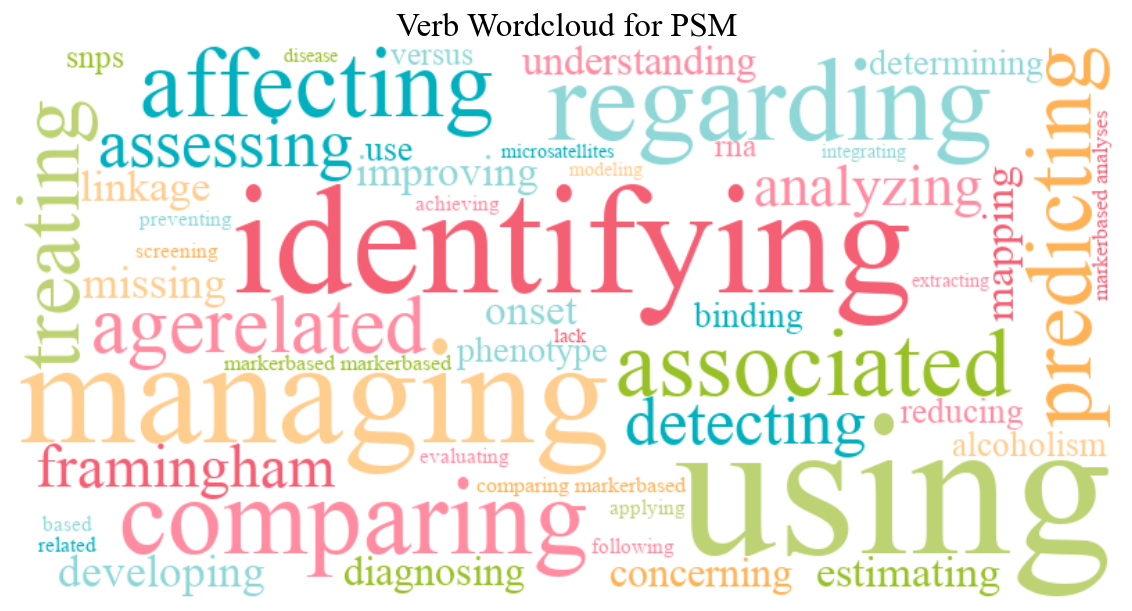}
    \caption{Verb}
  \end{subfigure}
  \hfill
  \begin{subfigure}[t]{0.23\textwidth}
    \includegraphics[width=\textwidth]{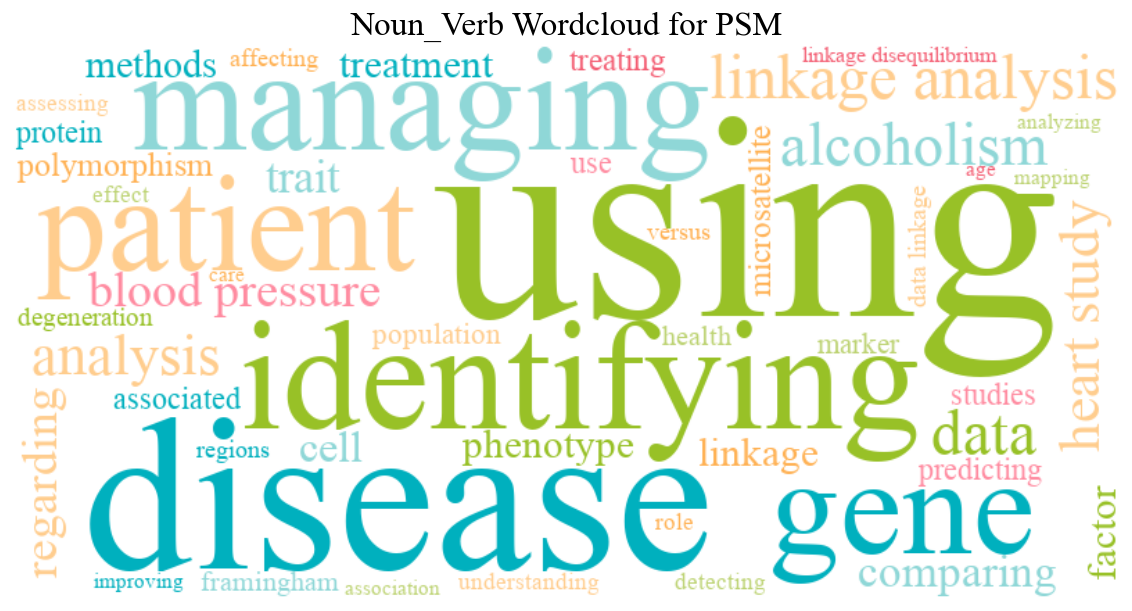}
    \caption{Noun \& Verb}
  \end{subfigure}
  \hfill
  \begin{subfigure}[t]{0.23\textwidth}
    \includegraphics[width=\textwidth]{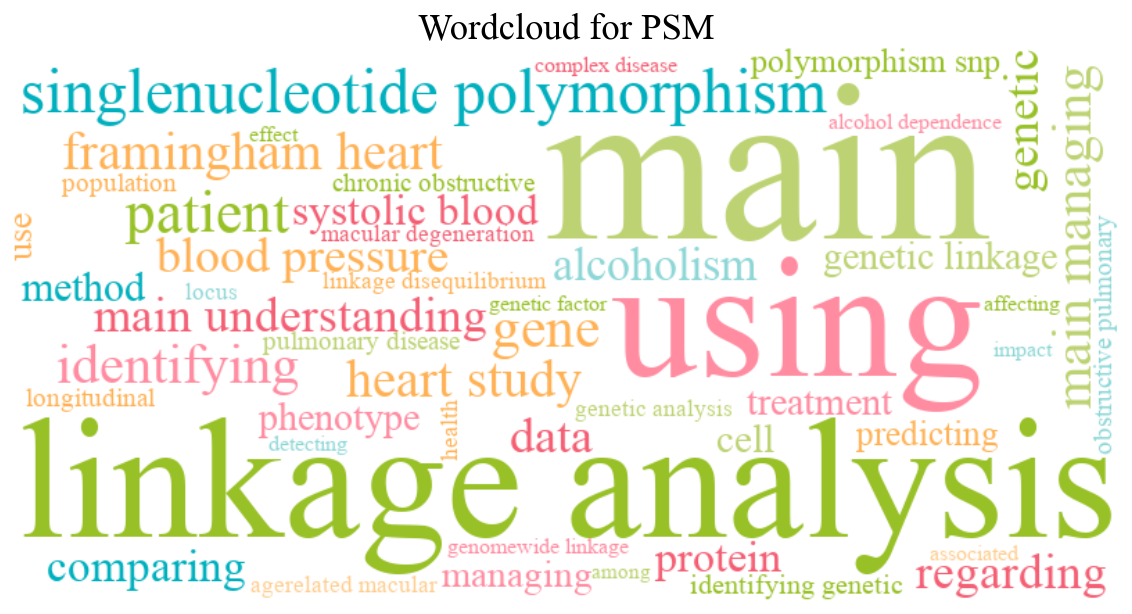}
    \caption{All Words}
  \end{subfigure}
  \caption{\textbf{Frequent words in \textbf{PSM-type} questions:} Showing terms like \textit{analysis}, \textit{affecting}, and \textit{identifying}, highlighting analytical reasoning about genetic mechanisms and disease traits.}
\end{figure}

\begin{figure}[H]
  \centering
  \begin{subfigure}[t]{0.23\textwidth}
    \includegraphics[width=\textwidth]{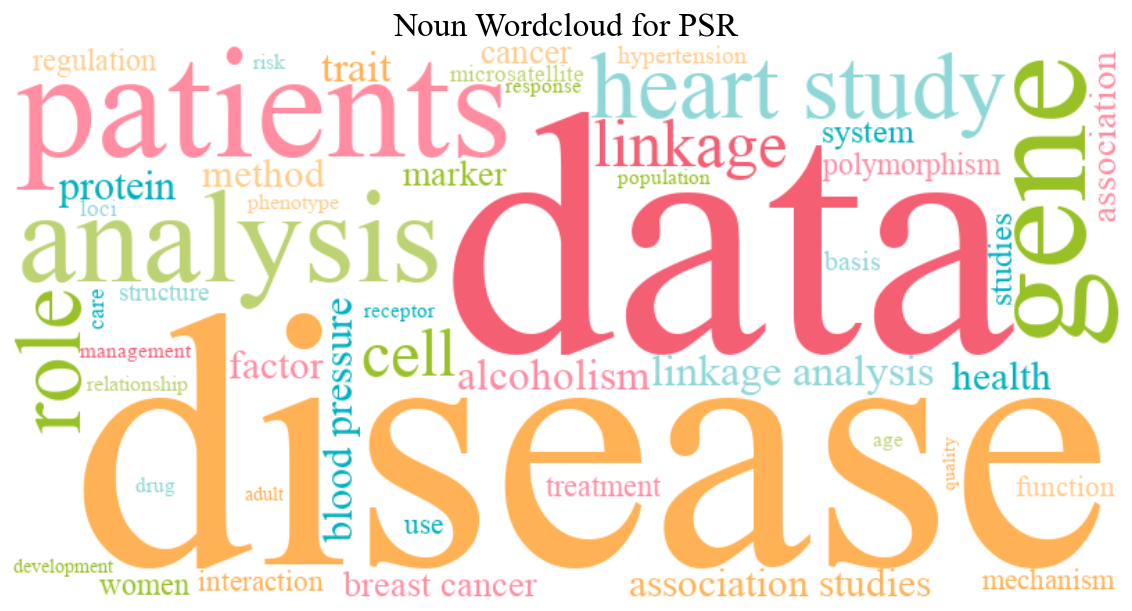}
    \caption{Noun}
  \end{subfigure}
  \hfill
  \begin{subfigure}[t]{0.23\textwidth}
    \includegraphics[width=\textwidth]{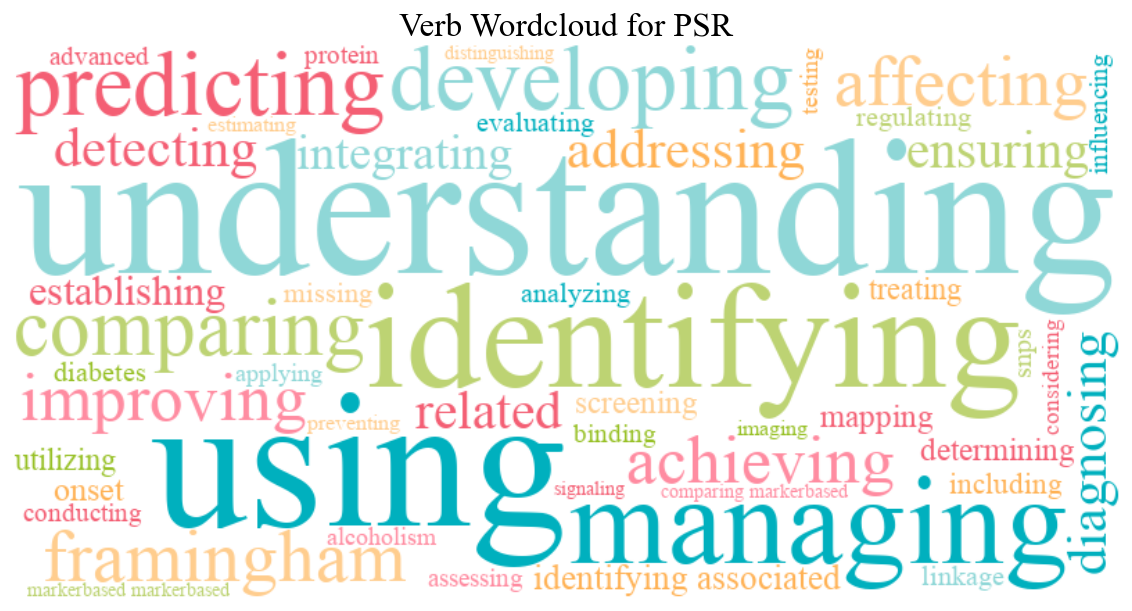}
    \caption{Verb}
  \end{subfigure}
  \hfill
  \begin{subfigure}[t]{0.23\textwidth}
    \includegraphics[width=\textwidth]{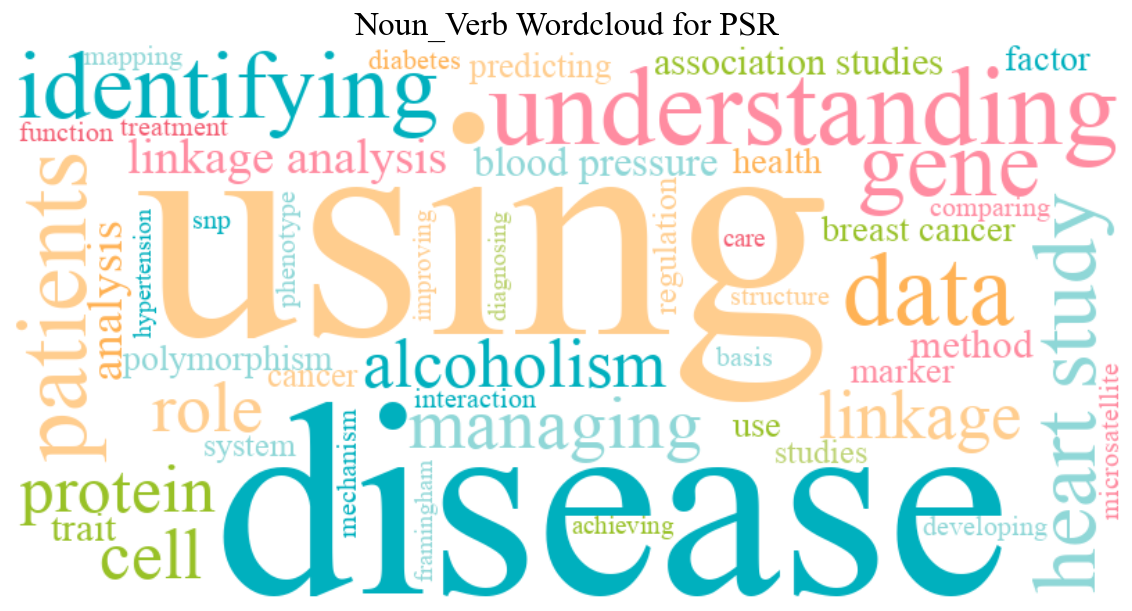}
    \caption{Noun \& Verb}
  \end{subfigure}
  \hfill
  \begin{subfigure}[t]{0.23\textwidth}
    \includegraphics[width=\textwidth]{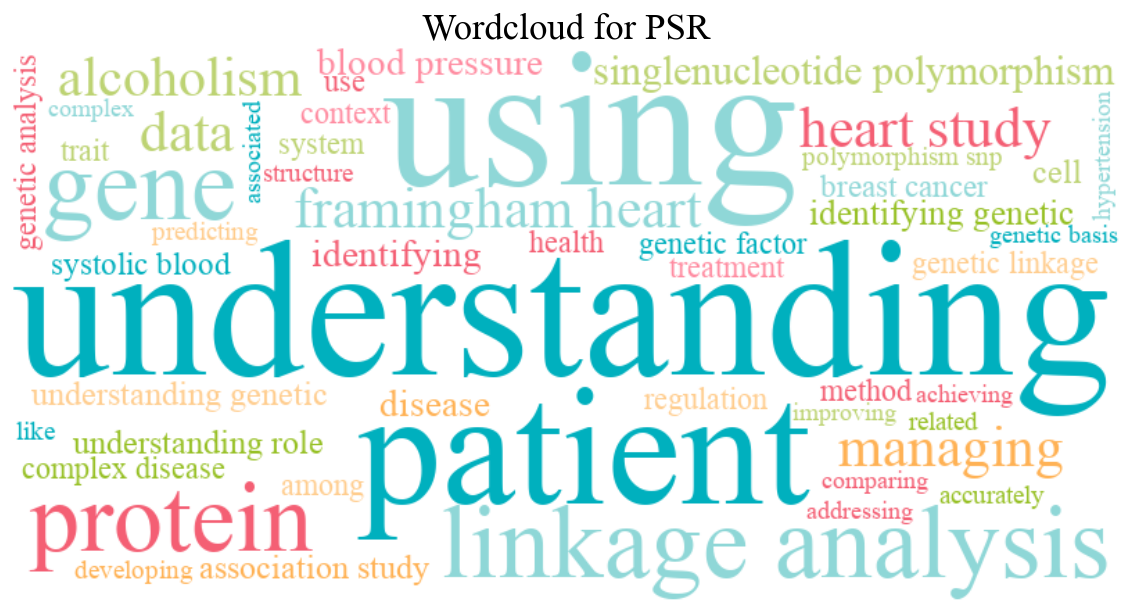}
    \caption{All Words}
  \end{subfigure}
  \caption{\textbf{Frequent words in \textbf{PSR-type} questions:} Suggesting a focus on result-oriented reasoning about biomedical solutions or interventions.}
\end{figure}

\begin{figure}[H]
  \centering
  \begin{subfigure}[t]{0.23\textwidth}
    \includegraphics[width=\textwidth]{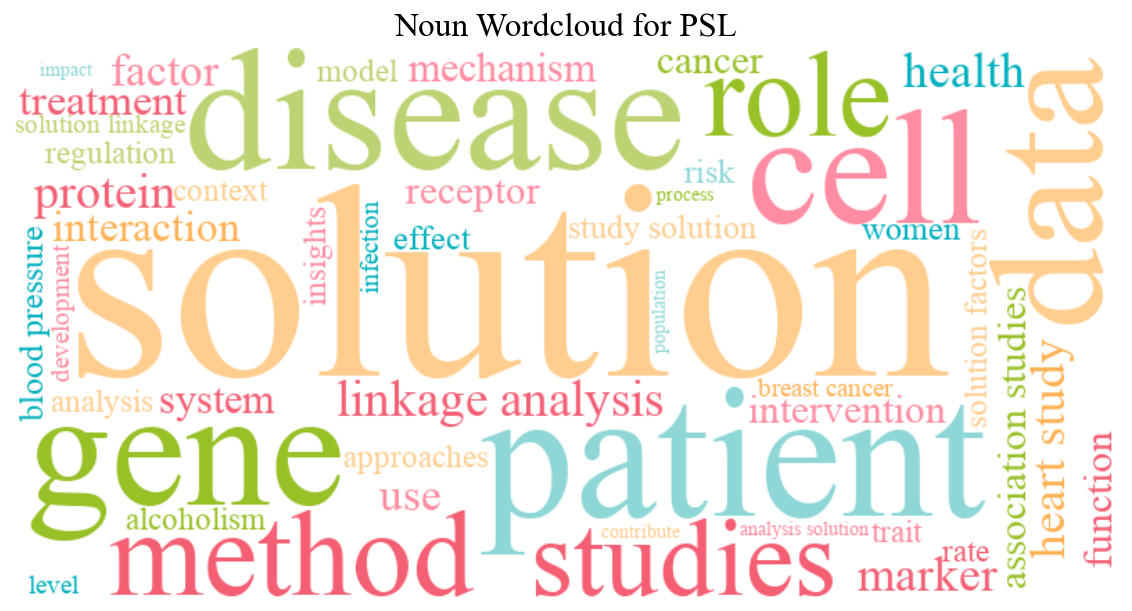}
    \caption{Noun}
  \end{subfigure}
  \hfill
  \begin{subfigure}[t]{0.23\textwidth}
    \includegraphics[width=\textwidth]{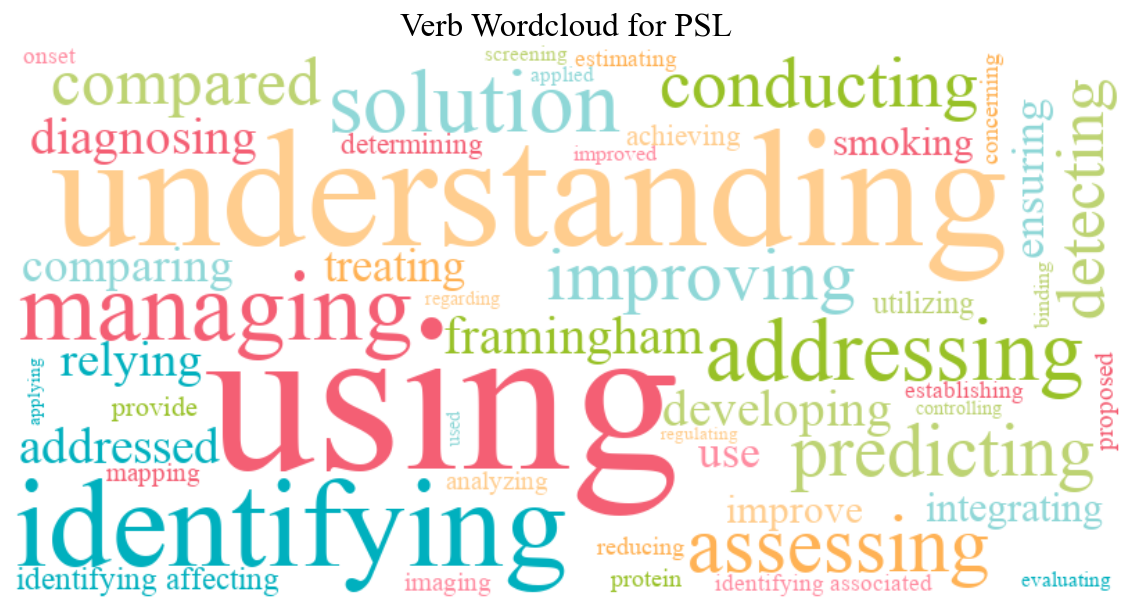}
    \caption{Verb}
  \end{subfigure}
  \hfill
  \begin{subfigure}[t]{0.23\textwidth}
    \includegraphics[width=\textwidth]{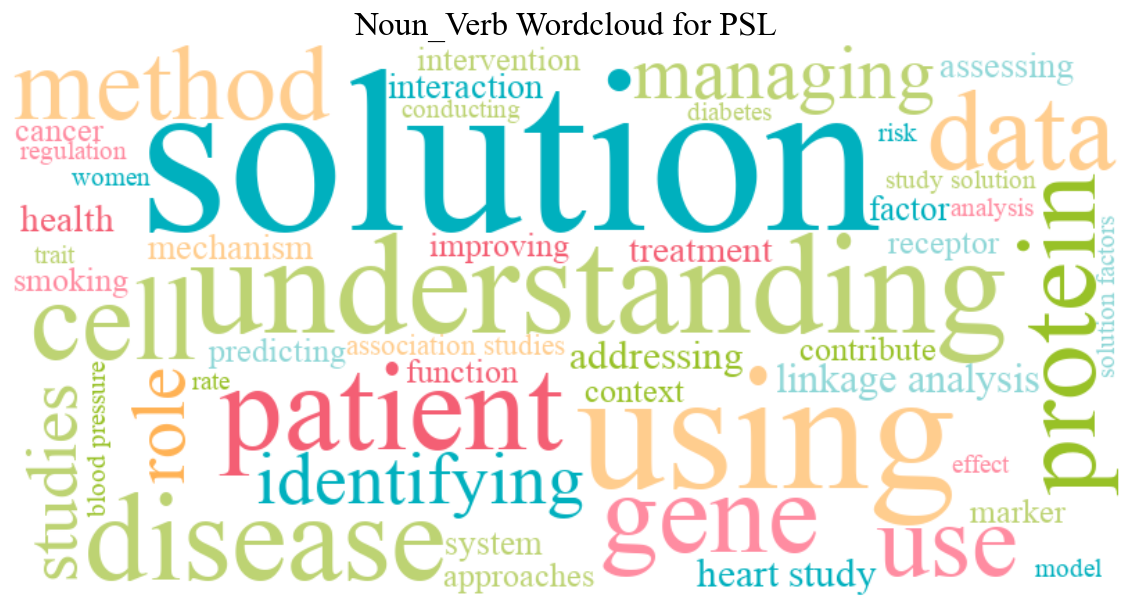}
    \caption{Noun \& Verb}
  \end{subfigure}
  \hfill
  \begin{subfigure}[t]{0.23\textwidth}
    \includegraphics[width=\textwidth]{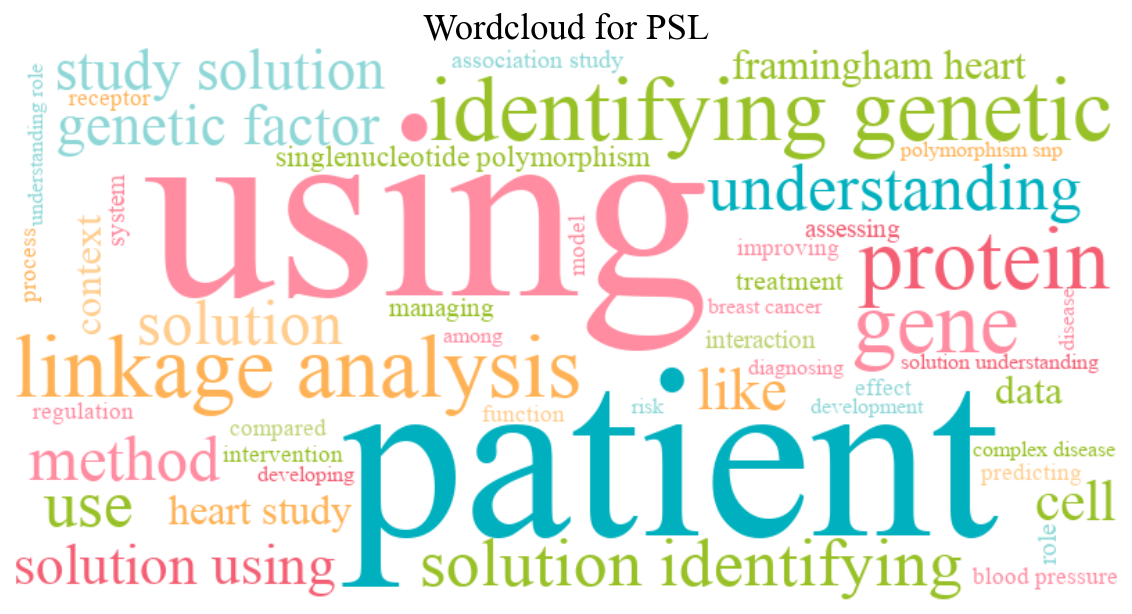}
    \caption{All Words}
  \end{subfigure}
  \caption{\textbf{Frequent words in \textbf{PSL-type} questions:} Showing terms like \textit{conducting}, \textit{addressing}, and \textit{diagnosing}, highlighting evaluation limitations and follow up aspects of solutions.}
\end{figure}

\begin{figure}[H]
  \centering
  \begin{subfigure}[t]{0.23\textwidth}
    \includegraphics[width=\textwidth]{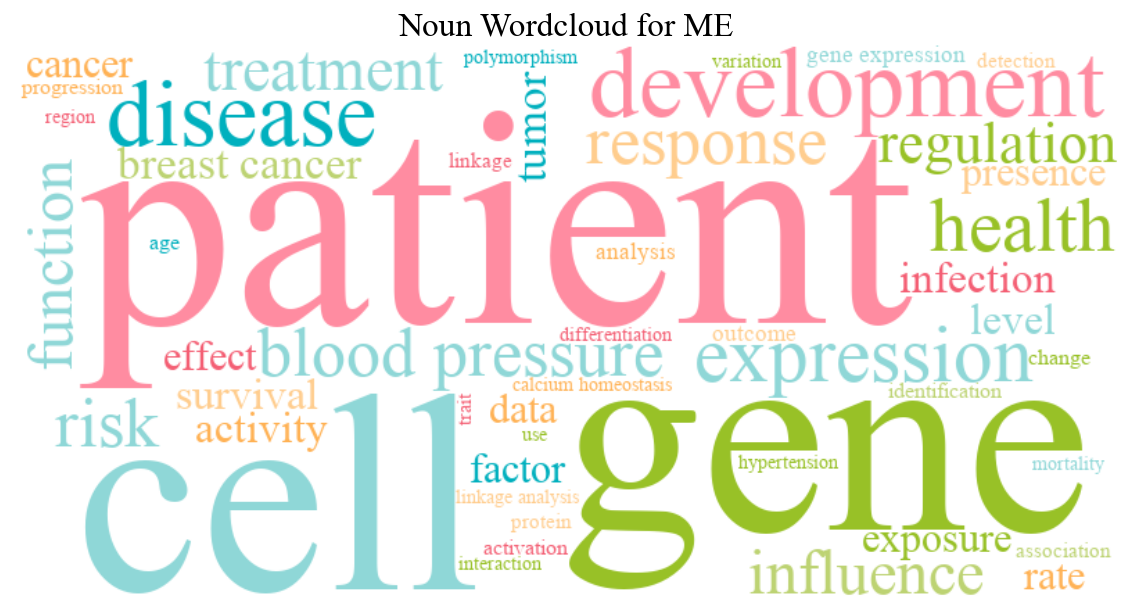}
    \caption{Noun}
  \end{subfigure}
  \hfill
  \begin{subfigure}[t]{0.23\textwidth}
    \includegraphics[width=\textwidth]{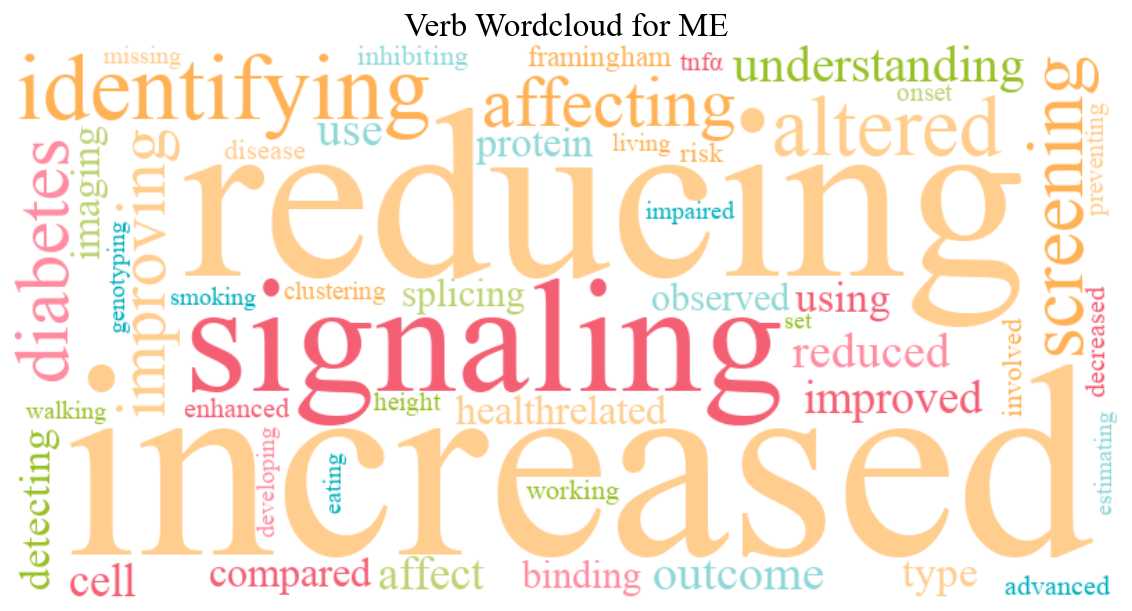}
    \caption{Verb}
  \end{subfigure}
  \hfill
  \begin{subfigure}[t]{0.23\textwidth}
    \includegraphics[width=\textwidth]{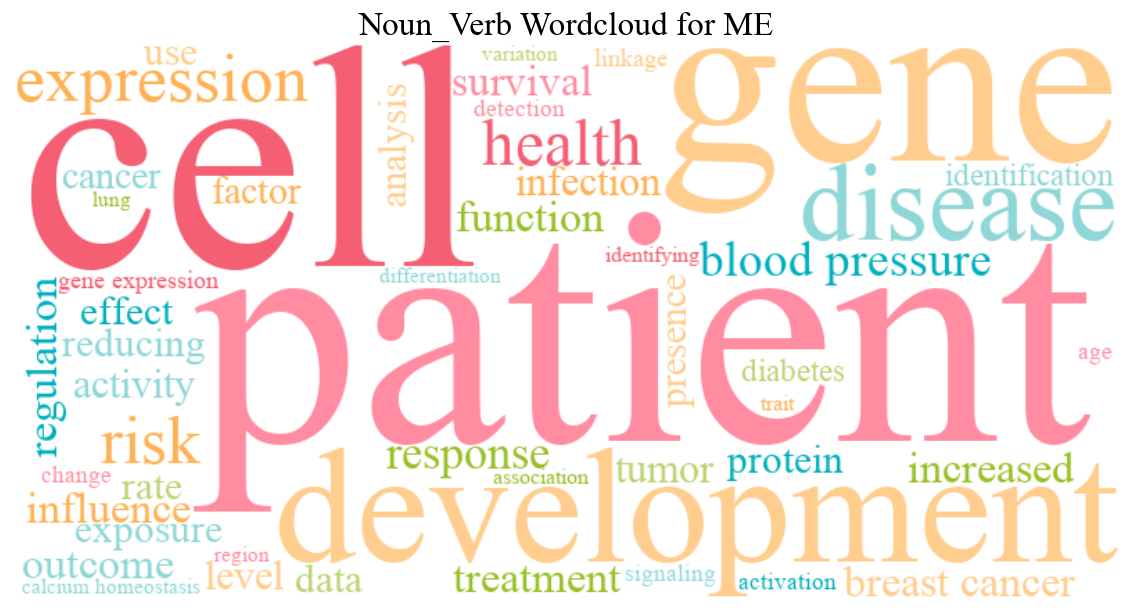}
    \caption{Noun \& Verb}
  \end{subfigure}
  \hfill
  \begin{subfigure}[t]{0.23\textwidth}
    \includegraphics[width=\textwidth]{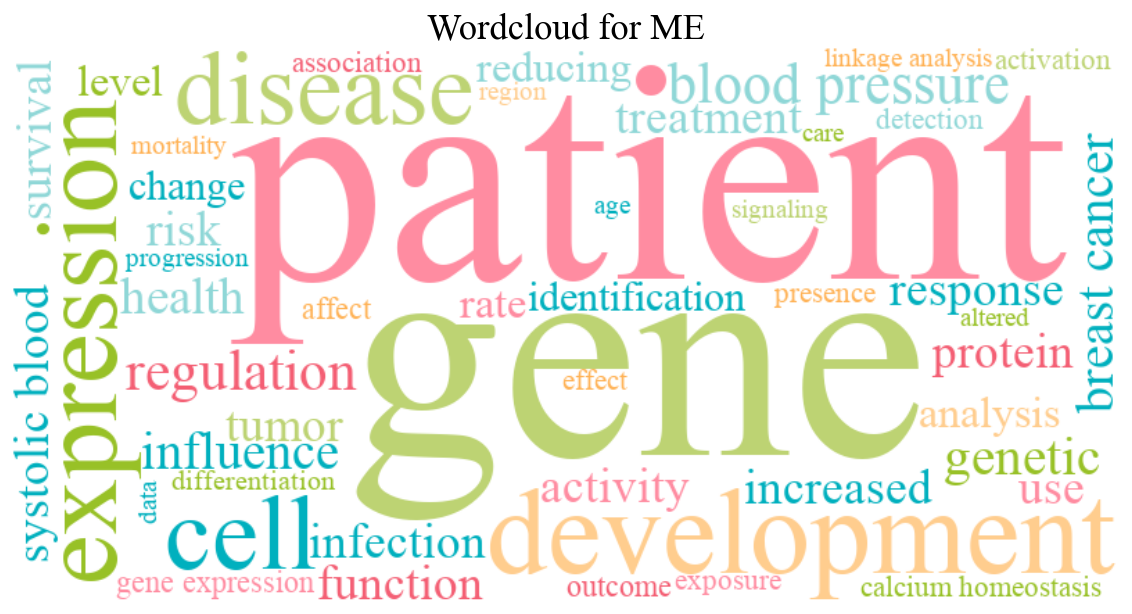}
    \caption{All Words}
  \end{subfigure}
  \caption{\textbf{Frequent words in \textbf{ME-type} questions:} Showing terms like \textit{increased}, \textit{reducing}, and \textit{signaling}, indicating a focus on causal mechanisms and biological effects.}
\end{figure}

\begin{figure}[H]
  \centering
  \begin{subfigure}[t]{0.23\textwidth}
    \includegraphics[width=\textwidth]{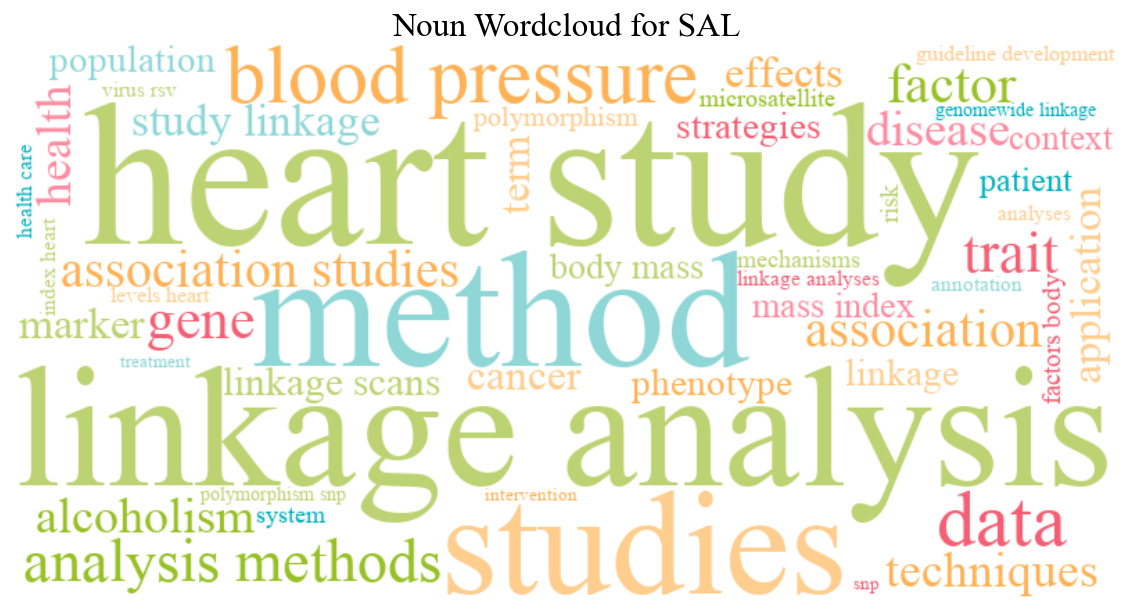}
    \caption{Noun}
  \end{subfigure}
  \hfill
  \begin{subfigure}[t]{0.23\textwidth}
    \includegraphics[width=\textwidth]{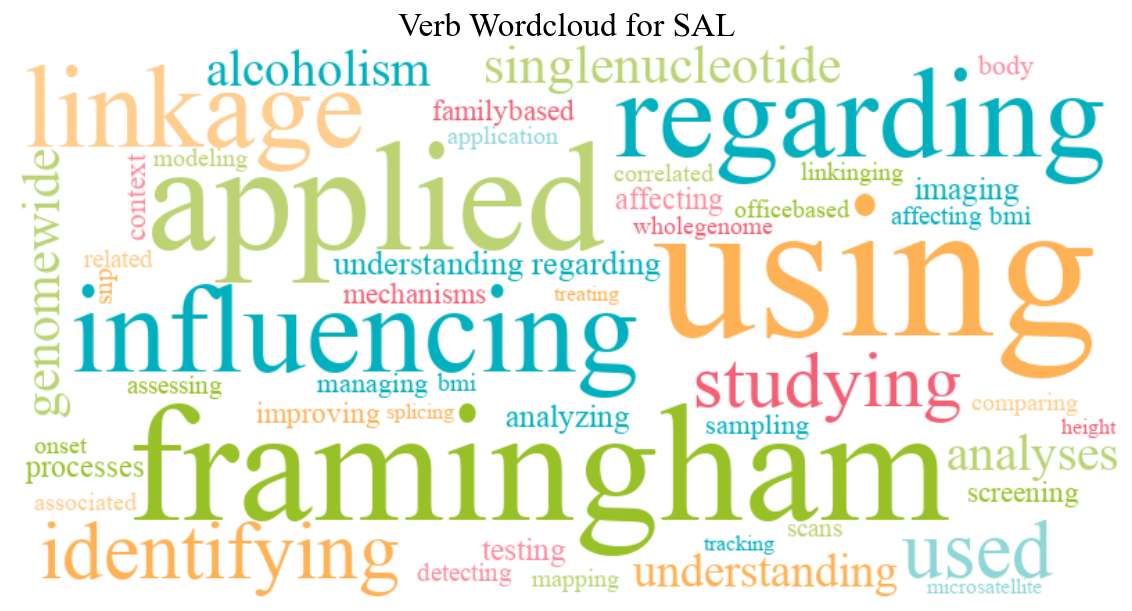}
    \caption{Verb}
  \end{subfigure}
  \hfill
  \begin{subfigure}[t]{0.23\textwidth}
    \includegraphics[width=\textwidth]{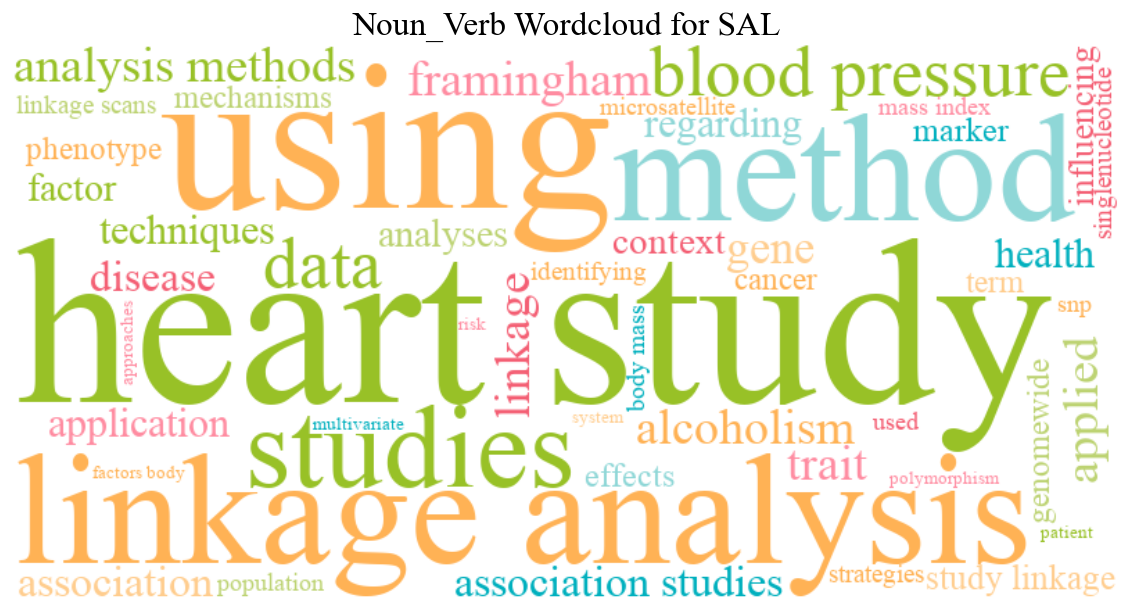}
    \caption{Noun \& Verb}
  \end{subfigure}
  \hfill
  \begin{subfigure}[t]{0.23\textwidth}
    \includegraphics[width=\textwidth]{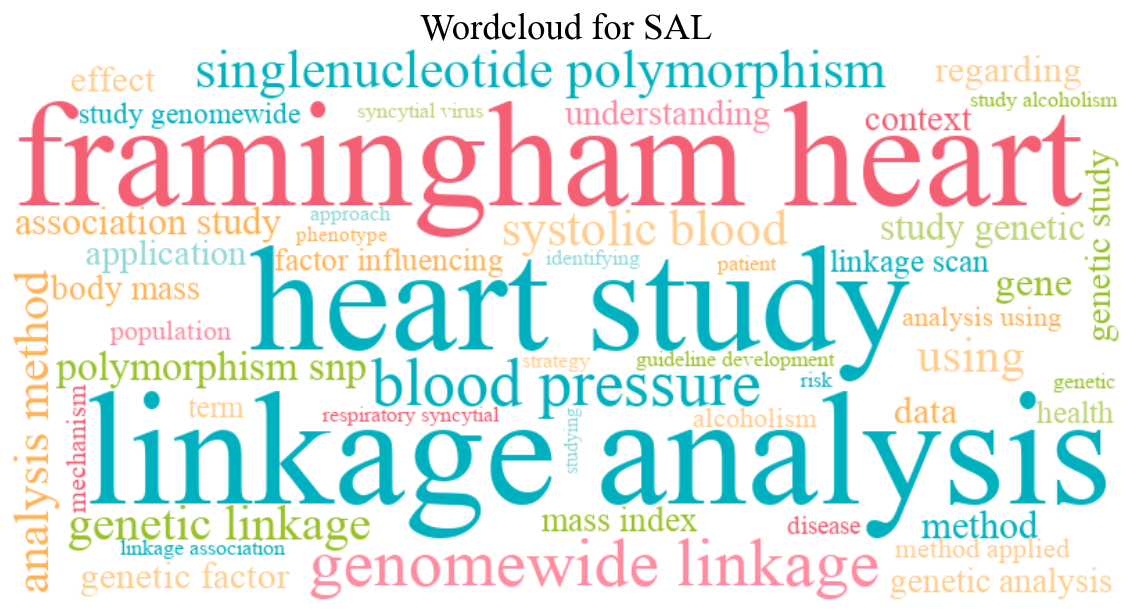}
    \caption{All Words}
  \end{subfigure}
  \caption{\textbf{Frequent words in \textbf{SAL-type} questions:} Showing terms like \textit{heart} and \textit{blood}, indicating strengths and limitations of  biological studies.}
\end{figure}

\label{appendix:paragraph}

\begin{figure}[H]
  \centering
  \begin{subfigure}[t]{0.23\textwidth}
    \includegraphics[width=\textwidth]{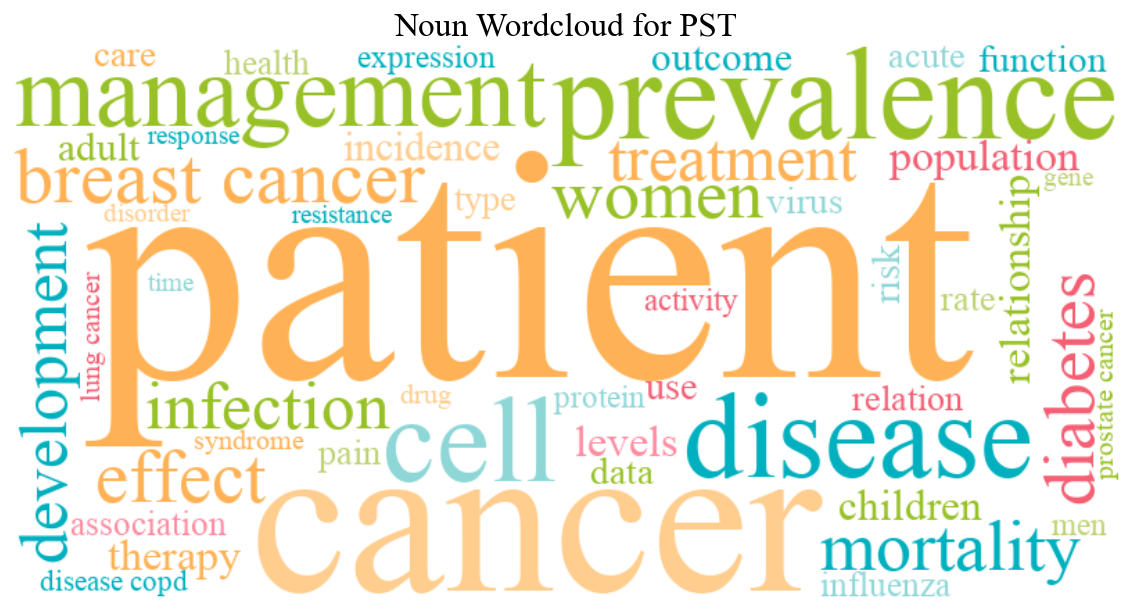}
    \caption{Noun}
  \end{subfigure}
  \hfill
  \begin{subfigure}[t]{0.23\textwidth}
    \includegraphics[width=\textwidth]{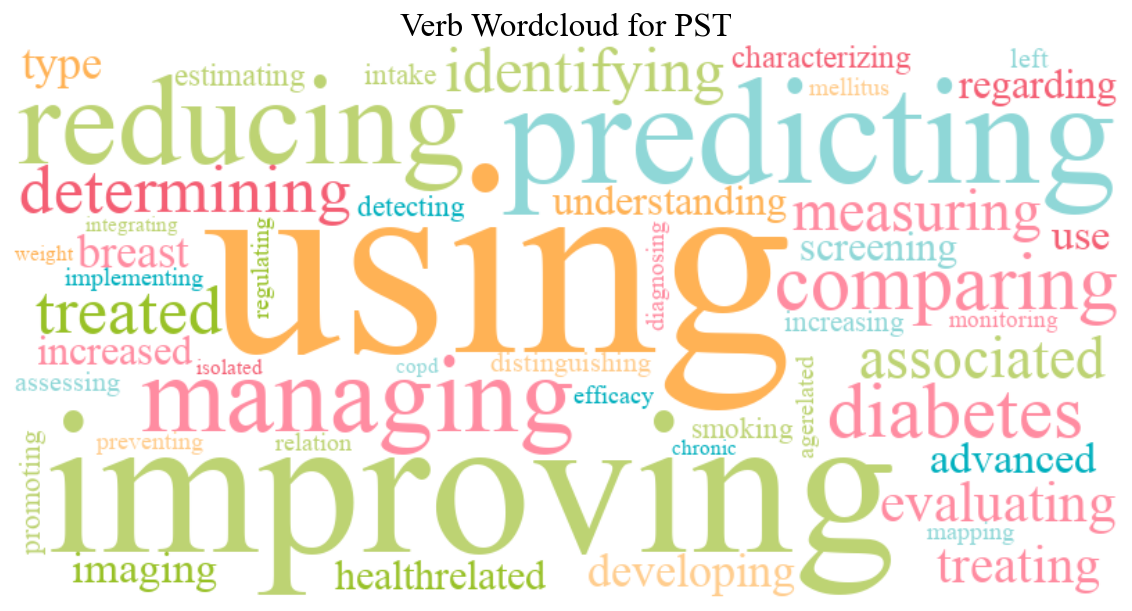}
    \caption{Verb}
  \end{subfigure}
  \hfill
  \begin{subfigure}[t]{0.23\textwidth}
    \includegraphics[width=\textwidth]{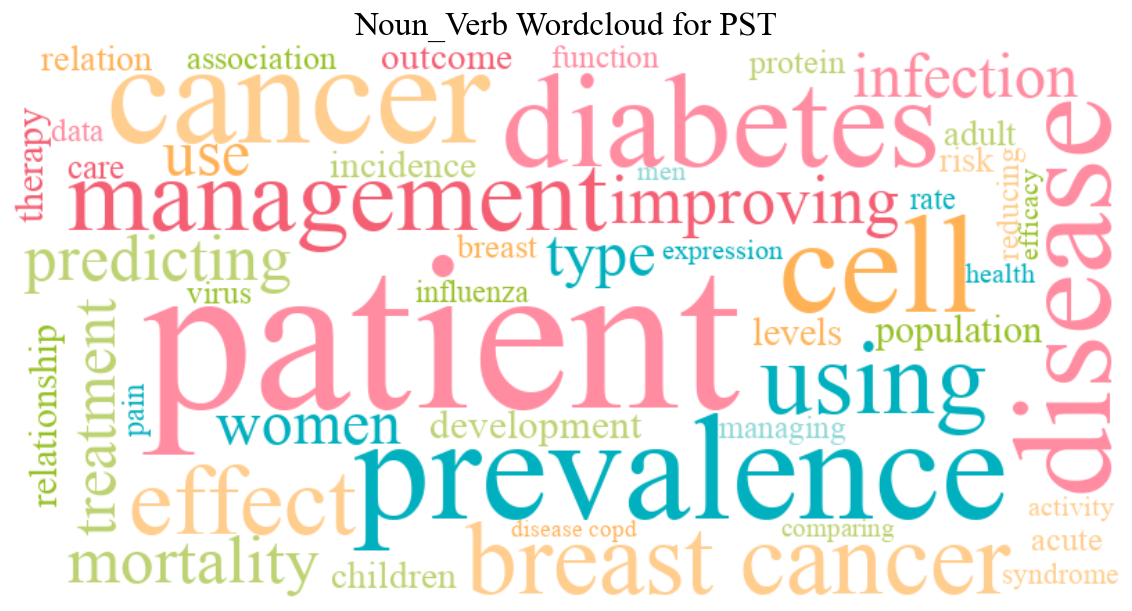}
    \caption{Noun \& Verb}
  \end{subfigure}
  \hfill
  \begin{subfigure}[t]{0.23\textwidth}
    \includegraphics[width=\textwidth]{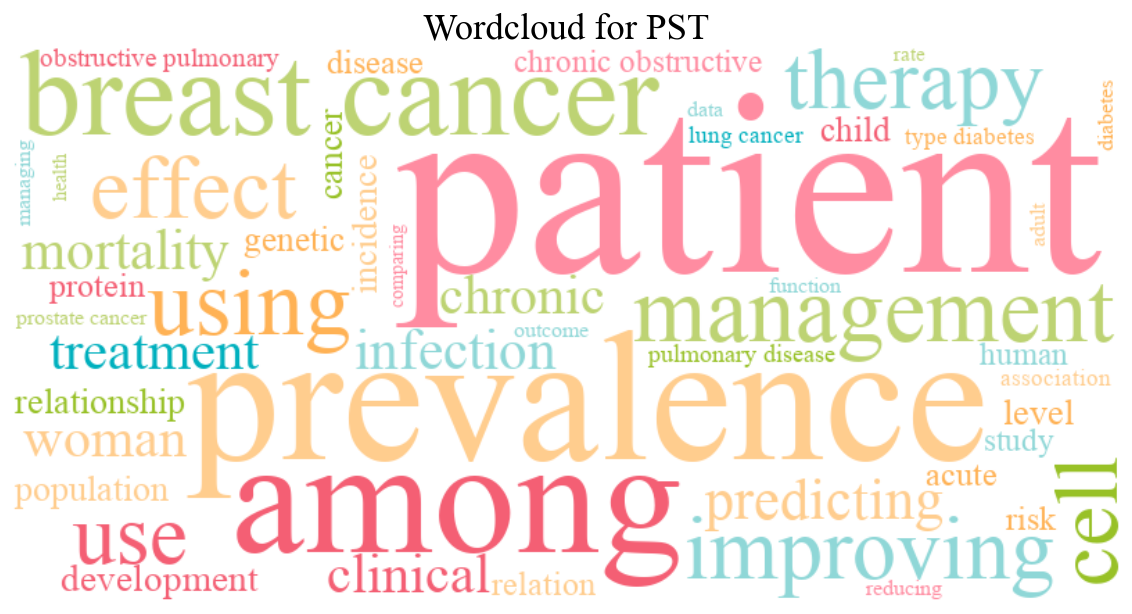}
    \caption{All Words}
  \end{subfigure}
  \caption{\textbf{Frequent words in \textbf{PST-type} questions:} Showing terms like \textit{predicting}, \textit{patient}, and \textit{infection}, highlighting challenges and strategies based on tabular data.}
\end{figure}

\begin{figure}[H]
  \centering
  \begin{subfigure}[t]{0.23\textwidth}
    \includegraphics[width=\textwidth]{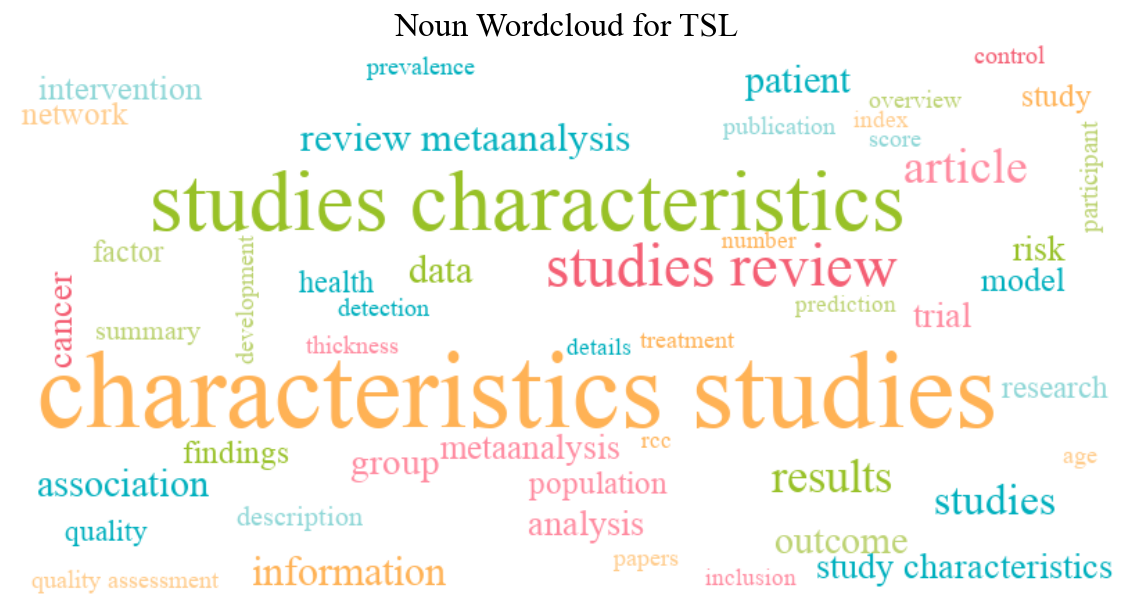}
    \caption{Noun}
  \end{subfigure}
  \hfill
  \begin{subfigure}[t]{0.23\textwidth}
    \includegraphics[width=\textwidth]{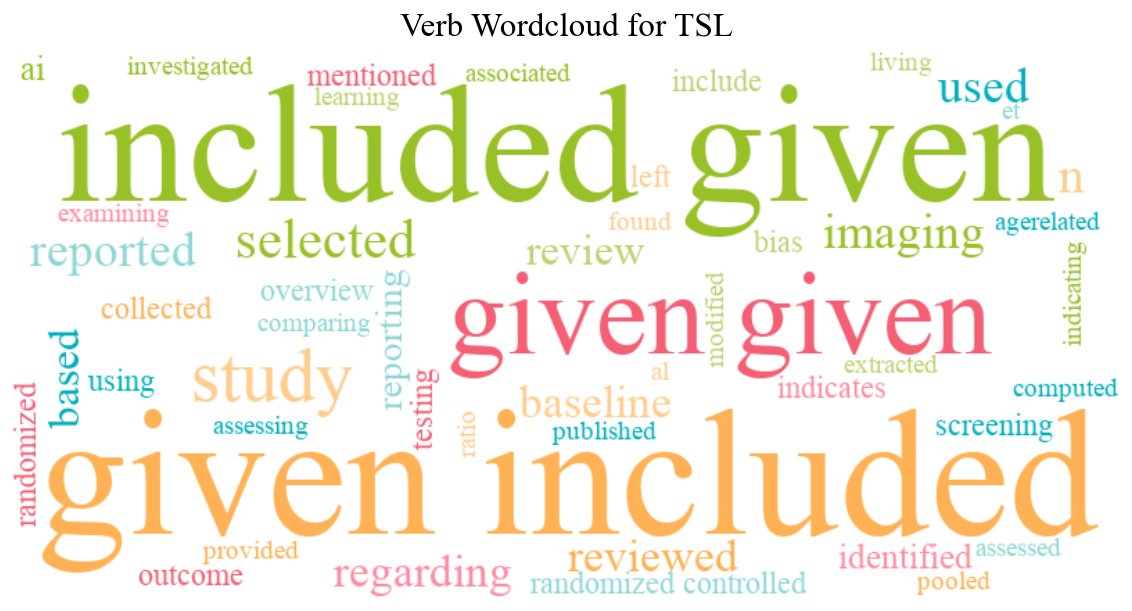}
    \caption{Verb}
  \end{subfigure}
  \hfill
  \begin{subfigure}[t]{0.23\textwidth}
    \includegraphics[width=\textwidth]{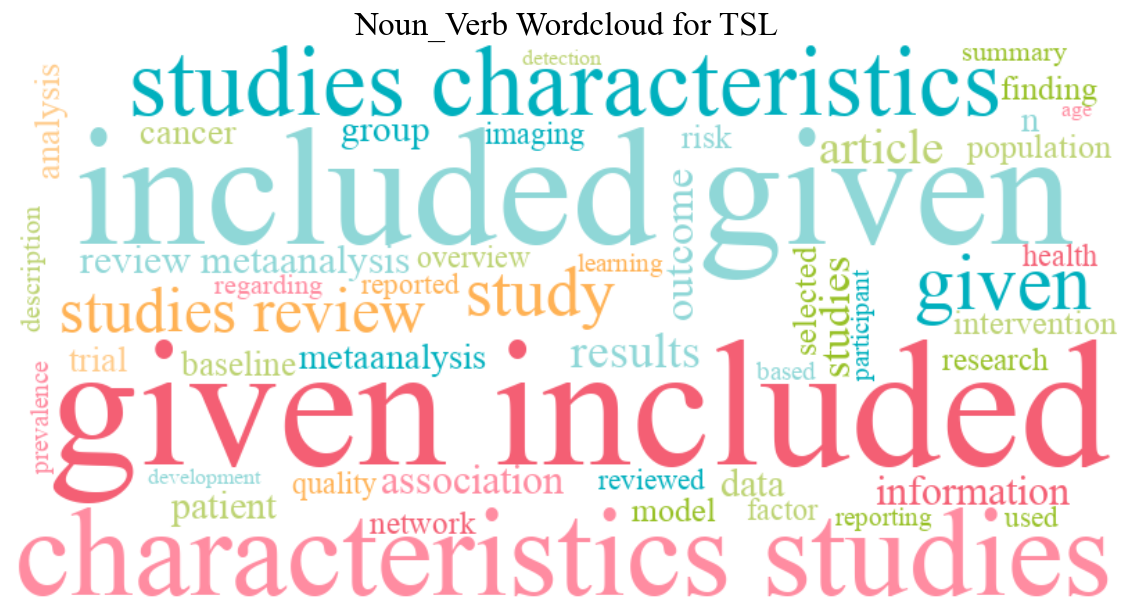}
    \caption{Noun \& Verb}
  \end{subfigure}
  \hfill
  \begin{subfigure}[t]{0.23\textwidth}
    \includegraphics[width=\textwidth]{Figures/wordcloud/Noun_Wordcloud_for_reference_limitation_.png}
    \caption{All Words}
  \end{subfigure}
  \caption{\textbf{Frequent words in \textbf{TSL-type} questions:} Showing terms like \textit{given}, \textit{included}, and \textit{characteristics}, reflecting their deterministic, caption-based generation with limited lexical variety.}
\end{figure}

\subsection{Answer Dispersion}
\label{appendix:answer_dispersion}



\begin{figure}[t]
\centering
\includegraphics[width=\linewidth]{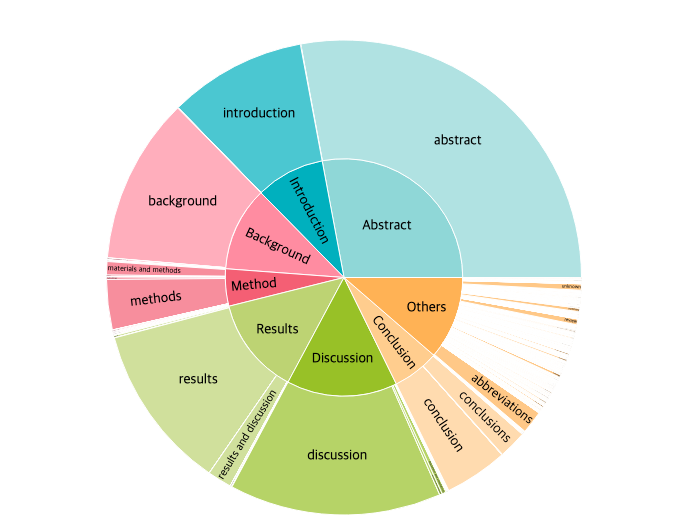}
\caption{\textbf{Section Mapping Results.}}
\label{fig:section_mapping_chart}
\end{figure}

\paragraph{Super-Section Mapping Analysis}
To facilitate section-level analysis, we standardized a diverse set of original section titles into 8 unified Super-Section tags. This mapping aimed to reduce the variation caused by inconsistent section titles (e.g., ``Materials and Methods", ``Subjects and Methods", ``Methodology") and to improve the interpretability of section-based statistics. The results of standardization are described in Table~\ref{tab:super_section_details} and Figure~\ref{fig:section_mapping_chart}.

\begin{table}[t]
\centering
\small
\begin{tabular}{M{2cm}|M{5.3cm}}
\hline
\textbf{Super-Section} & \textbf{Detail Section Title} \\ \hline
Abstract & \textbf{abstract}, etc. \\ \hline
Introduction & \textbf{introduction}, etc.\\ \hline
Background & \textbf{background}, \textbf{background and recent developments}, background and overview, etc.\\ \hline
Method & \textbf{method(s)}, material(s) and method(s), method and implementation, etc. \\ \hline
Results & \textbf{result(s)}, \textbf{results and discussion}, implementation and results, result analysis, etc. \\ \hline
Discussion & \textbf{discussion}(s), \textbf{discussion and conclusions}, limitations, summary and discussion, etc \\ \hline
Conclusion & \textbf{conclusion(s)}, conclusions and future directions, conclusion and perspectives, etc.\\ \hline
Others & case report, \textbf{review}, dataset and evaluation strategy, \textbf{abbreviations}, etc.\\ \hline
\end{tabular}
\caption{\textbf{Detailed results of Super-Section Mapping}, with frequent section titles bolded.}
\label{tab:super_section_details}
\end{table}

\begin{figure}[!h]
  \centering
  \begin{subfigure}[t]{0.24\columnwidth}
    \includegraphics[width=\linewidth,height=1.6cm]{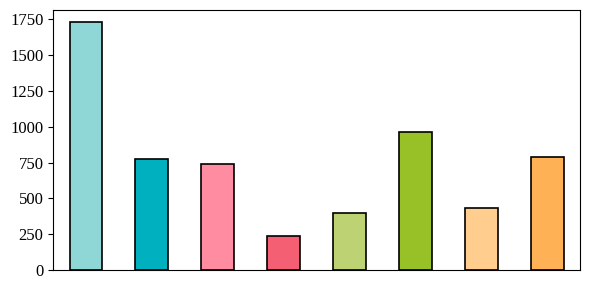}
    \vspace{-0.6cm}
    \caption{PS}
    \label{fig:sec_dist_ps}
  \end{subfigure}
  \begin{subfigure}[t]{0.24\columnwidth}
    \includegraphics[width=\linewidth,height=1.6cm]{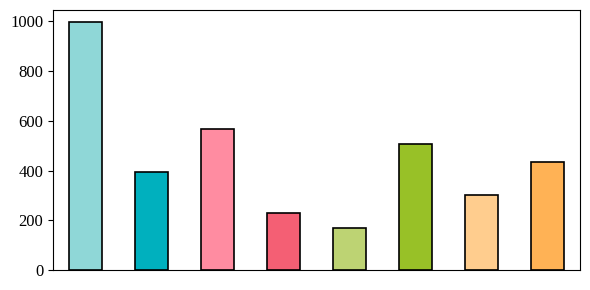}
    \vspace{-0.6cm}
    \caption{PSM}
    \label{fig:sec_dist_psm}
  \end{subfigure}
  \begin{subfigure}[t]{0.24\columnwidth}
    \includegraphics[width=\linewidth,height=1.6cm]{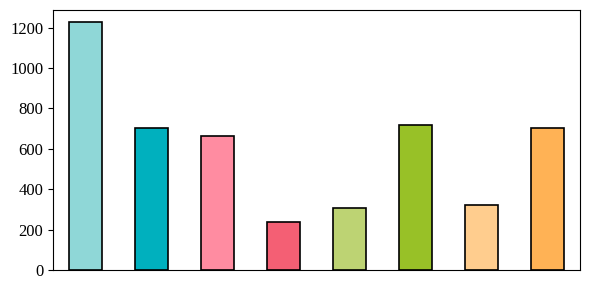}
    \vspace{-0.6cm}
    \caption{PSR}
    \label{fig:sec_dist_psr}
  \end{subfigure}
  \begin{subfigure}[t]{0.24\columnwidth}
    \includegraphics[width=\linewidth,height=1.6cm]{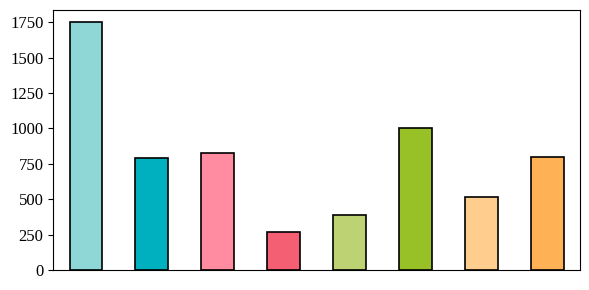}
    \vspace{-0.6cm}
    \caption{PSL}
    \label{fig:sec_dist_psl}
  \end{subfigure}

  \vspace{0.1cm}
  
  \begin{subfigure}[t]{0.24\columnwidth}
    \includegraphics[width=\linewidth,height=1.6cm]{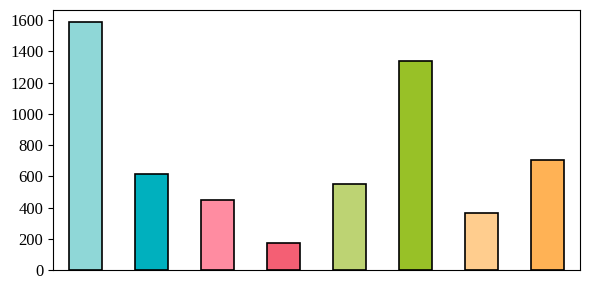}
    \vspace{-0.6cm}
    \caption{ME}
    \label{fig:sec_dist_me}
  \end{subfigure}
  \begin{subfigure}[t]{0.24\columnwidth}
    \includegraphics[width=\linewidth,height=1.6cm]{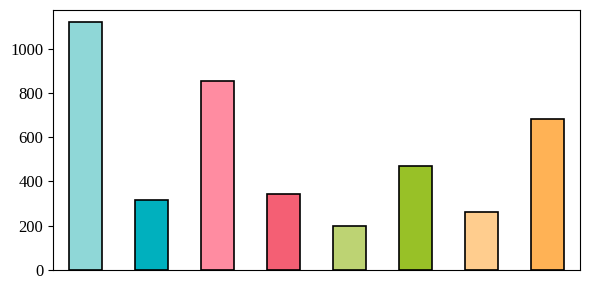}
    \vspace{-0.6cm}
    \caption{SAL}
    \label{fig:sec_dist_sal}
  \end{subfigure}
  \begin{subfigure}[t]{0.24\columnwidth}
    \includegraphics[width=\linewidth,height=1.6cm]{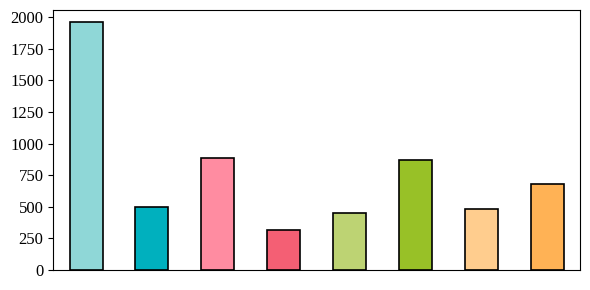}
    \vspace{-0.6cm}
    \caption{FF}
    \label{fig:sec_dist_ff}
  \end{subfigure}
  \begin{subfigure}[t]{0.24\columnwidth}
    \includegraphics[width=\linewidth,height=1.6cm]{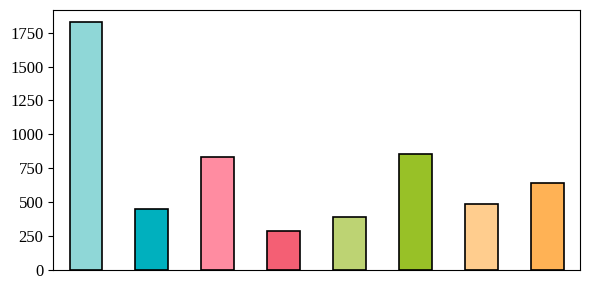}
    \vspace{-0.6cm}
    \caption{SS}
    \label{fig:sec_dist_ss}
  \end{subfigure}
  \begin{subfigure}[t]{0.24\columnwidth}
    \includegraphics[width=\linewidth,height=1.6cm]{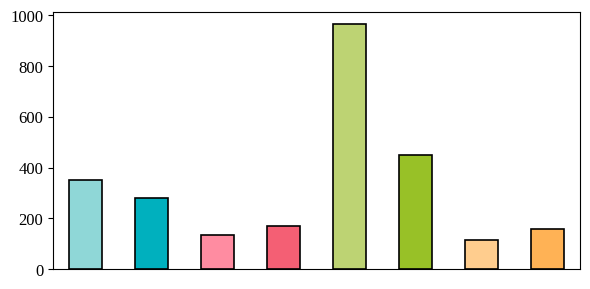}
    \vspace{-0.6cm}
    \caption{PST}
    \label{fig:sec_dist_pst}
  \end{subfigure}
  \begin{subfigure}[t]{0.24\columnwidth}
    \includegraphics[width=\linewidth,height=1.6cm]{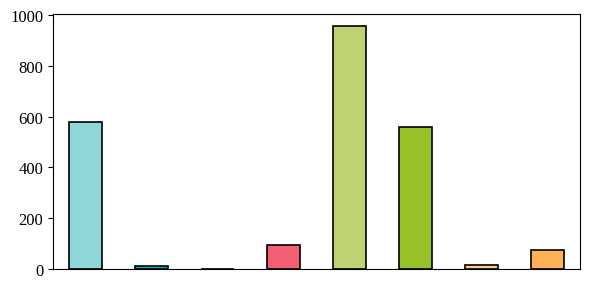}
    \vspace{-0.6cm}
    \caption{TSL}
    \label{fig:sec_dist_tsl}
  \end{subfigure}
  \begin{subfigure}[t]{0.24\columnwidth}
    \includegraphics[width=\linewidth,height=1.6cm]{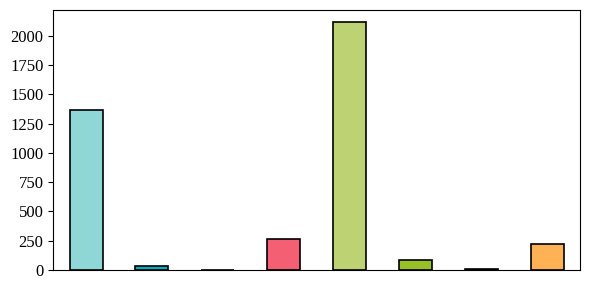}
    \vspace{-0.6cm}
    \caption{TS}
    \label{fig:sec_dist_ts}
  \end{subfigure}
  \begin{subfigure}[t]{\columnwidth}
    \includegraphics[width=\linewidth,height=1cm]{Figures/legend_sections_white_small.png}
  \end{subfigure}
  \vspace{-0.7cm}
  \caption{\textbf{Section-level distribution of answer spans across the 11 Q types.}}
  \label{fig:super-section-dist-each-qtype}
  \vspace{-0.8em}
\end{figure}

\paragraph{Section-level Answer Dispersion}
Figure~\ref{fig:super-section-dist-each-qtype} presents the distribution of answer snippet across QA types. Paragraph-Oriented types predominantly retrieved answers from the Abstract, aligning with their fan-hop retrieval strategy. In contrast, Table-Oriented types mostly extracted snippets from the Results and Discussion sections, reflecting the presence of structured, interpretive content found in them. These patterns highlight that each Q type draws from distinct sections of the document, underscoring the need for models to understand section-level semantics when reasoning over scientific papers.

\paragraph{Table Oriented Section Analysis}
\label{sec:table_oriented_section_anaylsis}
We analyzed the origin of tables (Table~\ref{tab:table_section_count}) and found that most were extracted from the \textit{Results} section. This supports our decision to include only documents with at least one table in result section for PST. TSL, TS tables also mainly came from Results, reflecting their common role in summarizing outcomes across studies. These patterns show that the dataset aligns with typical scientific reporting, where key findings are often presented in results tables.

\renewcommand{\arraystretch}{1.1}
\begin{table}[!h]
\small
\centering
\begin{tabular}{M{2cm}|M{1.2cm}|M{1.2cm}|M{1.2cm}}
\hline
\textbf{Section} & \textbf{PST} & \textbf{TS} & \textbf{TSL} \\
\hline
Abstract      & 5   & 3   & 3   \\

Introduction  & 1   &16   & 5   \\

Background    & 1   & 2   & 0   \\

Method        &33   &132  &47   \\

Results       &\textbf{830} &\textbf{1057} &\textbf{478} \\

Discussion    & 1   &41   & 4   \\

Conclusion    & 0   & 2   & 1   \\

Others        & 3   &112  &35   \\
\hline
\end{tabular}
\caption{\textbf{Distribution of Table Origins Across Table-Oriented types} (PST, TS, TSL) with Section Mapping}
\label{tab:table_section_count}
\vspace{-0.8em}
\end{table}

\paragraph{Structural Distance Analysis}
\begin{figure*}[!h]
    \centering
    \includegraphics[width=\textwidth]{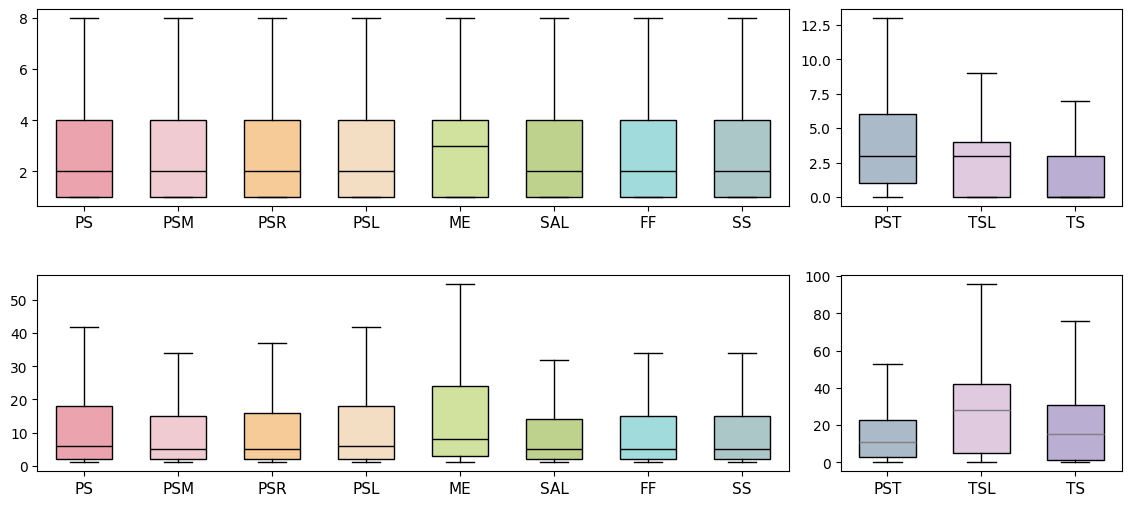}
    \caption{\textbf{Structural evidence distribution across question concept types.} We show section-level (top) and paragraph-level (bottom) positions of answer snippets for each concept. Fan-hop questions exhibit broadly distributed positions within documents, indicating global aggregation, whereas chain-hop questions show progressive structural between snippets, reflecting ordered multi-step reason-chaining processes. Notably, PSM, ME tend to retrieve evidence from relatively localized yet deeper structural regions, while PST, TSL, TS span wider structural ranges, often bridging descriptive and analytical sections. Overall, the diverse structural ranges covered across question concept types highlight the heterogeneity of reasoning difficulty and retrieval behaviors encoded in our taxonomy.}
    \label{fig:distance_by_qtypes}
\end{figure*}
We quantify how answer snippets are distributed across the document structure to verify that our 11 question concepts yield diverse, non-position-biased snippet layouts (Figure~\ref{fig:distance_by_qtypes}.
While fan-hop and chain-hop characterize two complementary reasoning patterns based on answer snippet aggregation and dependency structure, the underlying 11 question concept types span these patterns and induce distinct structural retrieval and traversal behaviors.
For \textbf{chain-hop}, we concatenate the two documents into a single sequence to account for cross-document distances, and compute the snippet-to-snippet \emph{distance} between consecutive steps based on snippet's position. We report distances at the element-wise and as an instance-wise average.
All box plots retain only values within the interquartile range (IQR; 25th--75th percentile).

\begin{figure*}[!t]
    \centering
    \includegraphics[width=\textwidth]{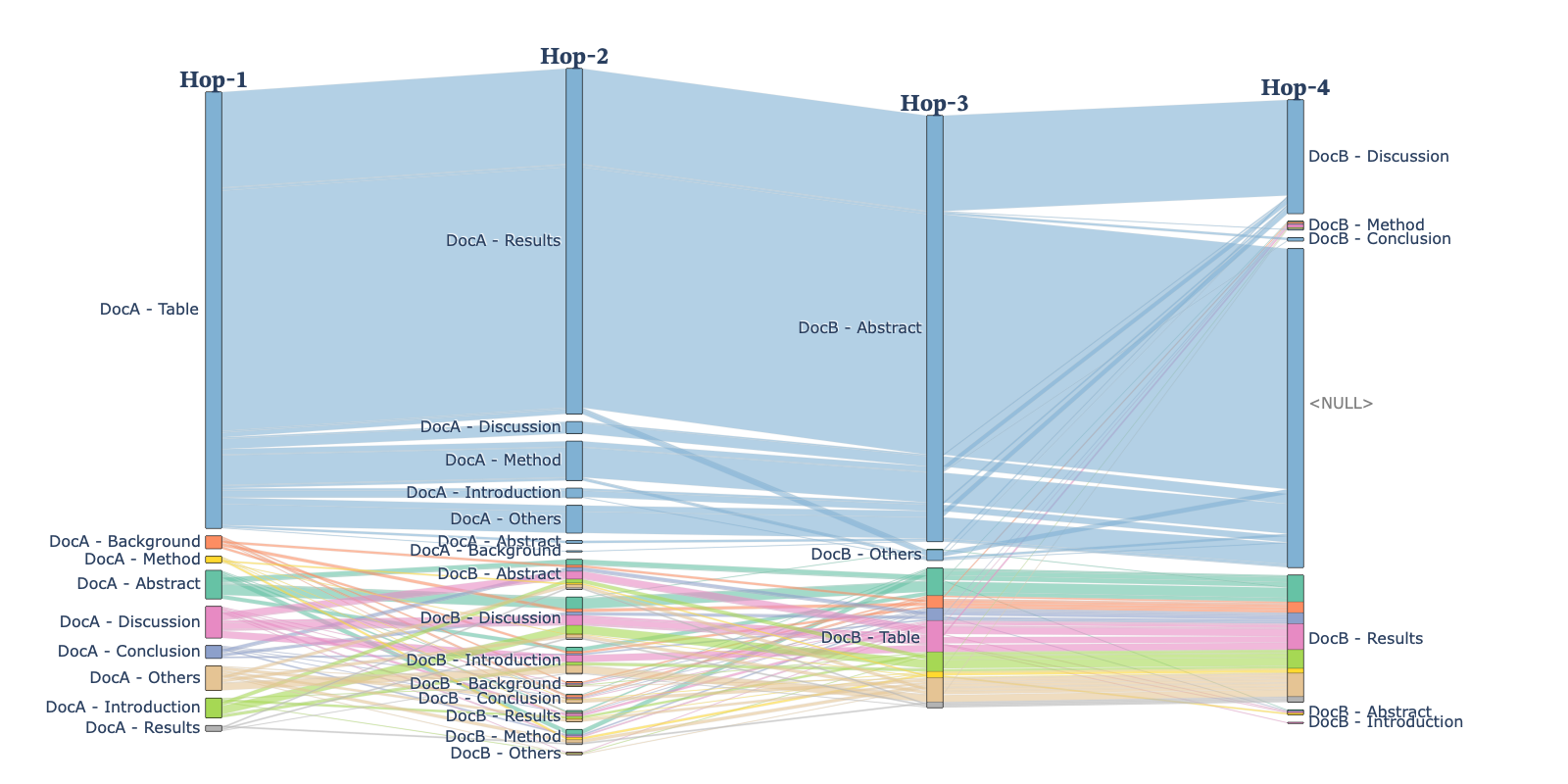}
    \caption{\textbf{Ground-truth chain-hop transitions across documents and sections (Sankey Diagram).} Each node denotes the document-section location of an answer snippet at each hop, and links represent hop-to-hop transitions in the ground-truth chain. Tables appear not only as endpoints but also as intermediate bridges. This diversity highlights the structural complexity of multi-document reasoning required by DocHop-QA}
    \label{fig:chainhop_sankey}
\end{figure*}

\paragraph{Chain-Hop Transition Analysis}
\label{appendix:cahin_hop_structure_analysis}
To analyze the structural diversity of chain-hop reasoning in DocHop-QA, we visualize ground-truth hop transitions across documents and document sections using a Sankey-style representation shown in Figure~\ref{fig:chainhop_sankey}. This analysis reveals that chain-hop questions exhibit diverse hop patterns, spanning document sections such as abstracts, background, methods, results, and conclusions, often across different documents. Notably, table entities appear not only as terminal answer evidence but also as intermediate nodes within reasoning chains, indicating that tables frequently serve as both sources and bridges of information. \\ These observations highlight that DocHop-QA chain-hop questions require models to perform structured, dependency-sensitive reasoning across heterogeneous document components, rather than relying on linear or section-local inference. The diversity of hop transitions underscores the need for multi-document understanding that integrates both textual and tabular evidence in a coherent reasoning process.

\section{Additional Experiments and Results}
\label{appendix:detailed_experiments_and_results}
\subsection{General Preprocessing}
\label{appendix:training_setup}
All documents are converted from PDF to XML and image formats. OCR and table detection are applied to extract layout-aware content for entity extraction tasks. We use a 70/10/20 train/dev/test split, ensuring that each document appears in only one split to prevent information leakage.

At the document level, we first extracted structured entities, including titles, abstracts, paragraphs, figures, and tables, from the XML version. For visual grounding, we align these XML entities with their corresponding positions in the PDF using the PaddleX OCR toolkit, and save each page of the PDF as an image. This enables us to generate both text-based and image-based inputs across tasks.

For entity-based tasks, at the sample level, we combined all entities from the relevant context documents into a unified list, assigning global indices. For instance, if Document 1 has 10 entities and Document 2 has 32, entities are indexed sequentially from [Entity 0] (e.g., \textless doc1\_title\textgreater) to [Entity 41] (e.g., \textless doc2\_paragraph20\textgreater). Answer labels are represented as index lists referencing this combined sequence. For models that are limited to one input image, such as LayoutLMv3, all pages from the context documents are merged into a single image and resized to comply with the model's input constraints.

For fine-tuning, all classification-based models are trained with binary cross-entropy loss with the AdamW optimizer for up to 50 epochs. Early stopping is applied with a patience of either 5 or 10 epochs, based on validation performance measured by sample-level recall. Training of open-sourced models is conducted locally using a single NVIDIA A100 GPU. Close-sourced model inference are executed via API calls. 

\vspace{-0.5em}
\subsection{Additional Task Descriptions}
\paragraph{Task 1: Generative Reasoning}
We test generative LLMs under three prompting strategies (question-only, zero-shot, one-shot) and three input types (text-only, image-only, text+image). Models capable of full-document ingestion, such as Gemini-2.5-Flash, Gemini-3-Pro, and GPT-4o, are given full XML and PDF files. Generated answers are evaluated with BLEU and ROUGE (1, 2, L, Lsum) against synthesized gold answers from reference context snippets using GPT-4.1.

For each QA instance, we used GPT-4.1 to synthesize a reference answer from the selected supporting evidence snippets produced by the answer-matching stage. The generation prompt explicitly instructed GPT-4.1 to answer the question using only the information contained in the provided evidence snippets, without relying on external knowledge or introducing unsupported information. Since DocHop-QA questions often contain multiple reasoning components, the model was additionally instructed to integrate the complementary evidence across snippets into a coherent answer that addresses the complete question rather than independently summarizing individual snippets.
We subsequently applied manual validation to the generated QA instances. Three authors cross-checked the question, selected evidence snippets, and synthesized reference answer to verify that (1) the answer was supported by the provided evidence, (2) the selected snippets collectively contained sufficient information to address the required components of the question, (3) the synthesized answer did not introduce claims inconsistent with or unsupported by the evidence, and (4) the question, evidence, and answer remained semantically consistent as a complete QA instance. We did not require a separate one-to-one annotation mapping every atomic statement in the reference answer to an individual snippet; rather, factual grounding was verified against the complete set of selected supporting snippets. QA instances that failed these grounding or quality checks were removed through our filtering pipeline.
This validation complements the conservative evidence-selection procedure in Appendix~\ref{appendix:detailed_answer_matching}, where instances with insufficiently aligned supporting snippets are filtered before gold-answer synthesis. The resulting gold answer should therefore be interpreted as an evidence-grounded reference synthesis, rather than the only valid answer to the question. Alternative answers may express the same supported information using different wording or organization, which also motivates our complementary human answerability analysis in Table~\ref{tab:human_answerability}.
\vspace{-0.7em}
\paragraph{Task 2: Structured Generative Answering}
This task uses generative models InternVL2-4b, Qwen2.5-VL-7B-Instruct, and GPT-4o to produce lists of entity indices. Inputs include labeled XML entities and document images. Prompts follow  zero-shot or one-shot formats. To avoid memory issues, we cap the number of entities and pages. Evaluation includes multi-label accuracy (Belong, Contain, and Overlap) and sample-level F1 based on spatial containment.
\vspace{-0.7em}
\paragraph{Task 3: BBox Entity Index Extraction}
Formulated as a discriminative, multi-label classification task, this evaluates how well models can identify the relevant answer-contained bounding boxes (e.g., paragraphs, figures, tables) in OCR document layouts. We first map each XML entity to one or more OCR-detected bounding boxes, which are compiled into a single 1000x1000 image per document set. We use two layout-aware vision-language models, LayoutLMv3 \cite{10.1145/3503161.3548112} and LayoutXLM \cite{Xu_2020}, that receive bounding box coordinates, associated OCR texts, and document images, and predict relevant bounding box indices. Entities that cannot be matched through OCR extraction are marked with index 0. The evaluation mirrors the metrics used in Task 2.

\vspace{-0.5em}

\begin{figure*}[t]
\centering

\begin{subfigure}[t]{0.32\textwidth}
    \centering
    \includegraphics[width=\linewidth]{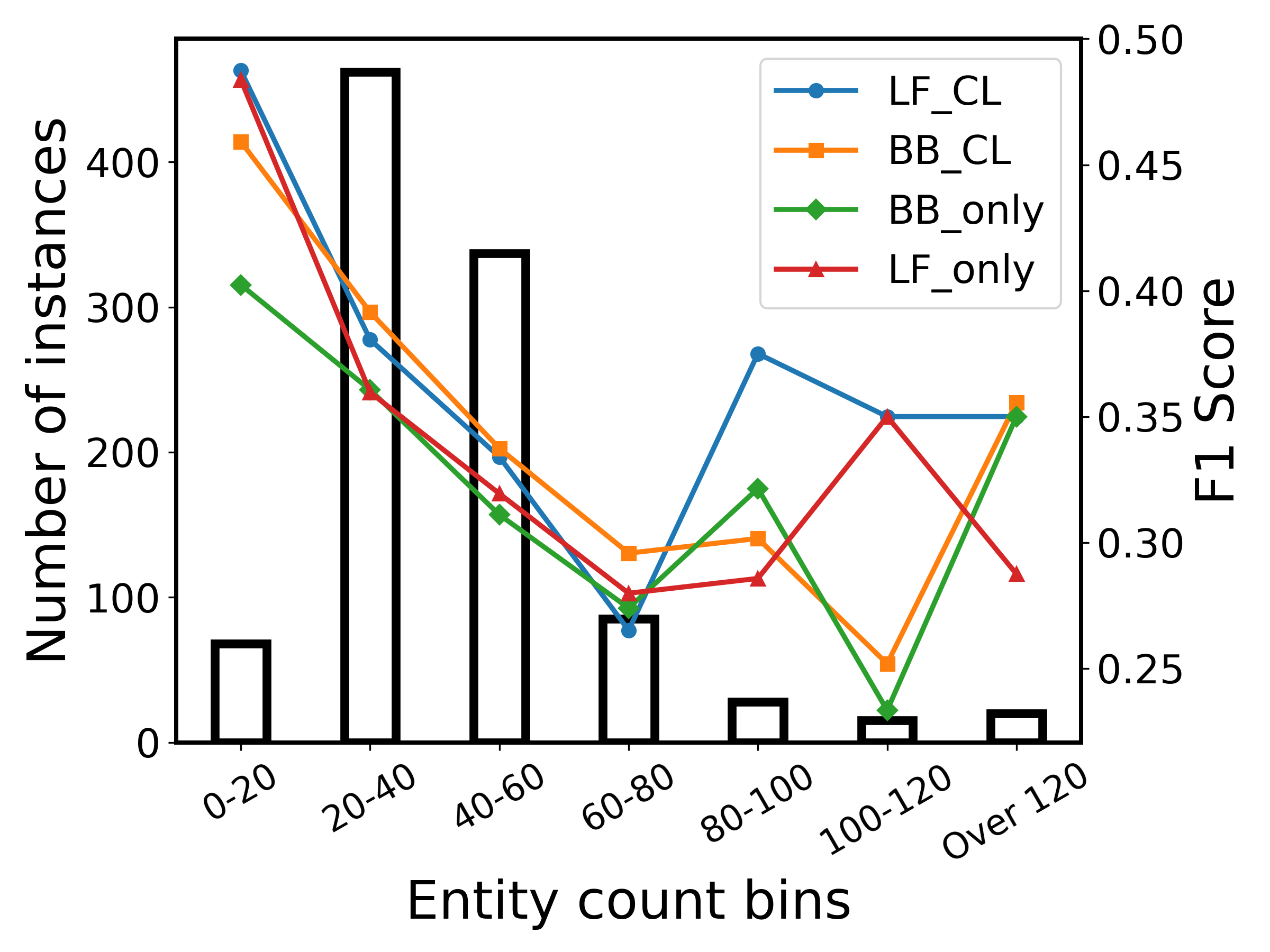}
    \caption{Paragraph \& Single}
    \label{fig:entcnt_f1_para_single_all}
\end{subfigure}
\hfill
\begin{subfigure}[t]{0.32\textwidth}
    \centering
    \includegraphics[width=\linewidth]{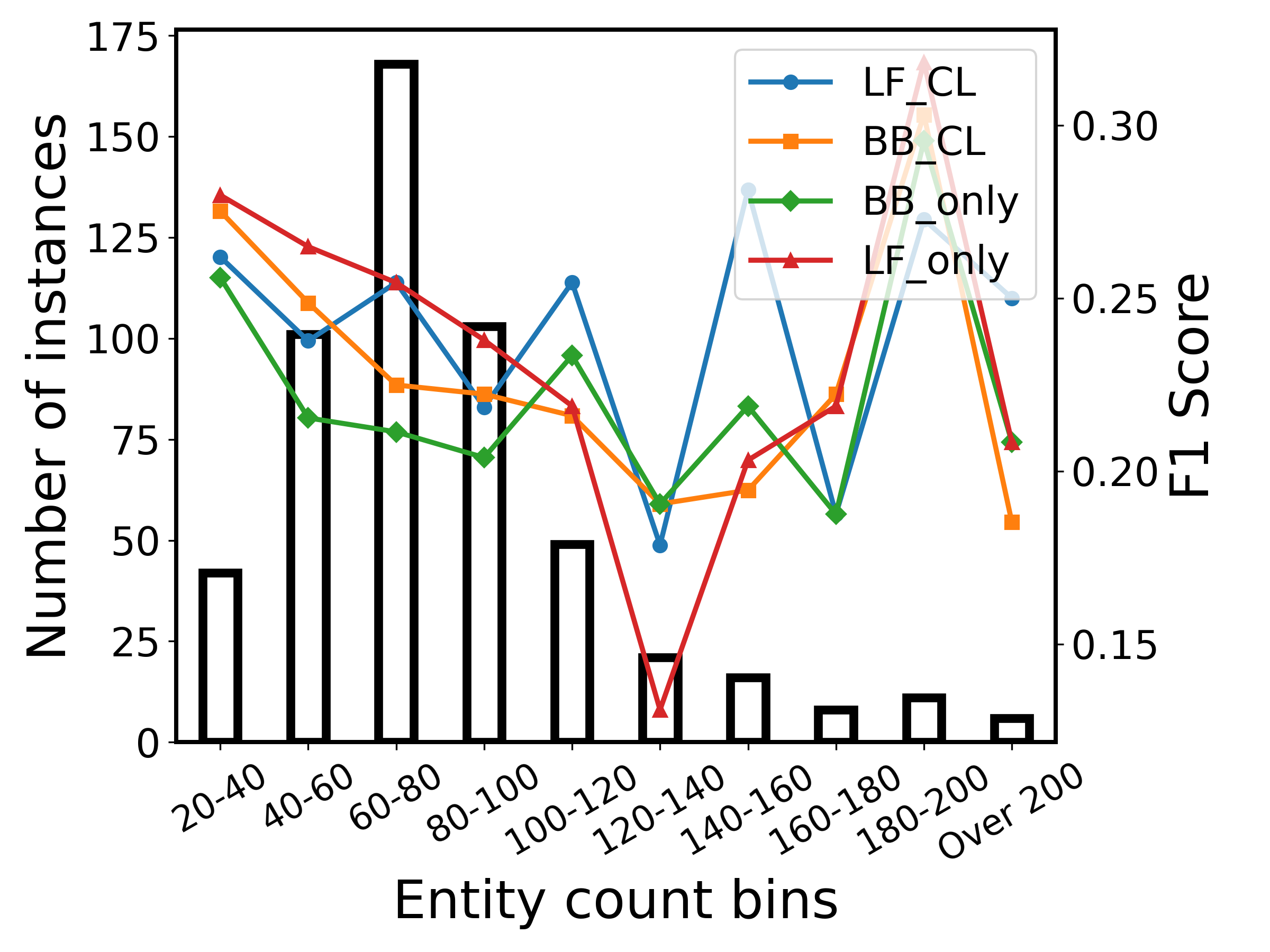}
    \caption{Paragraph \& Multi}
    \label{fig:entcnt_f1_para_multi_all}
\end{subfigure}
\hfill
\begin{subfigure}[t]{0.32\textwidth}
    \centering
    \includegraphics[width=\linewidth]{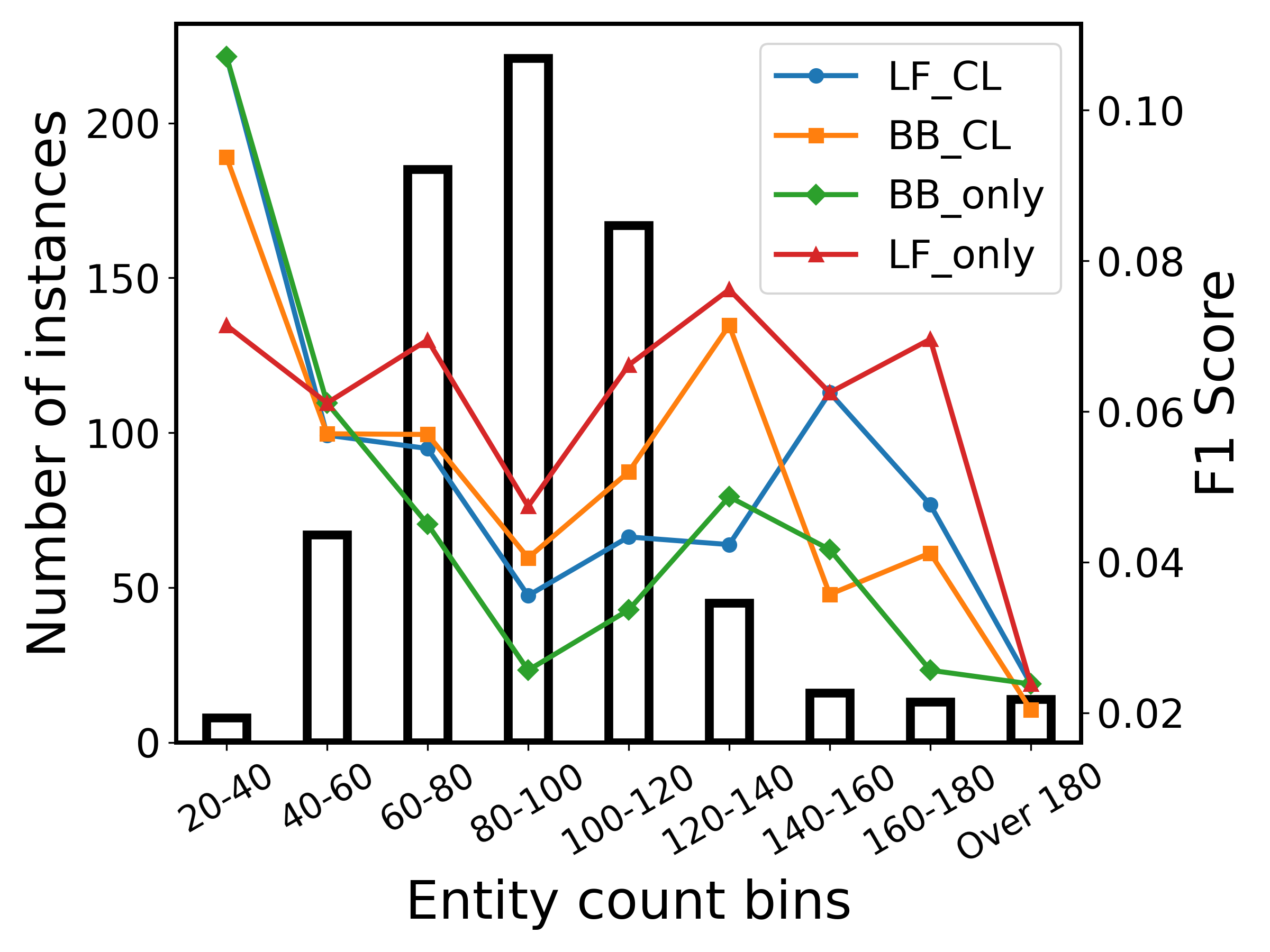}
    \caption{Table \& Multi}
    \label{fig:entcnt_f1_table_multi_all}
\end{subfigure}

\vspace{-0.8em}

\caption{
\textbf{Task 4: Entity Count vs. F1.}
Mean F1 per entity-count bin (lines) with bin sizes (bars).
The top row reports results over all instances, while the bottom row excludes zero-F1 cases.
Performance generally degrades as entity counts increase, with a more pronounced effect for paragraph-oriented settings.
}
\label{fig:task3_entitycount_f1}
\vspace{-1.2em}
\end{figure*}

\paragraph{Task 4: XML Entity Index Extraction}
This task further investigates multi-label entity classification without dependencies on bounding box extractions but instead, predicts indices directly from XML-defined entities, assessing how well models can reason over text-structured representations of documents. We examine two setups: 1) Concatenated inputs, where the question and all XML entities are fed as one sequence, and 2) Per-entity inputs, where text and optional CLIP-based image embeddings are pooled for each entity. Models tested include BigBird~\cite{zaheer2020bigbird} and Longformer~ \cite{Beltagy2020Longformer}. Metrics used are identical to Task 2.



\vspace{-0.5em}
\subsection{Additional Result Analysis}
\paragraph{Results of Task 1: Generative Reasoning} 

\begin{table*}[!t]
\scriptsize
\centering
\begin{tabularx}{\textwidth}{
    X | 
    X | 
    >{\centering\arraybackslash}p{0.10\columnwidth} | 
    >{\centering\arraybackslash}p{0.10\columnwidth} | 
    >{\centering\arraybackslash}p{0.20\columnwidth} | 
    >{\centering\arraybackslash}p{0.20\columnwidth}  
    >{\centering\arraybackslash}p{0.20\columnwidth}  
    >{\centering\arraybackslash}p{0.20\columnwidth}  
    >{\centering\arraybackslash}p{0.20\columnwidth}  
    }
    \hline
    \textbf{Model} & \textbf{Setup} & \textbf{Text} & \textbf{Image} 
    & \multicolumn{1}{c|}{\textbf{BLEU}} & \multicolumn{1}{c}{\textbf{Rouge1}} & \multicolumn{1}{c}{\textbf{Rouge2}} & \multicolumn{1}{c}{\textbf{RougeL}} & \multicolumn{1}{c}{\textbf{RougeLSum}} \\
    \hline
    InternVL & Question & - & \checkmark &         6.22 &	30.02 &	7.24 &	15.95 &	19.04 \\
    InternVL & Zero-shot & - & \checkmark &         5.54 &	28.67 &	7.03 &	14.62 &	19.46 \\
    InternVL & One-shot & - & \checkmark &         6.39 &	28.74 &	6.99 &	16.12 &	18.50 \\
    InternVL & Question & \checkmark & -  &         11.20 &	32.69 &	10.02 &	18.49 &	20.32 \\
    InternVL & Zero-shot & \checkmark & - &         12.65 &	33.36 &	11.11 &	18.93 &	21.40 \\
    InternVL & One-shot & \checkmark & -  &         14.53 &	32.11 &	11.99 &	19.56 &	20.91 \\
    InternVL & Question & \checkmark & \checkmark &         6.91 &	30.09 &	8.05 &	15.85 &	19.79 \\
    InternVL & Zero-shot & \checkmark & \checkmark &         6.91 &	30.06 &	8.48 &	15.55 &	20.60 \\
    InternVL & One-shot & \checkmark & \checkmark &         7.48 &	31.23 &	8.55 &	16.46 &	20.26 \\
    
    Qwen2.5 & Question & - & \checkmark &         15.06 &	36.64 &	13.43 &	20.79 &	23.89 \\
    Qwen2.5 & Zero-shot & - & \checkmark &         21.56 &	38.45 &	16.92 &	23.90 &	25.84 \\
    Qwen2.5 & One-shot & - & \checkmark &         12.40 &	34.48 &	11.49 &	19.50 &	22.53 \\
    Qwen2.5 & Question & \checkmark & - &         7.77 &	28.62 &	9.33 &	15.71 &	20.44 \\
    Qwen2.5 & Zero-shot & \checkmark & - &         9.66 &	30.81 &	10.93 &	17.28 &	21.96 \\
    Qwen2.5 & One-shot & \checkmark & - &         7.32 &	27.66 &	8.30 &	15.45 &	19.44 \\
    Qwen2.5 & Question & \checkmark & \checkmark &         9.80 &	31.89 &	10.70 &	17.82 &	22.12 \\
    Qwen2.5 & Zero-shot & \checkmark & \checkmark &         12.78 &	33.69 &	12.86 &	19.80 &	23.66 \\
    Qwen2.5 & One-shot & \checkmark & \checkmark &         9.45 &	31.35 &	9.96 &	17.94 &	21.40 \\
    Qwen3 & One-shot & \checkmark & - & 26.00 & - & - & 28.80 & - \\
    Qwen3 & One-shot & - & \checkmark & 18.20 & - & - & 21.00 & - \\
    Qwen3 & One-shot & \checkmark & \checkmark & 29.7 & - & - & 32.1 & - \\
    
    Gemini2.5 & Question & \checkmark & - &      7.34 &	28.26 &	10.91 &	15.28 &	20.44 \\
    Gemini2.5 & Zero-shot & \checkmark & - &      9.22 &	29.24 &	11.14 &	15.89 &	20.93 \\
    Gemini2.5 & One-shot & \checkmark & - &      11.94 &	24.55 &	12.85 &	16.81 &	19.41 \\
    Gemini2.5 & Question & - & \checkmark &      7.57 &	28.76 &	10.71 &	15.41 &	20.92 \\
    Gemini2.5 & Zero-shot & - & \checkmark &      9.03 &	29.62 &	12.08 &	16.20 &	21.88 \\
    Gemini2.5 & One-shot & - & \checkmark &      11.96 &	25.47 &	10.96 &	15.33 &	18.68 \\
    Gemini2.5 & Question & \checkmark & \checkmark &      7.38 &	28.85 &	10.55 &	15.15 &	20.46 \\
    Gemini2.5 & Zero-shot & \checkmark & \checkmark &      8.89 &	30.70 &	12.24 &	17.07 &	22.63 \\
    Gemini2.5 & One-shot & \checkmark & \checkmark &      11.71 &	24.57 &	12.61 &	16.30 &	18.95 \\
    Gemini3 & One-shot & \checkmark & \checkmark & \textbf{31.8} & - & - & \textbf{34.5} & - \\
    GPT4o & Question & - & \checkmark & 24.80 & 41.25 & 19.46 & 26.85 & 29.92 \\
    GPT4o & Zero-shot & - & \checkmark & 26.45 & 43.16 & 21.05 & 28.63 & 31.55 \\
    GPT4o & One-shot & - & \checkmark & 28.92 &  \textbf{45.79} &  \textbf{23.81} &  31.26 &  \textbf{34.12} \\
    \hline
\end{tabularx}
\caption{\textbf{Comprehensive results for generative reasoning task} using InternVL2, Qwen 2.5, Qwen 3, Gemini 2.5, Gemini 3, and GPT-4o across different prompt setups and input configuration. We measure performance through BLEU and ROUGE metrics.}
\label{tbl:generativeresults}
\vspace{-2em}
\end{table*}

As shown in Table~\ref{tbl:generativeresults}, the best performance is from Gemini3-Pro in a one-shot setting using text and image inputs with full-document ingestion, which achieves a BLEU score of 31.8 and strong performance across ROUGE-L (34.5).
This suggests that modern large vision-language models are more capable of locating relevant information and reasoning across multiple and complex document images and textual content.

Comparing input-processed architectures, Qwen2.5 and Qwen3 consistently outperforms InternVL2 in setups that utilize image input. This advantage likely stems from Qwen's greater input capacity, which allows it to process up to 100 document pages, while InternVL2 is limited to 10 with the simplest configuration. As a result, the Qwen models are better suited for large-scale document understanding, particularly when full visual context is required. However, InternVL2 demonstrates better performance when using text-only inputs compared to Qwen2.5, achieving a BLEU of 12.65 and ROUGE-2 of 11.11, which slightly surpasses Qwen2.5's text-only setup (BLEU: 9.66, ROUGE-2: 10.93). This implies that InternVL2’s language modeling capabilities are competitive when visual grounding is not required, and reaffirms the importance of architecture-modality alignment.

Interestingly, Qwen2.5 performs best when using image input alone, and worst when using only text. This indicates that Qwen2.5 is likely optimized for visual reasoning and may struggle to integrate text-only signals in the absence of layout structure. However, adding both image and text did not significantly boost performance, suggesting that naive concatenation of modalities may introduce redundancy or noise, and that more sophisticated cross-modal fusion strategies are needed for optimal performance on DocHop-QA. The more recent Qwen3, however, outperforms it's predecessor with better visual and textual alignment performing best (BLEU: 29.7, ROUGEL: 32.1) with both text and image inputs.

Prompt design plays a critical role for input processed generative QA performance. For Qwen2.5, zero-shot prompting consistently outperforms one-shot setups, likely due to the substantial length of context documents. Including a one-shot example in the prompt consumes valuable input space, which forces truncation of relevant information, especially under token limits. This finding highlights an important trade-off in prompting long-context models: example-rich prompts can backfire in high-volume settings like full-document QA.

We also evaluated Gemini 2.5 and Gemini 3 across XML and PDF inputs and GPT-4o with PDF inputs. Unlike InternVL2, Qwen2.5, and Qwen3, these models were accessed via API and thus were not bound by local memory constraints, allowing full document inputs without severe truncation. As a result, one-shot prompting performed best across most configurations, with top BLEU score from Gemini3 of 31.8 using whole XML \& PDF input.
Gemini3 outperforms any other models significantly, demonstrating its superior ability in document understanding and multimodal QA-based reasoning. Comparing to Qwen3-VL, which is one of the latest vision-language models Gemini3 also outperforms its results (BLEU: 29.7).
These results illustrate the potential of cloud-based LLMs for processing full-document QA tasks without aggressive input filtering, further expanding the applicability of DocHop-QA in real-world settings.

In summary, this task underscores how DocHop-QA enables robust benchmarking of free-form QA grounded in multimodal, multi-page and cross-document setups. Users can evaluate generative models under various prompting and input modality combinations, uncovering limitations in memory, alignment, and format-following behavior. 
\begin{figure}[h]
    \centering
    \begin{subfigure}[t]{0.49\columnwidth}
        \centering
        \includegraphics[width=\linewidth]{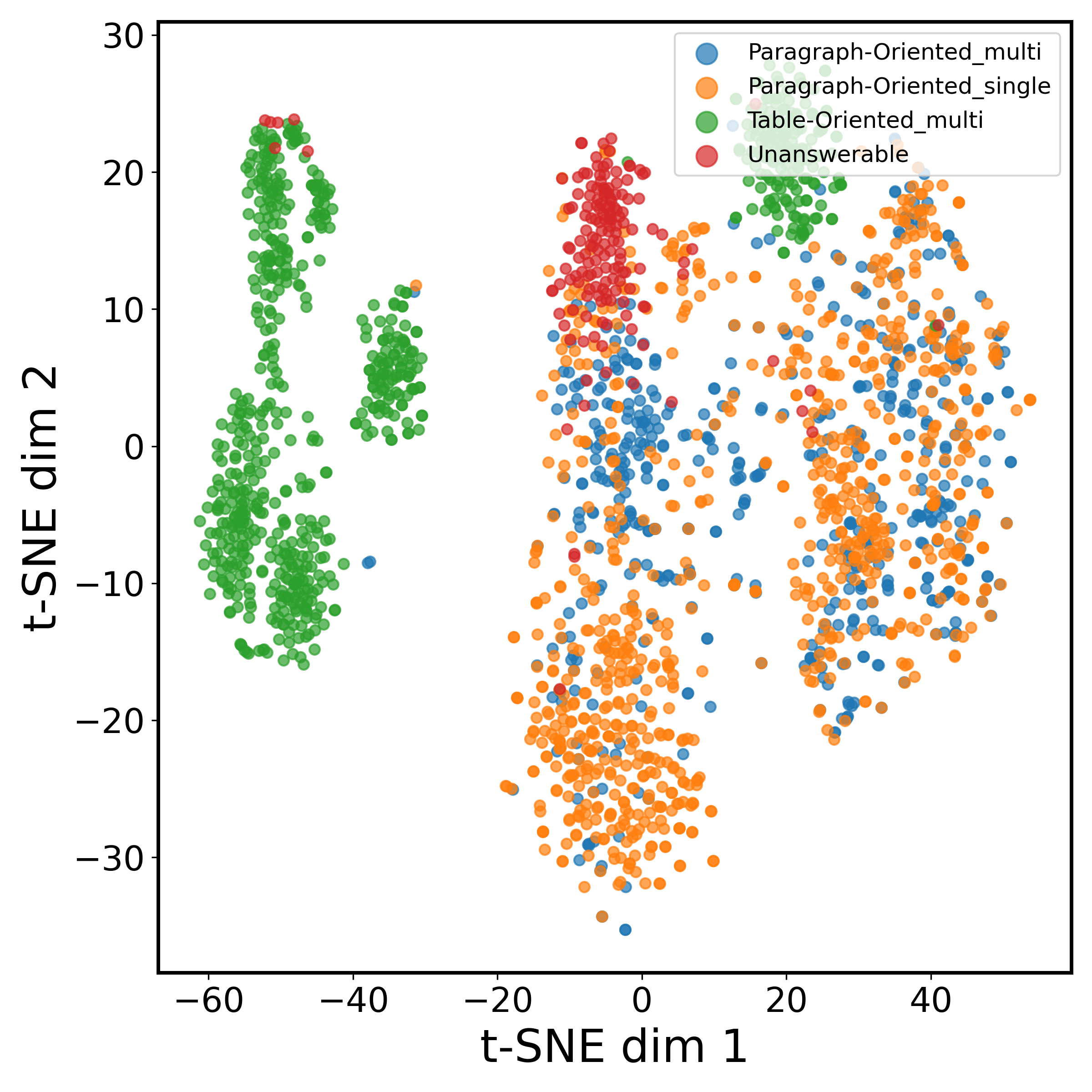}
        \caption{Image-only input}
        \label{fig:qwen_tsne_image}
    \end{subfigure}\hfill
    \begin{subfigure}[t]{0.49\columnwidth}
        \centering
        \includegraphics[width=\linewidth]{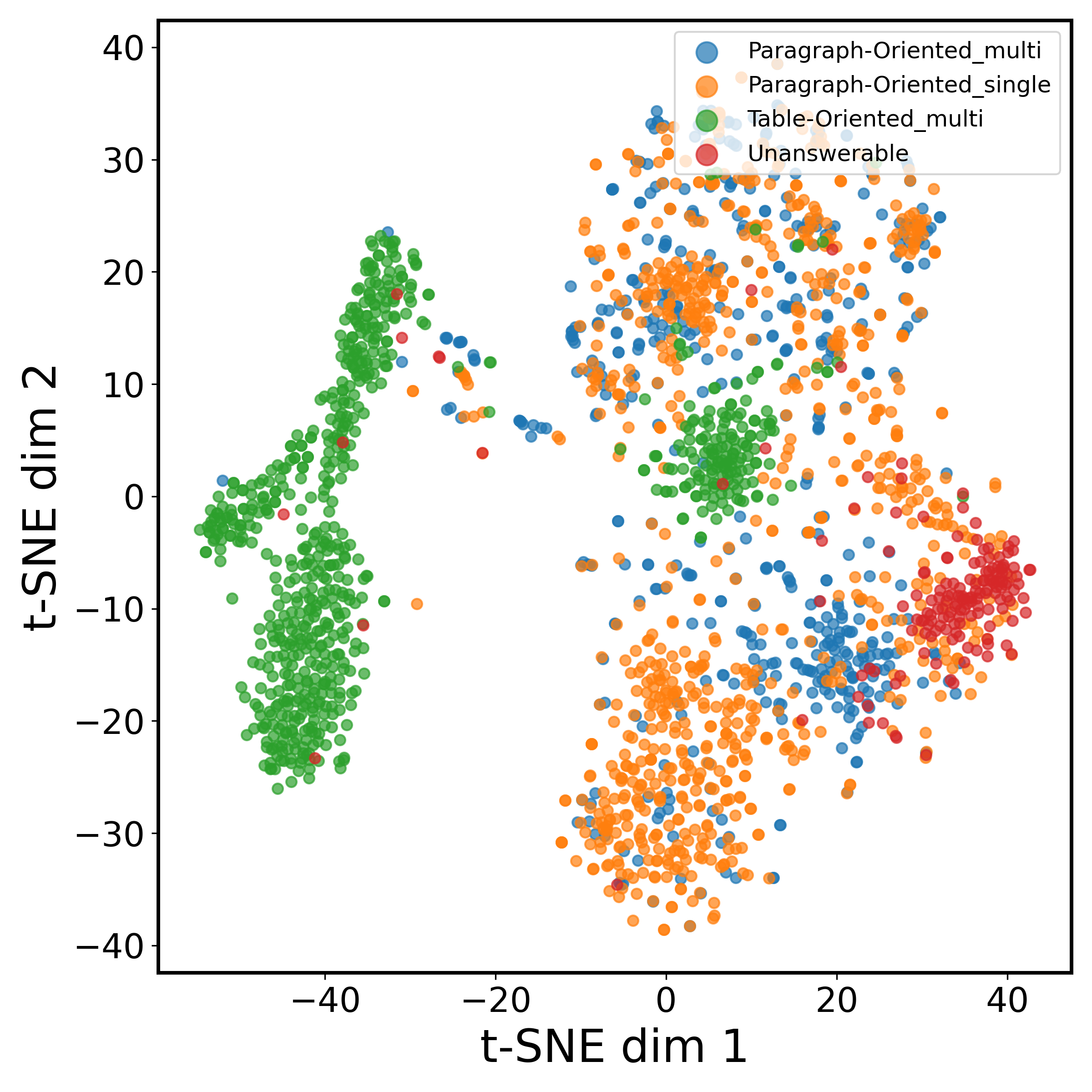}
        \caption{Text-only (XML) input}
        \label{fig:qwen_tsne_xml}
    \end{subfigure}
    \vspace{-0.5em}
    \caption{\textbf{Task 1: t-SNE of Qwen last layer hidden representations}
    Visualization under image-only and text-only (XML) inputs, analyzing how LLMs and VLMs differentiate question types and contextual complexity. Table and paragraph-oriented instances form separable clusters in both settings, and unanswerable cases show a distinct grouping, suggesting modality-consistent internal structuring of question types.}
    \label{fig:qwen_tsne}
    \vspace{-1.8em}
\end{figure}

\vspace{-0.7em}

\subparagraph{Embedding Analysis for Question Types}
We analyze the last-layer hidden states of the Qwen model using t-SNE in Figure~\ref{fig:qwen_tsne} to understand how LLMs and LVLMs internally differentiate question types and context complexities. Clear separation emerges between table-oriented and paragraph-oriented questions for both text-only and image-only inputs, suggesting that the model can distinguish question types at a coarse level, likely driven by surface-level cues in question formulation rather than explicit tabular reasoning.
This indicates an awareness of question format and required evidence type, but does not imply effective reasoning over tabular content, which is consistent with the substantially lower performance observed on table-oriented tasks. Additionally, we identify a distinct cluster corresponding to unanswerable questions, demonstrating that recent LLMs can partially detect answerability without fine-tuning. However, this separation is less pronounced for paragraph-oriented and fan-hop questions, where partial answers may be available but full reasoning chains cannot be completed. This observation reflects the inherent difficulty of DocHop-QA unanswerable cases and highlights the tendency of dataset-generation LLMs to introduce hallucinated intermediate hops. Overall, the similarity of clustering patterns across text-only and image-only inputs suggests consistent internal representations across modalities, while reinforcing the need for more robust reasoning and answerability-aware modeling.
\vspace{-1.5em}

\paragraph{Results of Task 2: Structured Generative Answering}
We prompt generative vision-language models (InternVL2, Qwen2.5-VL, GPT-4o) to output structured entity indices. Inputs include labeled entity texts and optionally, document page images. Prompts follow zero-shot or one-shot formats. Table~\ref{tbl:structuredgenerativeresults} shows that GPT-4o in the zero-shot, text-only configuration yields the highest F1 (16.21). Visual input degrades performance due to layout misalignment. One-shot prompts reduce input space, harming performance on long documents. This task highlights the potential of LLMs for structured reasoning, but also the limits of current multimodal alignment.

\vspace{-0.8em}
\paragraph{Results of Task3: BBox Entity Index Extraction}
This task features multi-label classification over OCR-extracted bounding boxes. As shown in Table~\ref{tbl:bboxresults}, 
LayoutLMv3 significantly outperforms LayoutXLM in overlap-based accuracy (95.52\%) and F1 (12.35), showing its robustness in spatially grounding questions. However, full containment accuracy is near zero, suggesting difficulties in aligning complex layouts.
These findings show that DocHop-QA is a valuable benchmark for pushing the limits of document layout understanding in QA settings. It reveals critical challenges in handling long, multi-page, multi-document inputs, while offering fine-grained bounding box supervision to support model development. DocHop-QA may be used to design new training strategies (e.g., multi-image stitching, hierarchical encoding) or pretraining tasks that improve entity-level reasoning under spatial and modality constraints.


\begin{table}[h]
\scriptsize
\centering
\setlength{\tabcolsep}{4pt} 
\begin{tabularx}{\columnwidth}{
    l |                      
    *{3}{>{\centering\arraybackslash}X} |  
    *{3}{>{\centering\arraybackslash}X}
}
\hline
& \multicolumn{3}{c|}{\textbf{Sample}} & \multicolumn{3}{c}{\textbf{Accuracy}} \\
\textbf{Model} & \textbf{F1} & \textbf{Re} & \textbf{Pr} & \textbf{Be} & \textbf{Co} & \textbf{Ov} \\
\hline
LayoutLMv3 & \textbf{12.35} & 34.97 & 8.47 & 0.00 & 0.04 & 95.52 \\
LayoutXLM  & 7.13 & 4.49 & 24.06 & 15.64 & 0.00 & 33.96 \\
\hline
\end{tabularx}
\caption{
\textbf{Results on BBox Entity Index Extraction} reporting sample-level F1, Recall (Re), and Precision (Pr), along with spatial accuracy: Belong (Be), Contain (Co), and Overlap (Ov), for LayoutLMv3 and LayoutXLM.
}
\vspace{-1.8em}
\label{tbl:bboxresults}
\end{table}

\begin{table}[t]
\scriptsize
\centering
\setlength{\tabcolsep}{3pt}
\renewcommand{\arraystretch}{1.1}
\begin{tabularx}{\columnwidth}{
    l
    c
    c|                       
    *{3}{>{\centering\arraybackslash}X}|  
    *{3}{>{\centering\arraybackslash}X}
}
\hline
& & & \multicolumn{3}{c|}{\textbf{Sample}}   
& \multicolumn{3}{c}{\textbf{Accuracy}} \\
\textbf{Setup} & \textbf{Text} & \textbf{Img}
& \textbf{F1} & \textbf{Re} & \textbf{Pr}
& \textbf{Be} & \textbf{Co} & \textbf{Ov} \\
\hline
Concat & BB & -- & 23.18 & 18.81 & 30.45 & 4.22 & 0.09 & 62.21 \\
Concat & LF & -- & 21.83 & 20.09 & 24.12 & 0.22 & 0.00 & 66.17 \\
PerEnt & BB & -- & 21.25 & 17.22 & 27.94 & 2.07 & 0.09 & 58.96 \\
PerEnt & LF & -- & 22.99 & 18.71 & 30.08 & 3.25 & 0.00 & 63.80 \\
PerEnt & BB & CL & \textbf{23.24} & 21.29 & 25.77 & 0.48 & 0.09 & 64.98 \\
PerEnt & LF & CL & 23.18 & 18.81 & 30.45 & 4.22 & 0.09 & 62.21 \\
\hline
\end{tabularx}
\caption{
\textbf{Results for XML Entity Index Extraction.} We compared concatenated (Concat) vs. per-entity (PerEnt) setups across text-only and text+image CLIP (CL) inputs with BigBird (BB) and Longformer (LF). 
}
\vspace{-2em}

\label{tbl:xmlresults}
\end{table}

\paragraph{Results of Task4: XML Entity Index Extraction}
We further investigate multi-label entity classification using XML-based entity indices. The best F1 score (23.24) was achieved using BigBird with both text and image inputs in the per-entity setup (Table \ref{tbl:xmlresults}), indicating that visual context improves model understanding and highlighting the value of layout-grounded visual features.
Model comparisons reveal complementary strengths: BigBird excels in capturing long-range context in the text-concatenated setup, while Longformer performs better on segmented inputs. Visual inputs further boost per-entity performance, showing the benefit of aligning text with document layout.
DocHop-QA exposes label bias and modality interaction challenges, providing a benchmark for developing models that integrate textual, visual, and positional cues in entity-level reasoning.

\vspace{-1em}
\subparagraph{Impact of Context Entity Cardinality on Model Performance}
To investigate how contextual complexity affects model performance, we analyze Task 4 performance as a function of the number of context entities involved in each question instance described in Figure~\ref{fig:task3_entitycount_f1}. Across both text-only and text-plus-image settings, we observe a general degradation in F1 score as the number of entities increases, particularly for paragraph-oriented instances. For single-document questions, performance declines noticeably as entity counts approach 80, while for multi-document questions, a similar trend appears around 100 entities. This degradation likely reflects increased noise from repetitive or overly detailed contextual information, such as experimental settings or hyperparameters that are irrelevant to the question. Interestingly, a modest performance increase is observed beyond these thresholds, which we attribute to the small number of instances in higher entity-count bins. For table-oriented, multi-document instances, 
overall scores remain lower than paragraph-oriented counterparts. This gap underscores the difficulty of contextualizing tabular information within large entity sets and suggests that improved mechanisms for entity selection and table-aware reasoning are crucial for addressing complex DocHop-QA instances.
\vspace{-0.9em}
\subsection{Hyperparameters}
\label{appendix:hyperparams}
We provide the best hyperparameter setups in Tables \ref{tbl:hyperparametersgenerative} to \ref{tbl:hyperparametersxml}. For the generative models in Task 1 and Task 2, we controlled the number of input entities and page images for input processed setups to manage memory issues. Maximum tokens are set to 1024 for all setups. Being a smaller model, InternVL2 only ran successfully with a maximum of 25 entities and 10 pages. Qwen2.5 could handle up to 200 entities and 100 page images but, interestingly, performed best with only 10 pages. We mirror the best setup from Qwen2.5 for experiments done for Qwen3. For fine-tuned models in Task 3 and Task 4, we explore different values for learning rate \{2e-01, 2e-02, 2e-03, 2e-04, 2e-05\}, early stop patience \{None, 5, 10\}, and warm up \{None, 10\}. For learning rates, 2e-02 generally produces a better F1 score, however, predicted indices tend to skew to the more frequent entity indices, raising recall but sacrificing precision. 
For closed-source models, we follow the hyperparameter setup of the best performing open-source model when possible.

\begin{table}[h]
\scriptsize
\centering
\setlength{\tabcolsep}{2.3pt}
\renewcommand{\arraystretch}{1.15}
\begin{tabularx}{\columnwidth}{
    X 
    c
    *{5}{>{\centering\arraybackslash}X}
}
\hline
\textbf{Model}
& \textbf{Setup}
& \textbf{Text}
& \textbf{Image}
& \textbf{\shortstack{Max\\Entities}}
& \textbf{\shortstack{Max\\Pages}}
& \textbf{\shortstack{Max \\ Tokens}} \\
\hline
InternVL & Question  & -- & \checkmark & --  & 10 & 1024 \\
InternVL & Zero-shot & -- & \checkmark & --  & 10 & 1024 \\
InternVL & One-shot  & -- & \checkmark & --  & 5  & 1024 \\
InternVL & Question  & \checkmark & -- & 25 & -- & 1024 \\
InternVL & Zero-shot & \checkmark & -- & 25 & -- & 1024 \\
InternVL & One-shot  & \checkmark & -- & 25 & -- & 1024 \\
InternVL & Question  & \checkmark & \checkmark & 25 & 5 & 1024 \\
InternVL & Zero-shot & \checkmark & \checkmark & 25 & 5 & 1024 \\
InternVL & One-shot  & \checkmark & \checkmark & 25 & 5 & 1024 \\
Qwen2.5    & Question  & -- & \checkmark & --  & 10 & 1024 \\
Qwen2.5     & Zero-shot & -- & \checkmark & --  & 10 & 1024 \\
Qwen2.5     & One-shot  & -- & \checkmark & --  & 10 & 1024 \\
Qwen2.5     & Question  & \checkmark & -- & 200 & -- & 1024 \\
Qwen2.5     & Zero-shot & \checkmark & -- & 200 & -- & 1024 \\
Qwen2.5     & One-shot  & \checkmark & -- & 50  & -- & 1024 \\
Qwen2.5     & Question  & \checkmark & \checkmark & 100 & 10 & 1024 \\
Qwen2.5     & Zero-shot & \checkmark & \checkmark & 100 & 10 & 1024 \\
Qwen2.5     & One-shot  & \checkmark & \checkmark & 50  & 10 & 1024 \\
\hline
\end{tabularx}
\caption{\textbf{Generative reasoning best model hyperparameters.}}
\label{tbl:hyperparametersgenerative}
\vspace{-1em}
\end{table}

\begin{table}[h]
\scriptsize
\centering
\setlength{\tabcolsep}{2.3pt}
\renewcommand{\arraystretch}{1.15}
\begin{tabularx}{\columnwidth}{
    X
    c
    *{5}{>{\centering\arraybackslash}X}
}
\hline
\textbf{Model}
& \textbf{Setup}
& \textbf{Text}
& \textbf{Image}
& \textbf{\shortstack{Max\\Entities}}
& \textbf{\shortstack{Max\\Pages}}
& \textbf{\shortstack{Max\\Tokens}} \\
\hline
InternVL & Zero-shot & \checkmark & -- & 25  & -- & 1024 \\
InternVL & Zero-shot & \checkmark & \checkmark & 25 & 5 & 1024 \\
InternVL & One-shot  & \checkmark & -- & 25  & -- & 1024 \\
InternVL & One-shot  & \checkmark & \checkmark & 25 & 5 & 1024 \\
Qwen2.5     & Zero-shot & \checkmark & -- & 100 & -- & 1024 \\
Qwen2.5     & Zero-shot & \checkmark & \checkmark & 50 & 10 & 1024 \\
Qwen2.5     & One-shot  & \checkmark & -- & 50  & -- & 1024 \\
Qwen2.5     & One-shot  & \checkmark & \checkmark & 50 & 10 & 1024 \\
\hline
\end{tabularx}
\caption{\textbf{Structured generative answering best model hyperparameters.}}
\label{tbl:hyperparametersstructgen}
\vspace{-0.8em}
\end{table}

\begin{table}[h]
\scriptsize
\centering
\setlength{\tabcolsep}{1.2mm}
\begin{tabularx}{\columnwidth}{
    X | 
    >{\centering\arraybackslash}p{0.15\columnwidth} |
    >{\centering\arraybackslash}p{0.12\columnwidth} |
    >{\centering\arraybackslash}p{0.12\columnwidth} |
    >{\centering\arraybackslash}p{0.1\columnwidth} |
    >{\centering\arraybackslash}p{0.1\columnwidth}
    }
    \hline
    \textbf{Model} &	\textbf{Learning Rate} &	\textbf{Batch Size} &	\textbf{Max Len} &	\textbf{Early Stop} &	\textbf{Warm Up} \\
    \hline
    LayoutLMv3 &    2e-02 &		32 &	512 &	10 &	10 \\
    LayoutXLM &	    2e-05 & 	64 &	512 &	10 &	10 \\
    
    \hline
\end{tabularx}
\caption{\textbf{BBox Entity Index Extraction best model hyperparameters.}}
\label{tbl:hyperparametersbbox}
\vspace{-1em}
\end{table}

\begin{table}[h]
\scriptsize
\centering
\setlength{\tabcolsep}{2.5pt}
\renewcommand{\arraystretch}{1.15}
\begin{tabularx}{\columnwidth}{
    l c c c
    >{\centering\arraybackslash}p{0.14\columnwidth}
    >{\centering\arraybackslash}p{0.12\columnwidth}
    >{\centering\arraybackslash}X
    >{\centering\arraybackslash}X
}
\hline
\textbf{Setup}
& \textbf{\shortstack{Text\\Emb}}
& \textbf{\shortstack{Img\\Emb}}
& \textbf{\shortstack{Max\\Len}}
& \textbf{\shortstack{Learning\\Rate}}
& \textbf{\shortstack{Early\\Stop}}
& \textbf{\shortstack{Batch\\Size}}
& \textbf{\shortstack{Warm\\Up}} \\
\hline
Concat & BB & --   & 2048 & 2e-02 & 5 & 16 & 10 \\
Concat & LF & --   & 2048 & 2e-02 & 5 & 8  & 10 \\
PerEnt & BB & --   & 4096 & 2e-02 & 0 & 64 & 0  \\
PerEnt & LF & --   & 4096 & 2e-02 & 0 & 64 & 0  \\
PerEnt & BB & CLIP & 4096 & 2e-02 & 5 & 64 & 10 \\
PerEnt & LF & CLIP & 4096 & 2e-02 & 5 & 64 & 10 \\
\hline
\end{tabularx}
\caption{\textbf{XML Entity Index Extraction best model hyperparameters.} Concat: Concatenated, PerEnt: Per Entity, BB: BigBird, LF: Longformer}
\label{tbl:hyperparametersxml}
\end{table}

\vspace{-0.5em}
\section{Ablation Studies}
\label{appendix:ablationstudies}

\paragraph{Human Answerability and Realism Analysis}
To further examine whether DocHop-QA reflects realistic and answerable scientific information-seeking scenarios, we conducted a human evaluation on 50 QA instances covering diverse reasoning concepts and hop types. Three domain experts (1 CS PhD, 1 Biomedical Engineering PhD, 1 Medical Science PhD) independently answered the questions and evaluated answerability confidence.  As shown in Table~\ref{tab:human_answerability}, 84\% of the questions were judged answerable with an average confidence score of 4.1/5. Human-generated answers also achieved substantially higher overlap with reference answers than current model outputs. Notably, even independently written human answers achieve a ROUGE-L score of only 58.6 against the synthesized gold reference, rather than approaching 100; this indicates that gold answers are not intended to represent the sole correct phrasing, and that our lexical-overlap evaluation implicitly tolerates variation in correct responses. These results suggest that DocHop-QA does not primarily consist of ill-posed or artificially constructed queries, but instead requires evidence-grounded multi-document reasoning.

\begin{table}[h]
\centering
\small

\begin{tabular}{lc}
\toprule
\textbf{Metric} & \textbf{Result} \\
\midrule
Answerable (\%) & 84 \\
Confidence (1--5) & 4.1 \\
Human vs. Gold (ROUGE-L) & 58.6 \\
\bottomrule
\end{tabular}%

\caption{Human evaluation on 50 instances.}
\label{tab:human_answerability}
\vspace{-0.8em}
\end{table}



\clearpage
\section{Supplementary Materials}
\label{appendix:supplementary}

\begin{table}[!ht]
\small
\centering
\begin{tabular}{p{0.95\linewidth}}
\toprule
\textbf{\textit{Few-shot Example A}} (high scores) \\
\textbf{Question:} How does L-carnitine differ from black tea in protecting LDL and liver lipids from oxidation? \\
\textbf{Snippets:} Snippet 1 describes L-carnitine protecting LDL lipids and proteins. Snippet 2 describes black tea reducing liver lipid peroxidation and maintaining glutathione levels. \\
\vspace{0.2em}
\textbf{Output:} \\
\texttt{\{"reality\_fluency": \{"score": 4, "reason": "Clear, scientifically meaningful comparative question."\},} \\
\texttt{"accuracy": \{"score": 4, "reason": "Both snippets directly correspond to the two subjects of the question."\},} \\
\texttt{"completeness": \{"score": 3, "reason": "Mechanisms are described but direct comparative efficacy data is missing."\}\}} \\
\midrule
\textbf{\textit{Few-shot Example B}} (mid scores) \\
\textbf{Question:} How do DNMT epigenetic mechanisms in lung cancer and Arab-American health disparities differ in disease prevention? \\
\textbf{Snippets:} All 5 snippets focus on DNMT and lung tumorigenesis; Arab-American health disparities are not mentioned. \\
\vspace{0.2em}
\textbf{Output:} \\
\texttt{\{"reality\_fluency": \{"score": 3, "reason": "Complex but interpretable as a comparative research question."\},} \\
\texttt{"accuracy": \{"score": 2, "reason": "Snippets only cover one of the two required topics."\},} \\
\texttt{"completeness": \{"score": 2, "reason": "Only half the question can be addressed; the other topic has no coverage."\}\}} \\
\midrule
\textbf{\textit{Few-shot Example C}} (mid-high scores) \\
\textbf{Question:} What surgical approaches for Moyamoya disease and pulmonary artery stenosis show better outcomes? \\
\textbf{Snippets:} Snippet 1 details revascularization for Moyamoya. Snippet 2 covers balloon dilation for pulmonary stenosis. Snippet 3 discusses general pediatric cardiac outcomes. \\
\vspace{0.2em}
\textbf{Output:} \\
\texttt{\{"reality\_fluency": \{"score": 3, "reason": "Somewhat complex but clearly understandable as a surgical outcomes comparison."\},} \\
\texttt{"accuracy": \{"score": 3, "reason": "Two snippets directly address the two conditions; the third is loosely related."\},} \\
\texttt{"completeness": \{"score": 3, "reason": "Relevant surgical detail provided for both, though direct outcome comparisons are not explicit."\}\}} \\
\bottomrule
\end{tabular}
\caption{\textbf{Few-shot examples provided to the LLM evaluator.} Examples A, B, and C illustrate high, mid, and mid-high scoring cases respectively.}
\label{tab:eval-fewshot}
\end{table}

\begin{table}[!t]
\small
\begin{tabular}{p{0.95\linewidth}}
\toprule
\textbf{\textit{System Prompt:}} You are an evaluator for a multi-hop QA dataset called DocHop-QA. Evaluate each QA instance on the following three criteria using a 4-point scale. \\
\vspace{0.2em}
\textbf{1. Reality/Fluency:} 4=realistic and coherent; 3=slightly awkward but interpretable; 2=unnatural or hard to interpret; 1=incomprehensible. \\
\textbf{2. Accuracy:} 4=all snippets directly relevant; 3=mostly relevant; 2=tangentially related; 1=entirely unrelated. \\
\textbf{3. Completeness:} 4=all parts addressed; 3=mostly covered; 2=partial coverage only; 1=no useful information. \\
\vspace{0.2em}
\textbf{\textit{Input:}} \textbf{Question:} \{question\} \quad \textbf{Snippets:} \{snippets\} \\
\vspace{0.2em}
\textbf{\textit{Output:}} \texttt{\{"reality\_fluency": \{"score": <1-4>, "reason": "..."\}, "accuracy": \{"score": <1-4>, "reason": "..."\}, "completeness": \{"score": <1-4>, "reason": "..."\}\}} \\
\bottomrule
\end{tabular}
\caption{\textbf{LLM evaluator prompt for dataset quality assurance.} Three criteria are assessed on a 4-point scale with structured JSON output.}
\label{tab:eval-prompt}
\end{table}

\end{document}